\documentclass[journal]{IEEEtran}

\usepackage[utf8]{inputenc}
\usepackage[T1]{fontenc}
\usepackage{graphicx}
\DeclareGraphicsExtensions{.pdf,.png}
\usepackage{xcolor}
\usepackage[cmex10]{amsmath}
\usepackage{amsthm,amssymb,amsfonts,mathrsfs,bm,mathtools}
\mathtoolsset{centercolon=true}
\DeclareFontFamily{OT1}{pzc}{}
\DeclareFontShape{OT1}{pzc}{m}{it}{<-> s * [1.10] pzcmi7t}{}
\DeclareMathAlphabet{\mathpzc}{OT1}{pzc}{m}{it}
\DeclareFontFamily{U}{BOONDOX-calo}{\skewchar\font=45 }
\DeclareFontShape{U}{BOONDOX-calo}{m}{n}{<-> s*[1.05] BOONDOX-r-calo}{}
\DeclareFontShape{U}{BOONDOX-calo}{b}{n}{<-> s*[1.05] BOONDOX-b-calo}{}
\DeclareMathAlphabet{\mathcalboondox}{U}{BOONDOX-calo}{m}{n}
\SetMathAlphabet{\mathcalboondox}{bold}{U}{BOONDOX-calo}{b}{n}
\DeclareMathAlphabet{\mathbcalboondox}{U}{BOONDOX-calo}{b}{n}

\usepackage{newfloat}
\usepackage[caption=false]{subfig}
\usepackage{setspace}
\usepackage[inline]{enumitem}
\usepackage{algorithm}
\usepackage{algpseudocode}

\usepackage{etoolbox}
\makeatletter
\expandafter\patchcmd\csname\string\algorithmic\endcsname{\itemsep\z@}{\itemsep=.6ex plus1pt}{}{}
\makeatother
\AtBeginEnvironment{algorithmic}{\setstretch{1}}
\usepackage[hyphens]{url}
\usepackage[noadjust,sort,compress]{cite}

\usepackage{xspace}

\usepackage{tikz}
\usetikzlibrary{
  calc,
  trees,
  positioning,
  chains,
  decorations.pathreplacing,
  decorations.pathmorphing,
  decorations.shapes,
  decorations.text,
  decorations.markings,
  shapes,
  shapes.geometric,
  shapes.symbols,
  matrix,
  patterns,
  intersections,
  fit
}

\pgfkeys{tikz/mymatrixenv/.style={decoration=brace, nodes =
{font=\footnotesize}, every left delimiter/.style = {xshift=3pt}, 
every right delimiter/.style={xshift=-3pt}}}
\pgfkeys{tikz/mymatrix/.style={matrix of math nodes, left
delimiter=[,right delimiter= {]}, inner sep=1pt, column sep=.25em,
row sep=.25em, nodes={inner sep=0pt}}}
\pgfkeys{tikz/mymatrixbrace/.style={decorate,thick}}
\newcommand\mymatrixbraceoffseth{.5em}
\newcommand\mymatrixbraceoffsetv{.5em}

\tikzset{%
  add/.style args={#1 and #2}{to path={%
($(\tikztostart)!-#1!(\tikztotarget)$)--($(\tikztotarget)!-#2!(\tikztostart)$)%
  \tikztonodes}}
} 

\newcommand*\mymatrixbraceleft[4][m]{
  \draw[mymatrixbrace] ($(#1.north east)!(#1-#2-1.north east)!(#1.south
  east)+    (\mymatrixbraceoffseth,0)$) 
  -- node[right=2pt] {#4} 
  ($(#1.north east)!(#1-#3-1.south east)!(#1.south east)+
  (\mymatrixbraceoffseth,0)$); 
}
\newcommand*\mymatrixbracetop[4][m]{
  \draw[mymatrixbrace] ($(#1.north west)!(#1-1-#2.north west)!(#1.north
  east)+(0,\mymatrixbraceoffsetv)$) 
  -- node[above=2pt] {#4} 
  ($(#1.north west)!(#1-1-#3.north east)!(#1.north
  east)+(0,\mymatrixbraceoffsetv)$); 
}

\newcommand{\Integer}{\mathbb{Z}}
\newcommand{\IntegerP}{\mathbb{Z}_{\geq 0}}
\newcommand{\IntegerPP}{\mathbb{Z}_{>0}}
\newcommand{\Real}{\mathbb{R}}

\newcommand{\RealPP}{\mathbb{R}_{>0}}

\newcommand\given{{\mathbin{}\mid\mathbin{}}}
\newcommand\vect[1]{\mathbf{#1}}

\providecommand\given{} % so it exists
\newcommand\SetSymbol[1][]{
  \nonscript\,#1\vert \allowbreak \nonscript\,\mathopen{}}
\DeclarePairedDelimiterX\Set[1]{\lbrace}{\rbrace}%
{ \renewcommand\given{\SetSymbol[\delimsize]} #1 }
\DeclarePairedDelimiterX\innerp[2]{\langle}{\rangle}{#1
  \mathop{}\delimsize\vert\mathop{} #2}
\DeclarePairedDelimiterX\norm[1]\lVert\rVert{\ifblank{#1}{\:\cdot\:}{#1}}

\DeclareMathOperator{\linspan}{span}
\DeclareMathOperator{\trace}{trace}

\DeclareMathOperator{\diag}{diag}

\DeclareMathOperator{\expect}{\mathbb{E}}

\DeclareMathOperator*{\Argmin}{arg\,min}

\DeclareMathOperator{\dist}{dist}
\DeclareMathOperator{\minor}{Minor}

\let\oldsqrt\sqrt
\def\sqrt{\mathpalette\DHLhksqrt}
\def\DHLhksqrt#1#2{%
\setbox0=\hbox{$#1\oldsqrt{#2\,}$}\dimen0=\ht0
\advance\dimen0-0.2\ht0
\setbox2=\hbox{\vrule height\ht0 depth -\dimen0}%
{\box0\lower0.4pt\box2}}

\newtheoremstyle{kostasstyle}% name of the style to be used
{2pt}% measure of space to leave above the theorem. E.g.: 3pt
{2pt}% measure of space to leave below the theorem. E.g.: 3pt
{\normalfont}% name of font to use in the body of the theorem
{0pt}% measure of space to indent
{\bfseries}% name of head font
{.}% punctuation between head and body
{1ex}% space after theorem head; " " = normal interword space
{}% Manually specify head

\theoremstyle{kostasstyle}

\newtheorem{prop}{Proposition}

\makeatletter

\newcommand*{\ie}{%
  \@ifnextchar{,}%
  {\textit{i.e.}}%
  {\textit{i.e.,}\@\xspace}%
}
\newcommand*{\eg}{%
  \@ifnextchar{,}%
  {\textit{e.g.}}%
  {\textit{e.g.,}\@\xspace}%
}
\newcommand*{\etc}{%
  \@ifnextchar{.}%
  {\textit{etc}}%
  {\textit{etc.}\@\xspace}%
}
\newcommand*{\etal}{%
  \@ifnextchar{.}%
  {\textit{et al}}%
  {\textit{et al.}\@\xspace}%
}
\newcommand*{\cf}{%
  \@ifnextchar{.}%
  {\textit{cf}}%
  {\textit{cf.}\@\xspace}%
}
\newcommand*{\aka}{%
  \@ifnextchar{,}%
  {\textit{a.k.a.}}%
  {\textit{a.k.a.}\@\xspace}%
}
\makeatother

\begin{document}

\title{Riemannian-geometry-based modeling and clustering of
  network-wide non-stationary\\ time series: The brain-network
  case}

\author{Konstantinos~Slavakis,~\IEEEmembership{Senior~Member,~IEEE,}
  Shiva~Salsabilian, David~S.~Wack, Sarah~F.~Muldoon,
  Henry~E.~Baidoo-Williams, Jean~M.~Vettel, Matthew~Cieslak, and
  Scott~T.~Grafton%
  \thanks{\textit{K.~Slavakis}\/ and \textit{S.~Salsabilian}\/
    are with the Dept.~of Electrical Eng., Univ.~at Buffalo (UB),
    The State University of New York (SUNY), NY 14260-2500, USA;
    Emails: \{kslavaki,shivasal\}@buffalo.edu. Tel:
    +1~(716)~645-1012. \textit{D.~S.~Wack}\/ is with the
    Depts.~of Nuclear Medicine and Biomedical Eng., UB (SUNY);
    Email: dswack@buffalo.edu. \textit{S.~F.~Muldoon}\/ is with
    the Dept.~of Mathematics and Computational and Data-Enabled
    Science and Eng.\ Program, UB (SUNY); Email:
    smuldoon@buffalo.edu. \textit{H.~E.~Baidoo-Williams}\/ is
    with the Dept.\ of Mathematics, UB (SUNY), and the US Army
    Research Laboratory, MD, USA; Email:
    henrybai@buffalo.edu. \textit{J.~M.~Vettel}\/ is with the US
    Army Research Laboratory, MD, USA, the Dept.~of Psychological
    and Brain Sciences, Univ.~of California, Santa Barbara, USA,
    and the Dept.~of Bioengineering, Univ.~of Pennsylvania, USA;
    Email: jean.m.vettel.civ@mail.mil. \textit{M.~Cieslak}\/ and
    \textit{S.~T.~Grafton}\/ are with Dept.~of Psychological and
    Brain Sciences, Univ.~of California, Santa Barbara, USA;
    Emails: mattcieslak@gmail.com,
    scott.grafton@psych.ucsb.edu.}%
  \thanks{Preliminary parts of this study can be found in
    \cite{Slavakis.SSP.16,
      Slavakis.Asilomar.16}. \textit{D.~S.~Wack}\/ receives
    research/grant support from the William~E.~Mabie, DDS, and
    Grace~S.~Mabie Fund. This work is also supported by the NSF
    awards Eager~1343860 and 1514056, and by the Army Research
    Laboratory through contract no.~W911NF-10-2-0022 from the
    U.S.\ Army research office. The content is solely the
    responsibility of the authors and does not necessarily
    represent the official views of the U.S.\ Army funding
    agency.}%
}

\maketitle

\begin{abstract}
  This paper advocates Riemannian multi-manifold modeling in the
  context of \textit{network-wide}\/ non-stationary time-series
  analysis. Time-series data, collected sequentially over time
  and across a network, yield features which are viewed as points
  in or close to a union of multiple submanifolds of a Riemannian
  manifold, and distinguishing disparate time series amounts to
  clustering multiple Riemannian submanifolds. To support the
  claim that exploiting the latent Riemannian geometry behind
  many statistical features of time series is beneficial to
  learning from network data, this paper focuses on brain
  networks and puts forth two feature-generation schemes for
  network-wide dynamic time series. The first is motivated by
  Granger-causality arguments and uses an auto-regressive moving
  average model to map low-rank linear vector subspaces, spanned
  by column vectors of appropriately defined observability
  matrices, to points into the Grassmann manifold.  The second
  utilizes (non-linear) dependencies among network nodes by
  introducing kernel-based partial correlations to generate
  points in the manifold of positive-definite
  matrices. Capitilizing on recently developed research on
  clustering Riemannian submanifolds, an algorithm is provided
  for distinguishing time series based on their geometrical
  properties, revealed within Riemannian feature
  spaces. Extensive numerical tests demonstrate that the proposed
  framework outperforms classical and state-of-the-art techniques
  in clustering brain-network states/structures hidden beneath
  synthetic fMRI time series and brain-activity signals generated
  from real brain-network structural connectivity matrices.
\end{abstract}

\begin{IEEEkeywords}
  Time series, (brain) networks, Riemannian manifold, clustering,
  ARMA model, partial correlations, kernels.
\end{IEEEkeywords}

\section{Introduction}\label{sec:intro}

Recent advances in brain science have highlighted the need to
view the brain as a complex network of interacting nodes across
spatial and temporal scales~\cite{Sporns.book,
  Bullmore.complex.09, Park.structural.13,
  Braun.brain.nets.15}. The emphasis on understanding the brain
as a network has capitalized on concurrent advances in
brain-imaging technology, such as electroencephalography (EEG)
and functional magnetic resonance imaging (fMRI), which assess
brain activity by measuring neuronal time
series~\cite{Braun.brain.nets.15, brain.tutorial.spm.13}.
% FMRI is a 4D neuro-imaging procedure that assesses dynamic brain
% activity by measuring blood-oxygen-level dependent (BOLD)
% neuronal time series~\cite{Ogawa.brain.90}.

Clustering is the unsupervised (no data labels available)
learning process of grouping data patterns into clusters based on
similarity~\cite{Theodoridis.pattern.09}. Time-series clustering
has emerged as a prominent tool in big-data analytics because not
only does it enable compression of high-dimensional and voluminous
data, \eg, one hour of electrocardiogram data occupies 1Gb of
storage~\cite{Aghabozorgi.time.series.clustering.15}, but it also
leads to discovery of patterns hidden beneath network-wide
time-series datasets. Indeed, data-mining and comparison of
functional connectivity patterns of the default-mode
  brain network of human subjects, \ie, brain regions that
remain active during \textit{resting-state}\/ periods in fMRI,
has enhanced understanding of brain disorders such as the
Alzheimer disease and autism~\cite{Buckner.unrest.07,
  Stam.Alzheimer.09, Greicius.default.04, Rombouts.altered.05},
depression~\cite{Greicius.depression.07}, anxiety, epilepsy and
schizophrenia~\cite{Broyd.default.09}.

\begin{figure}[!t]
  \centering
  \subfloat[States]
  {\includegraphics[width=.9\linewidth]{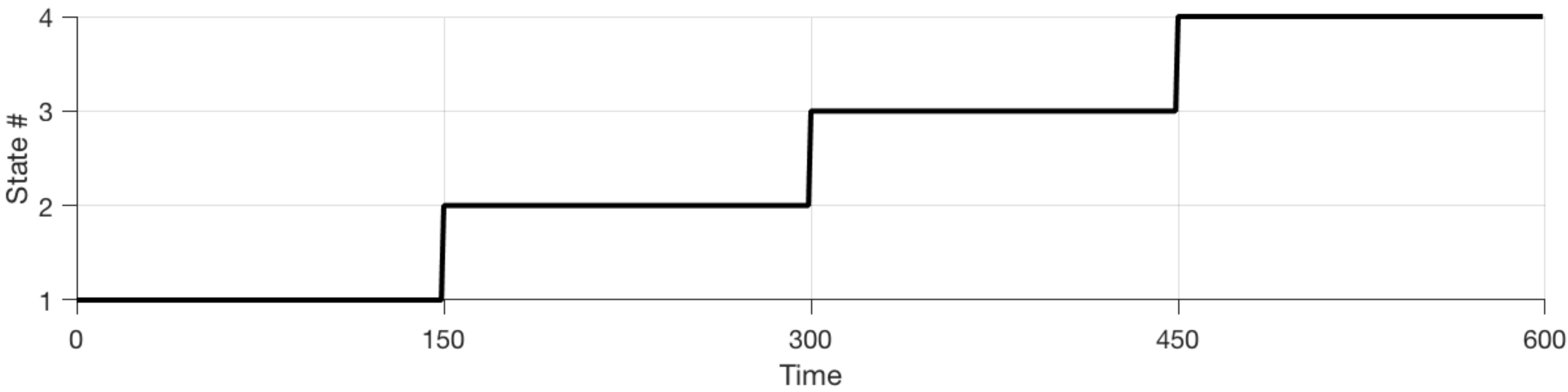}
    \label{fig:motivate.states}}\\
    \subfloat[]
  {\includegraphics[width=.15\linewidth]{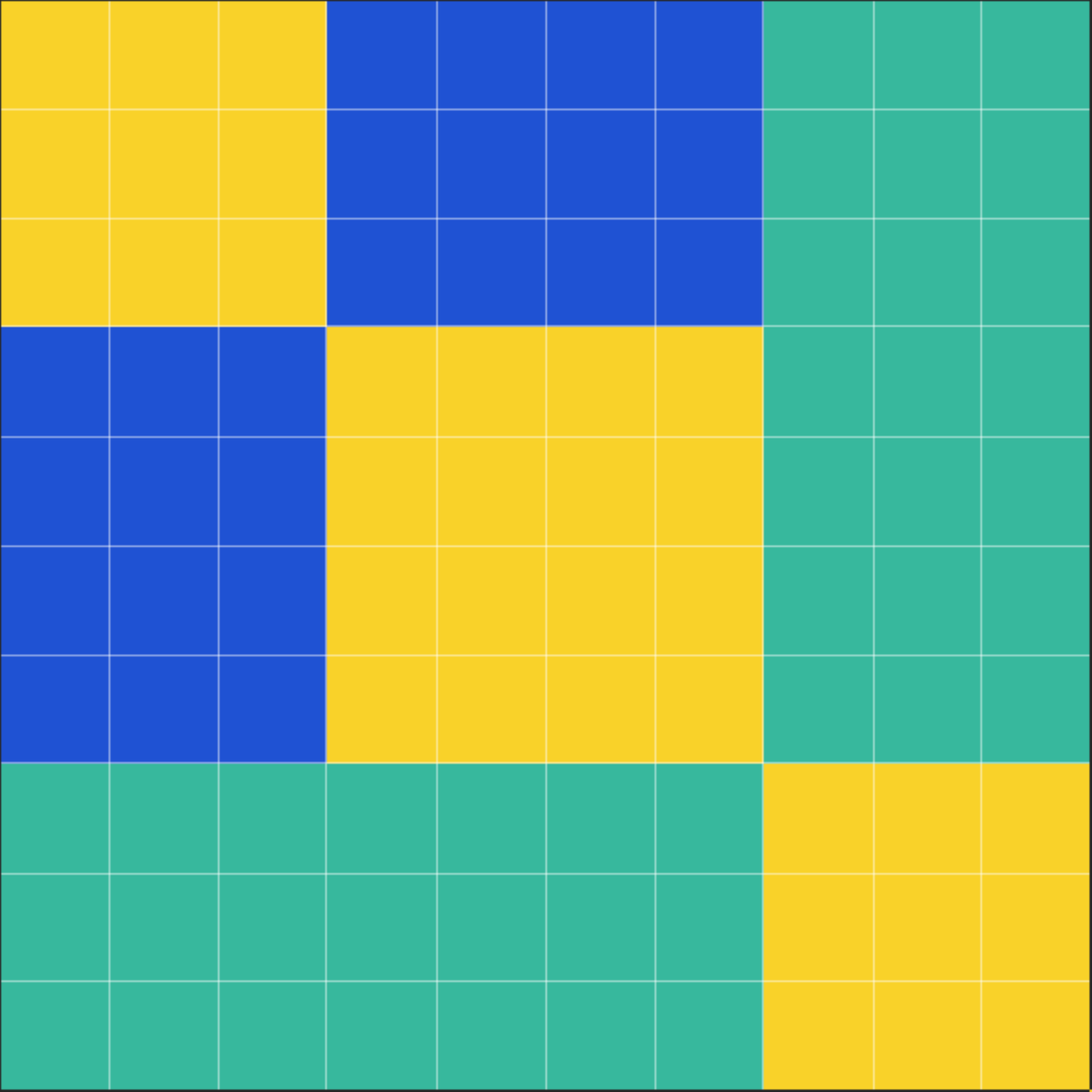}
    \label{fig:motivate.Conn.1}}\hfill
  \subfloat[]
  {\includegraphics[width=.15\linewidth]{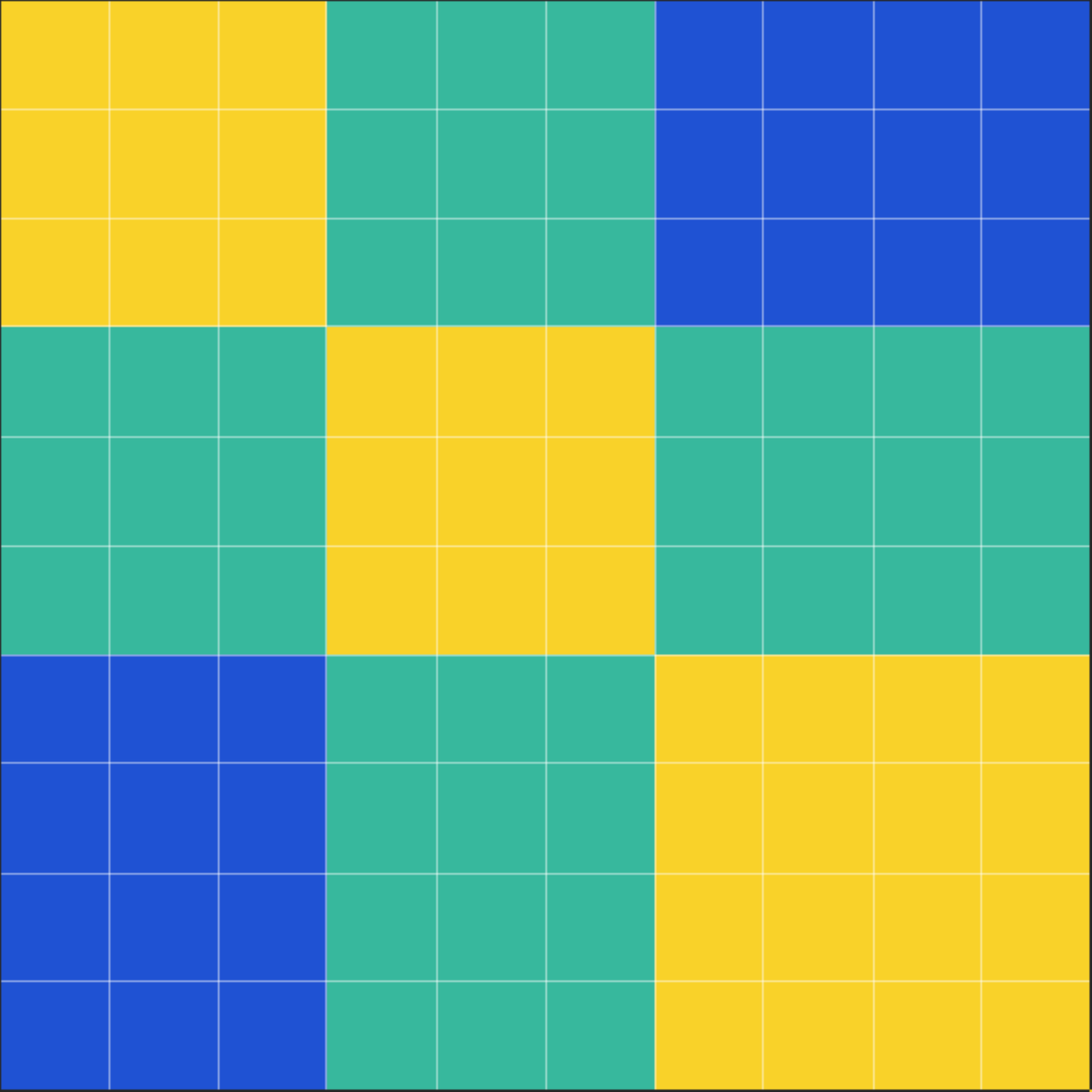}
    \label{fig:motivate.Conn.2}}\hfill
  \subfloat[]
  {\includegraphics[width=.15\linewidth]{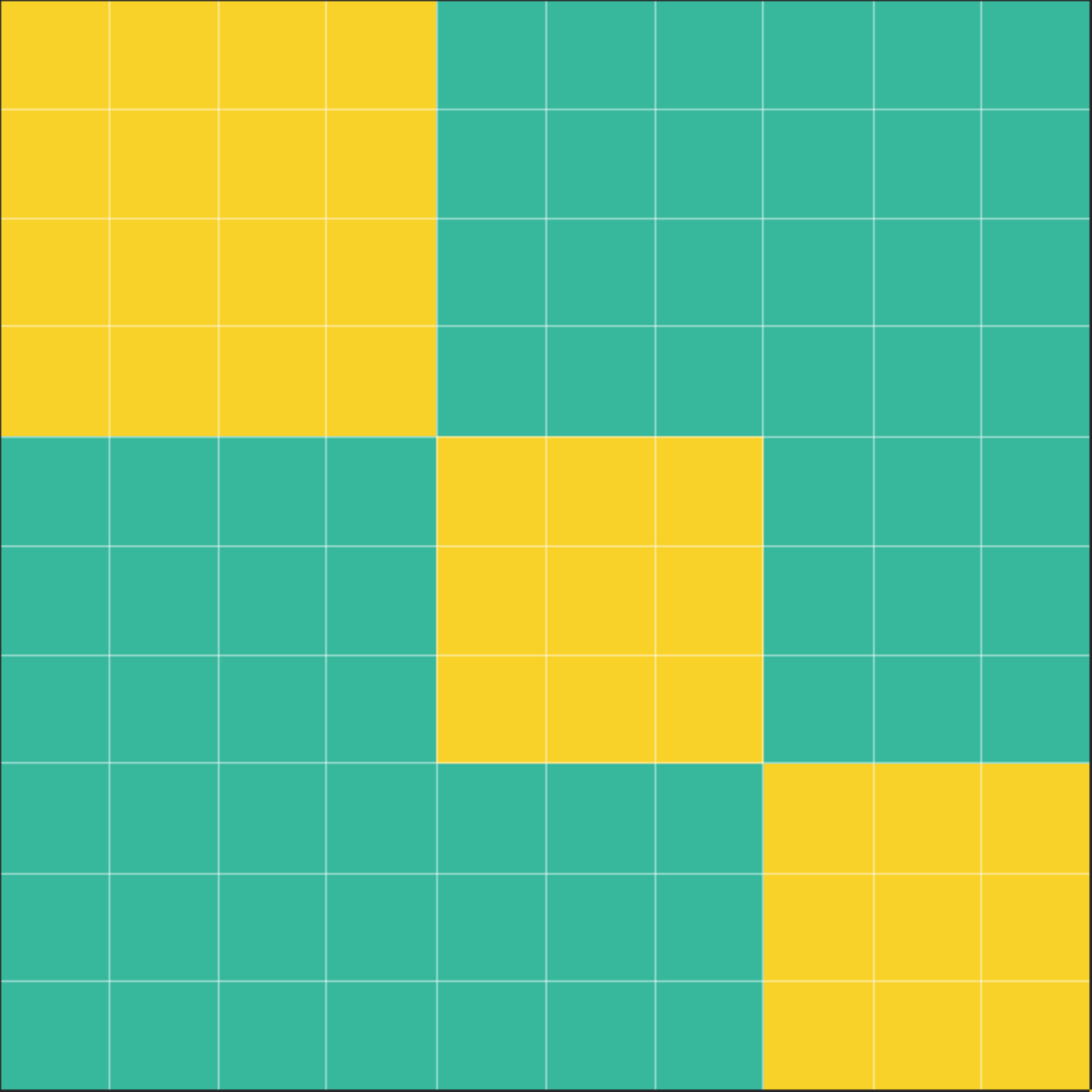}
    \label{fig:motivate.Conn.3}}\hfill
  \subfloat[]
  {\includegraphics[width=.15\linewidth]{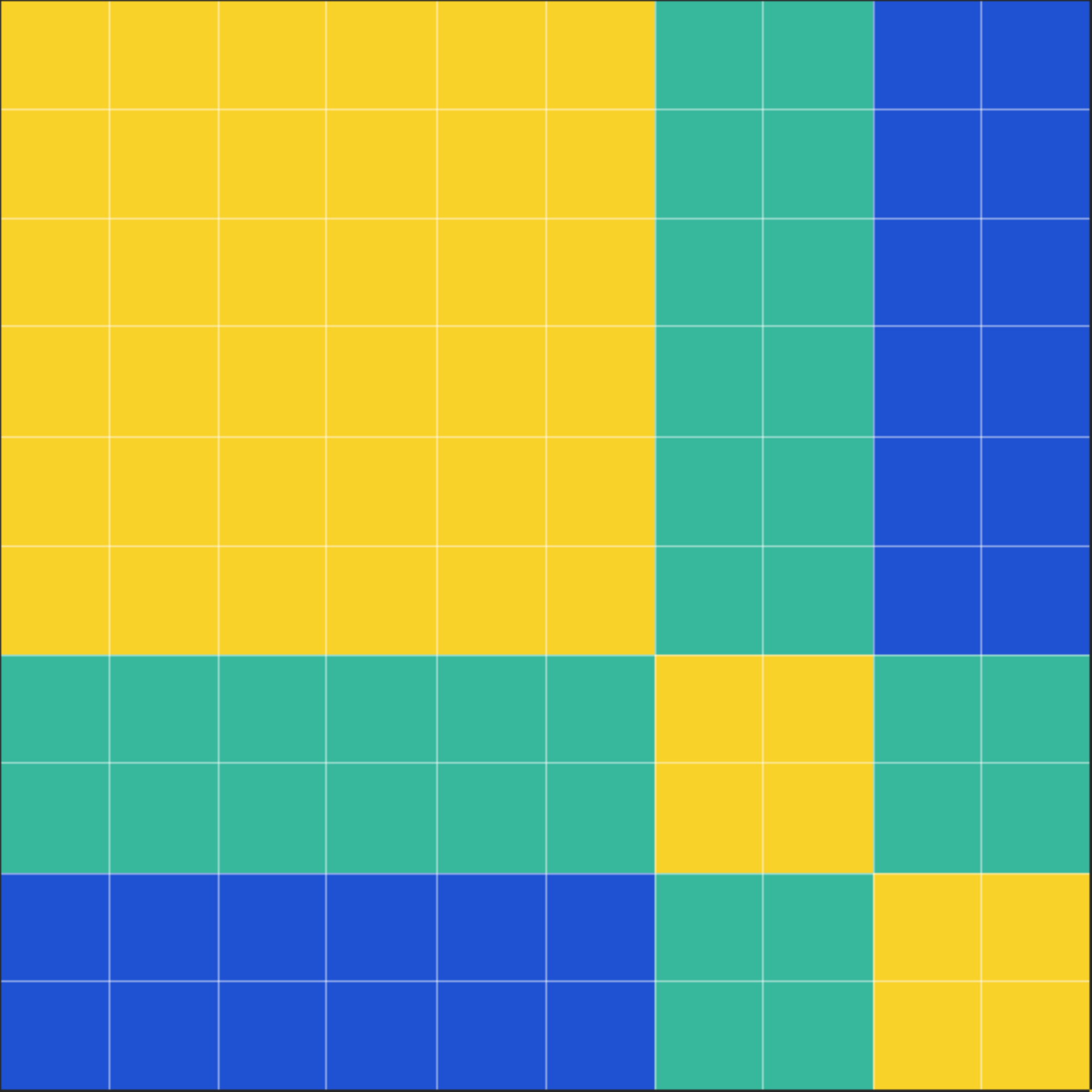}
    \label{fig:motivate.Conn.4}}\\
  \subfloat[Single-node BOLD time series]
  {\includegraphics[width=.9\linewidth]{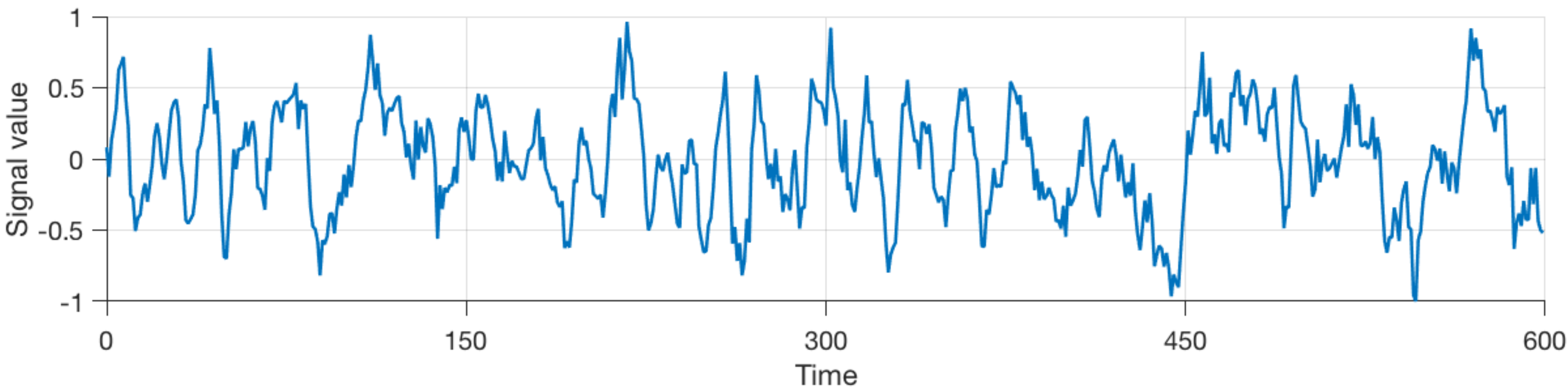}
    \label{fig:motivate.BOLD}}\\
  \subfloat[]
  {\includegraphics[width=.15\linewidth]{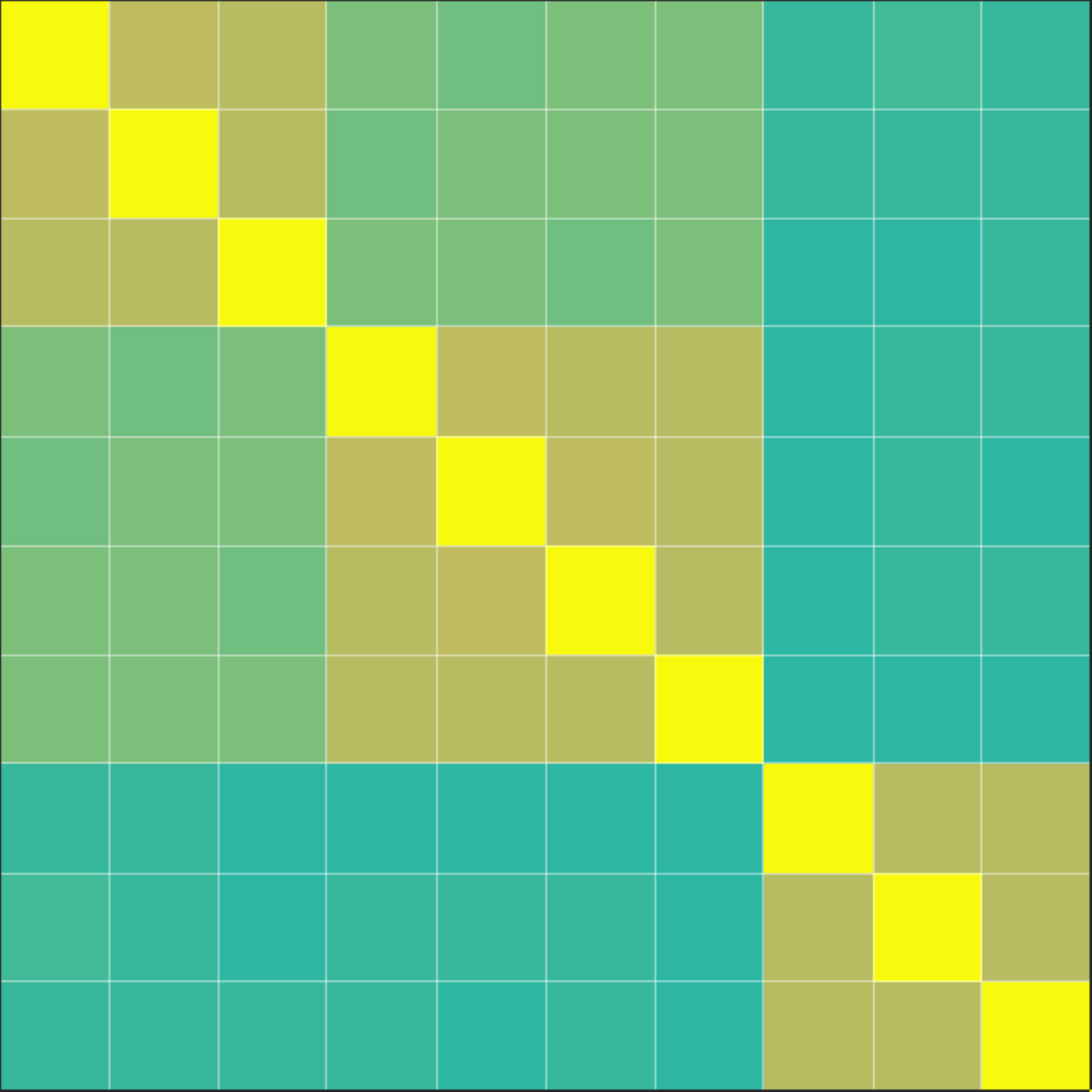}
    \label{fig:motivate.Cov.1}}\hfill
  \subfloat[]
  {\includegraphics[width=.15\linewidth]{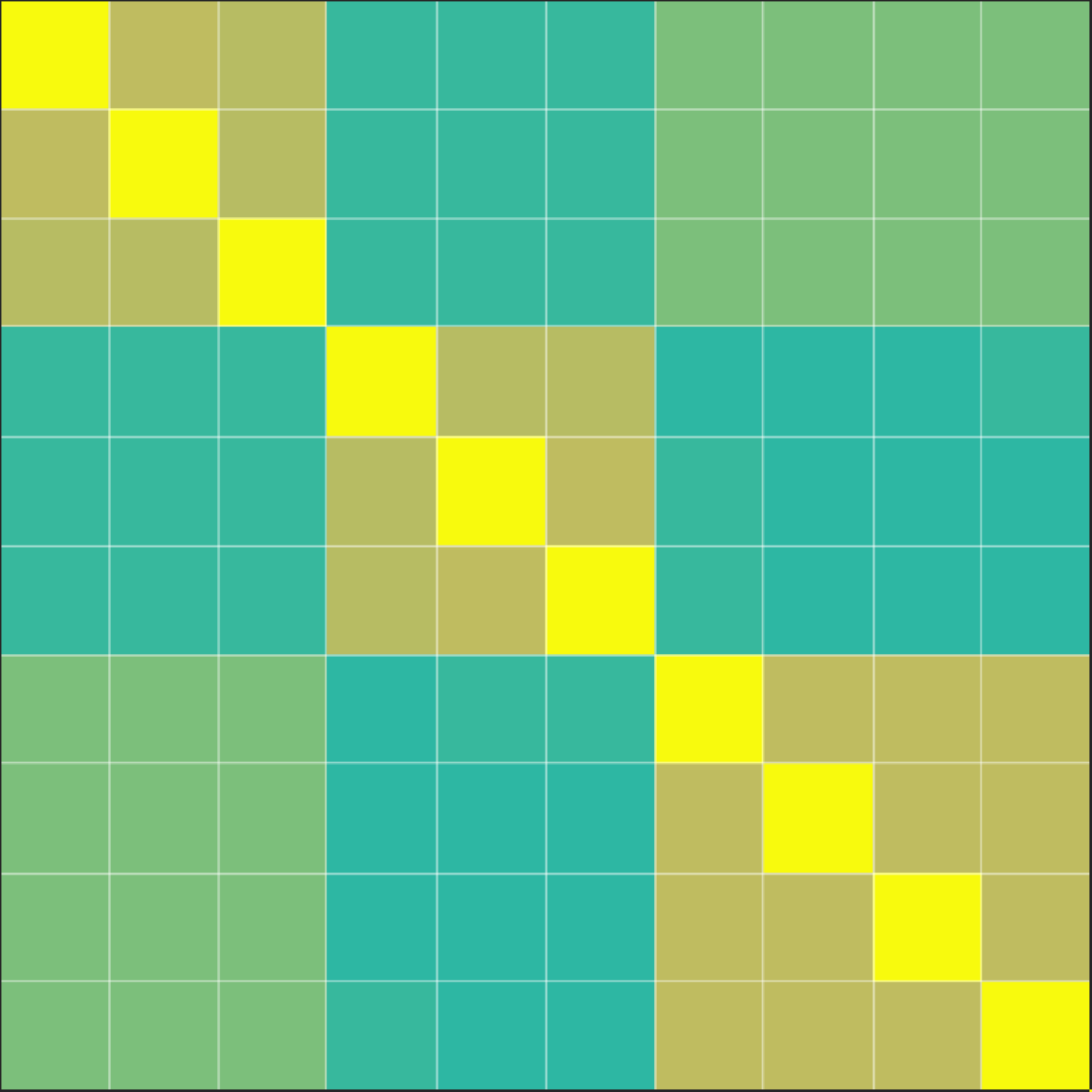}
    \label{fig:motivate.Cov.2}}\hfill
  \subfloat[]
  {\includegraphics[width=.15\linewidth]{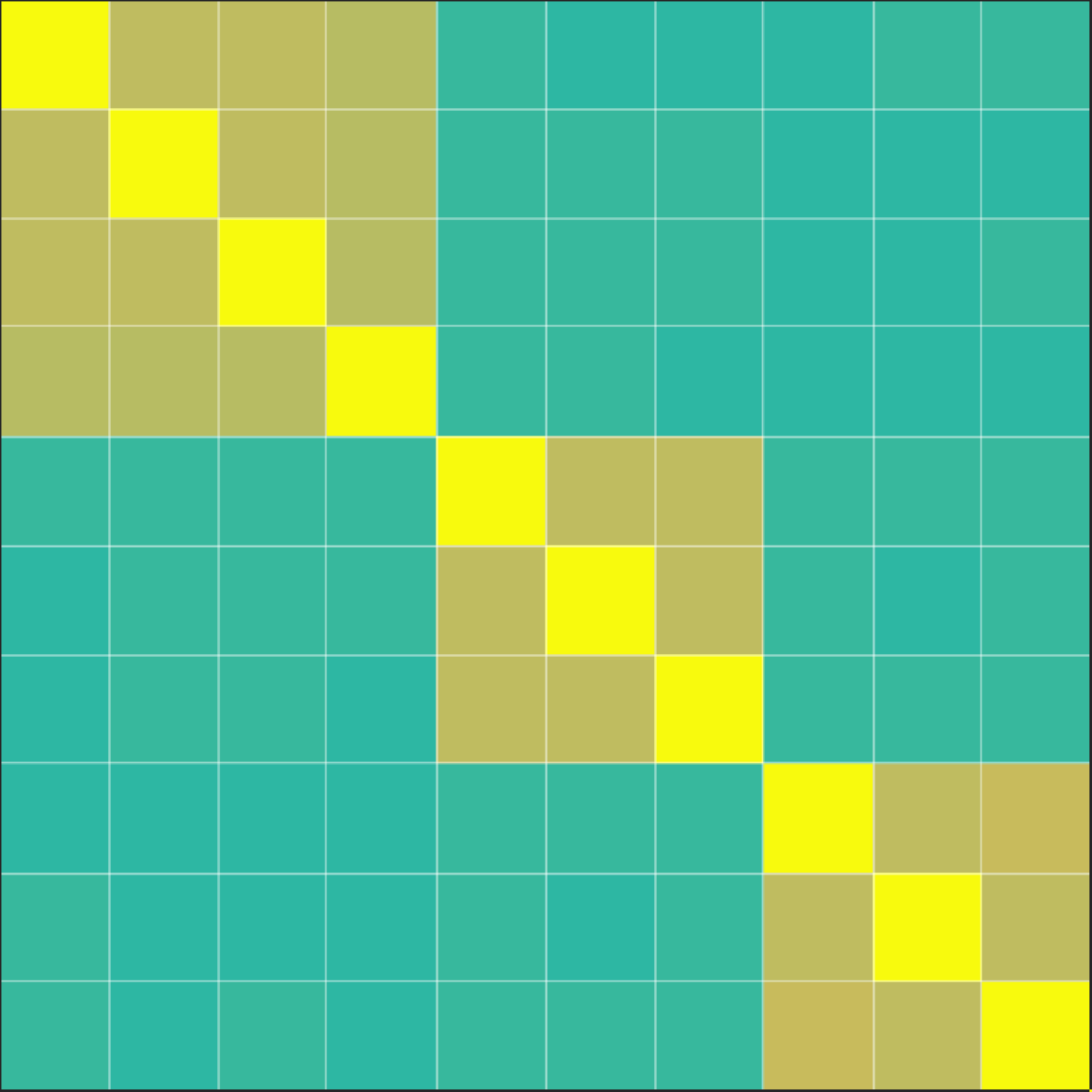}
    \label{fig:motivate.Cov.3}}\hfill
  \subfloat[]
  {\includegraphics[width=.15\linewidth]{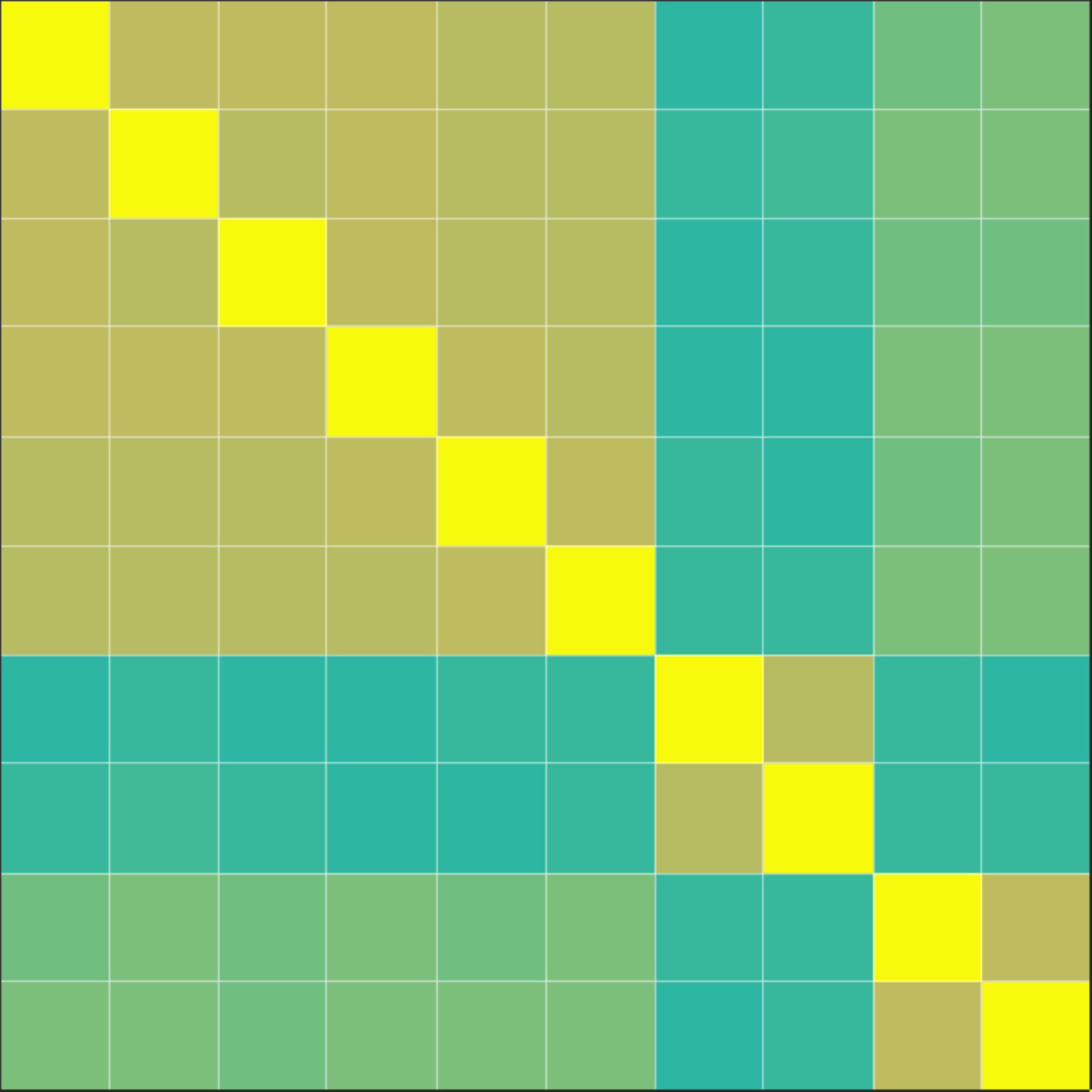}
    \label{fig:motivate.Cov.4}}\\
  \subfloat[]
  {\includegraphics[width=.15\linewidth]{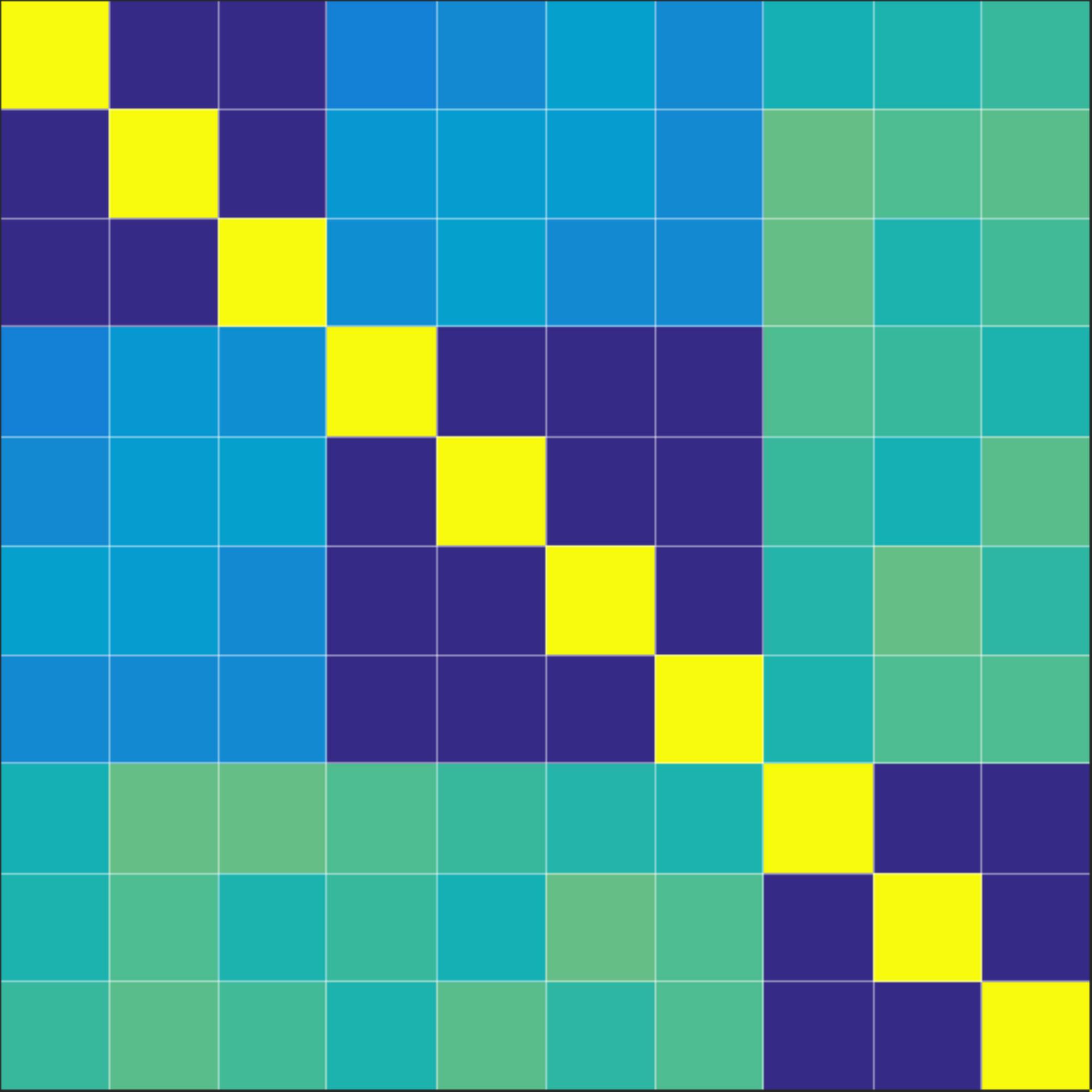}
    \label{fig:motivate.PC.1}}\hfill
  \subfloat[]
  {\includegraphics[width=.15\linewidth]{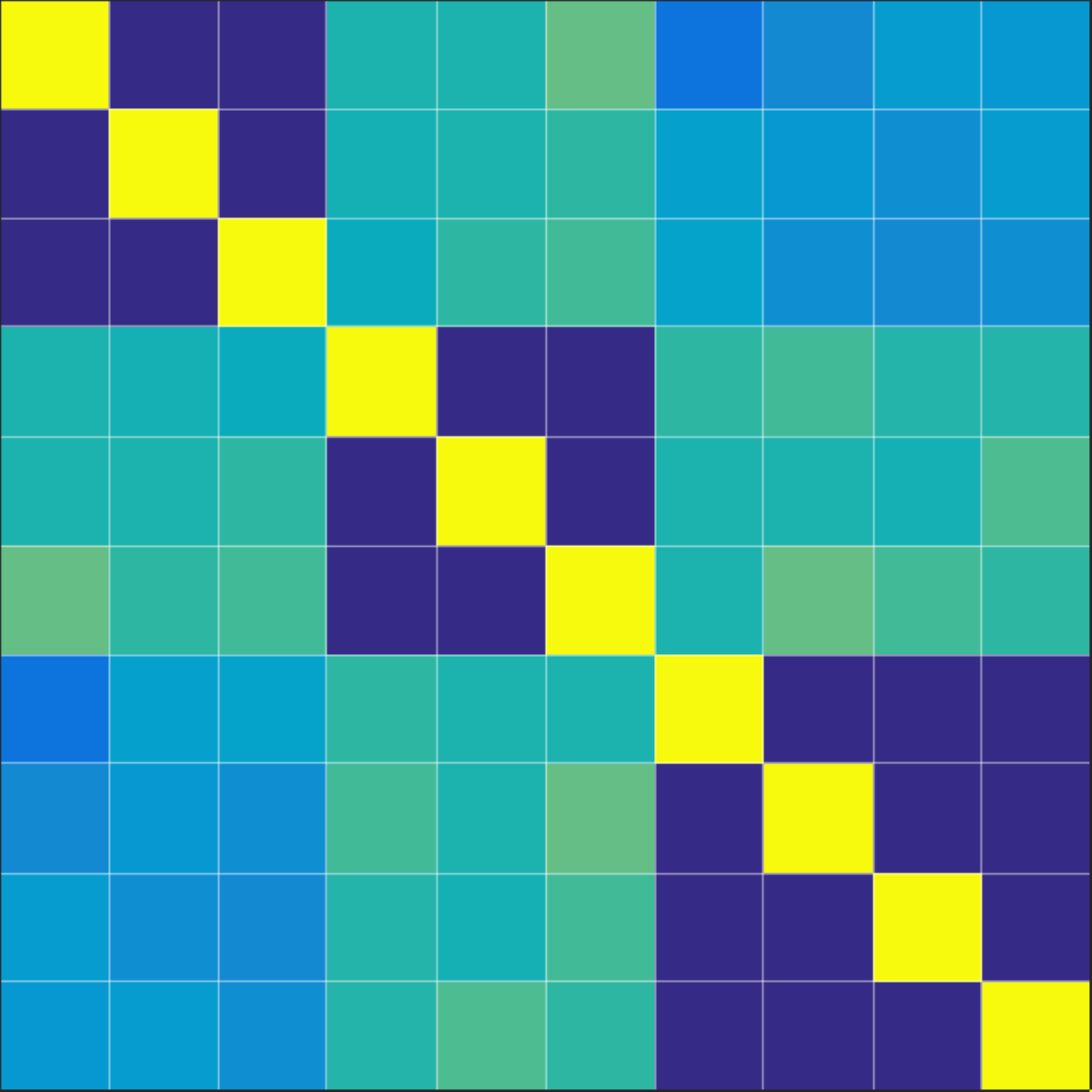}
    \label{fig:motivate.PC.2}}\hfill
  \subfloat[]
  {\includegraphics[width=.15\linewidth]{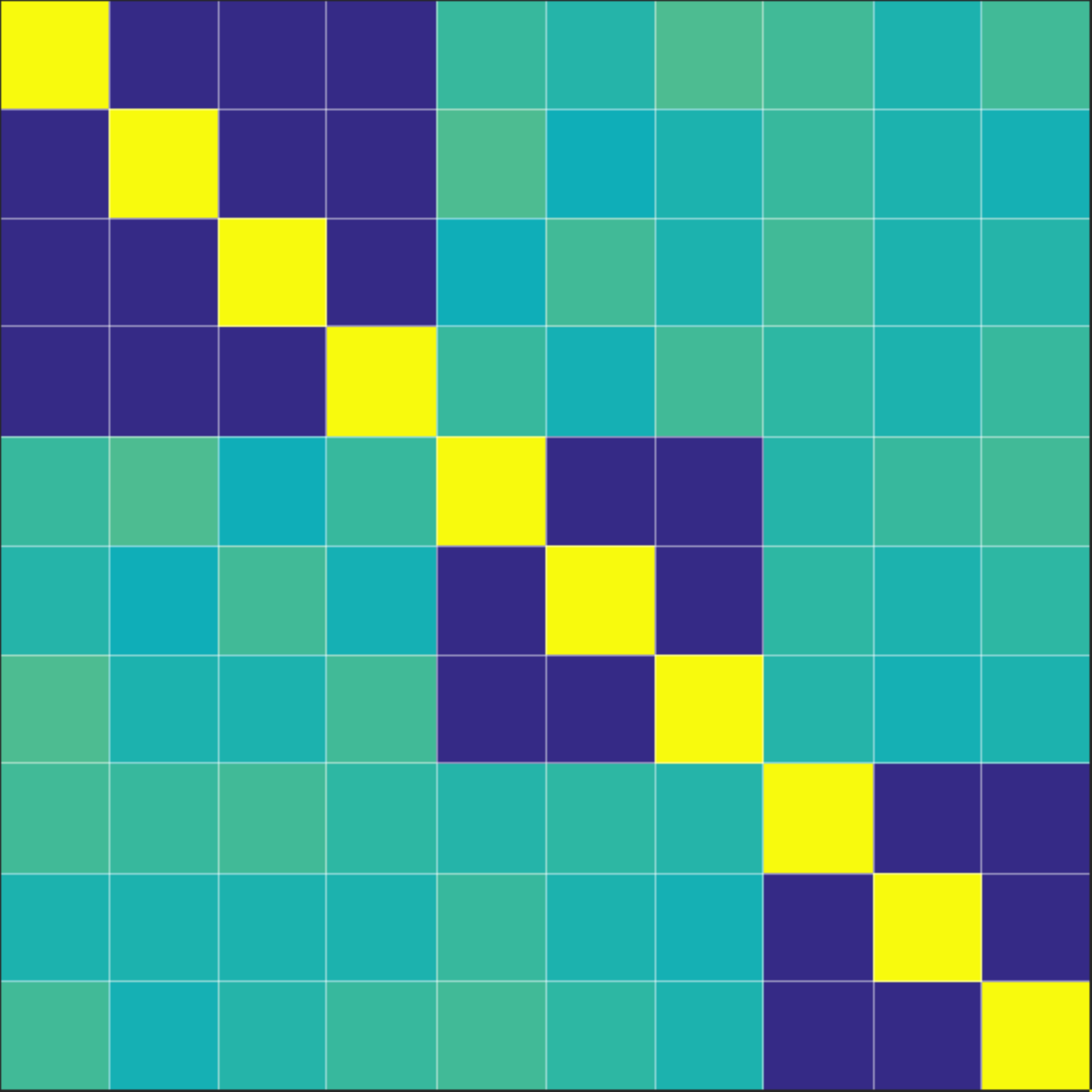}
    \label{fig:motivate.PC.3}}\hfill
  \subfloat[]
  {\includegraphics[width=.15\linewidth]{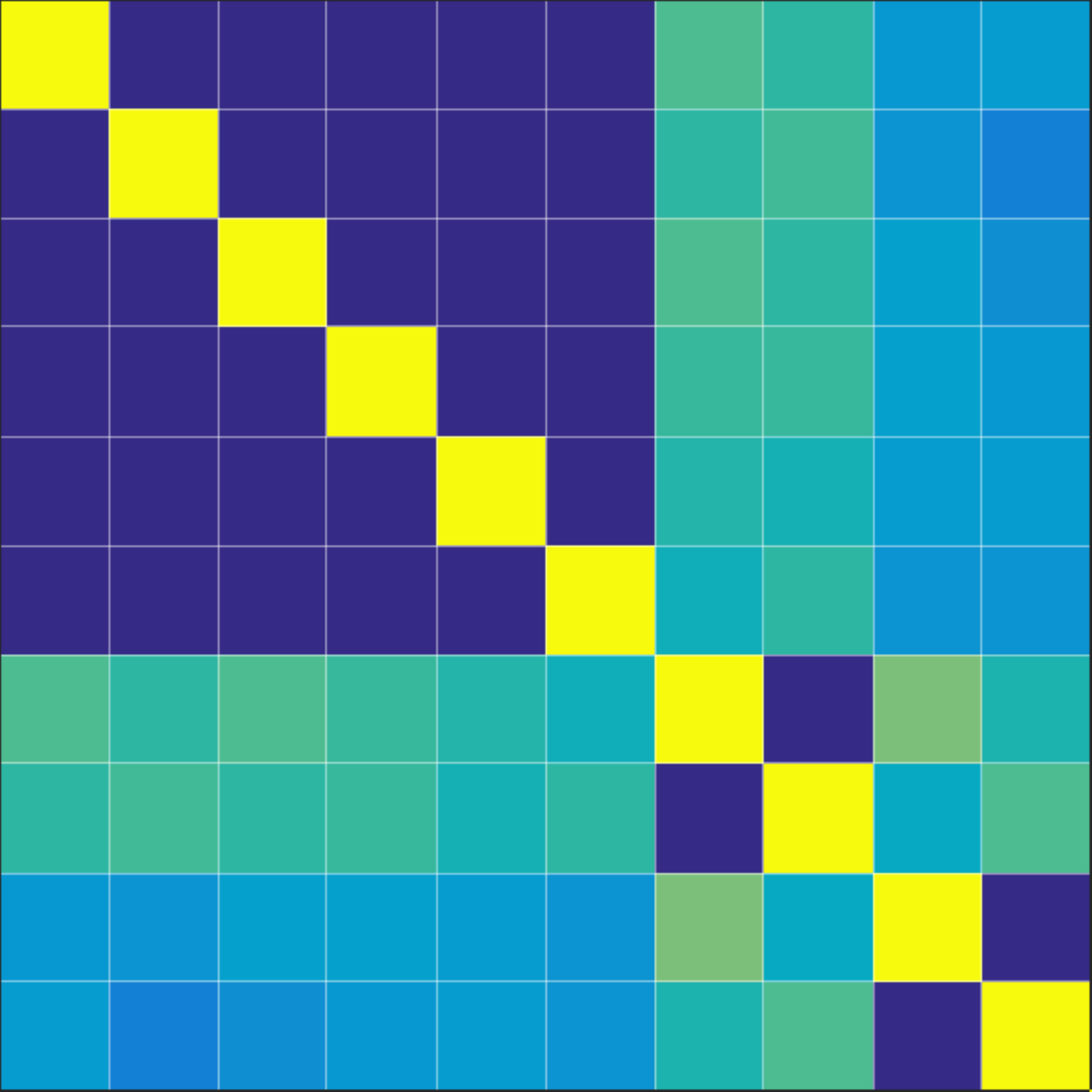}
    \label{fig:motivate.PC.4}}\\
  \subfloat[]
  {\includegraphics[width=.15\linewidth]{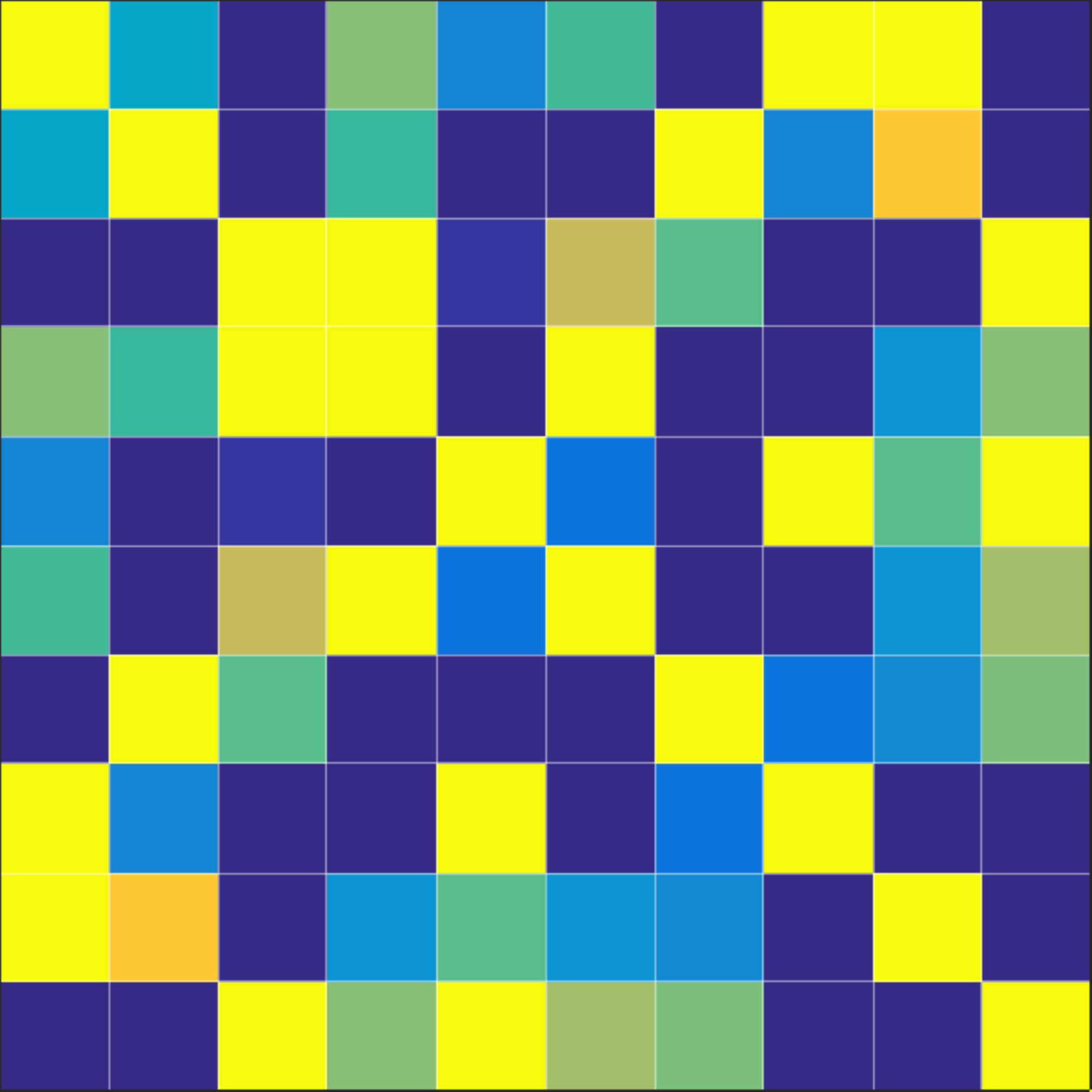}
    \label{fig:motivate.PC.single.1}}\hfill
  \subfloat[]
  {\includegraphics[width=.15\linewidth]{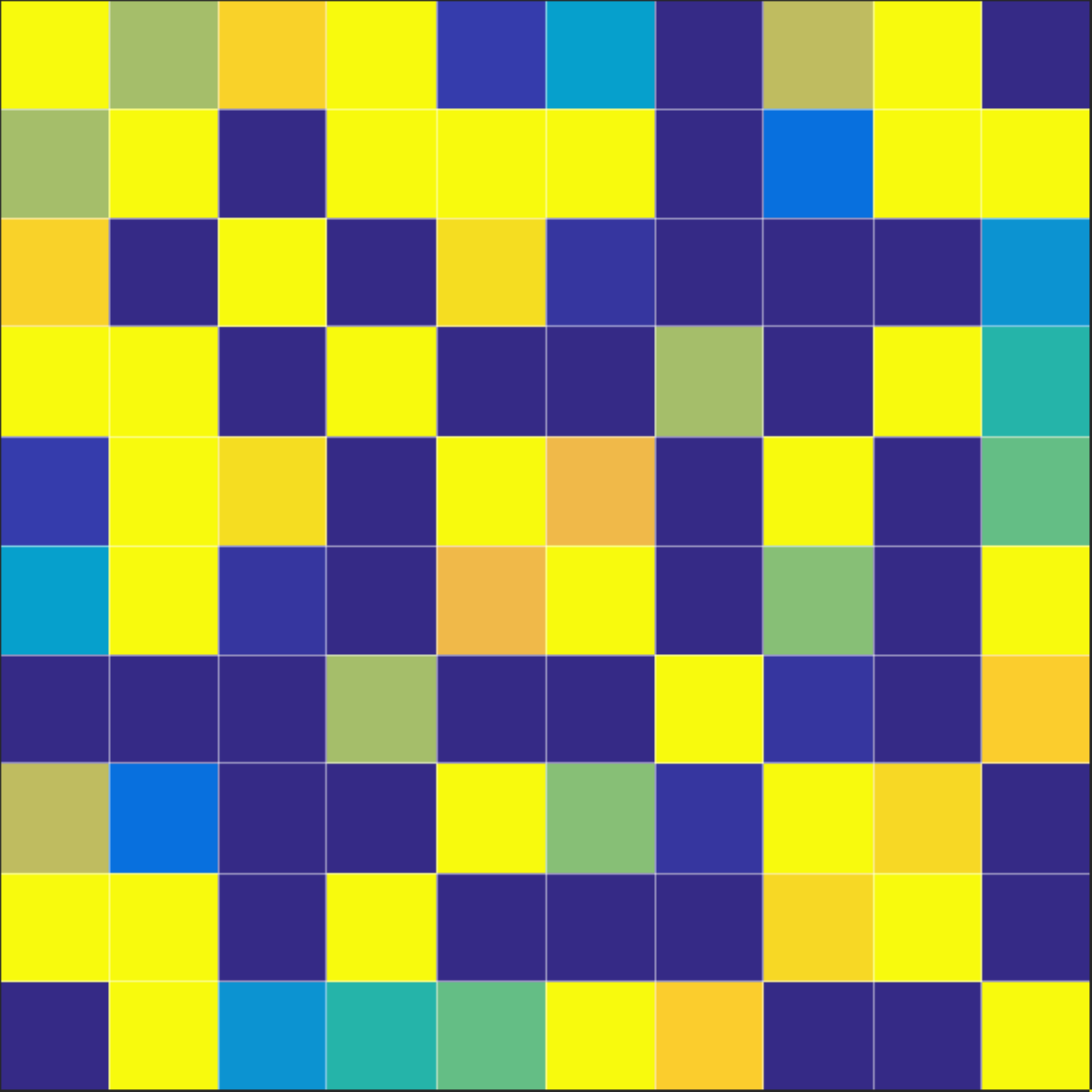}
    \label{fig:motivate.PC.single.2}}\hfill
  \subfloat[]
  {\includegraphics[width=.15\linewidth]{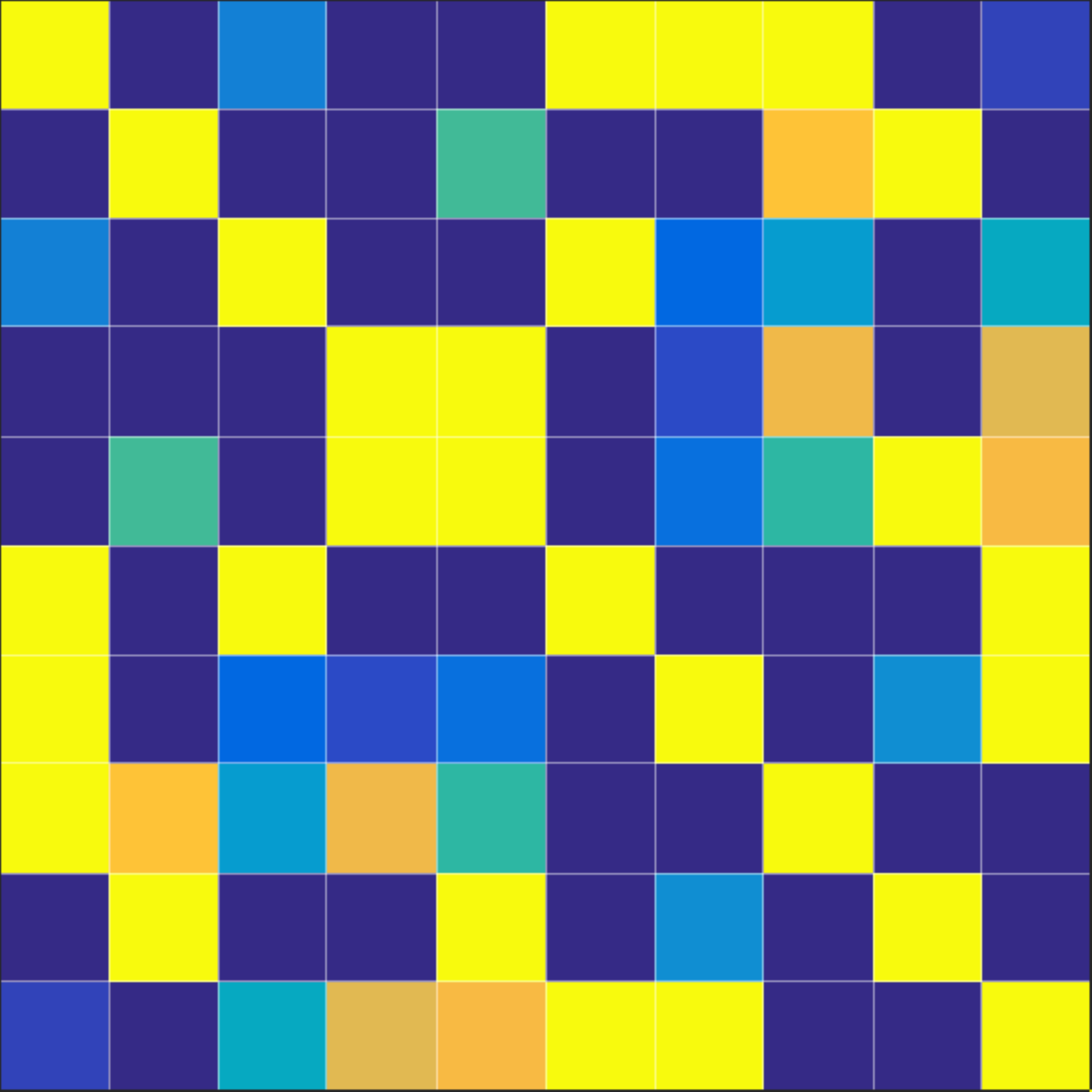}
    \label{fig:motivate.PC.single.3}}\hfill
  \subfloat[]
  {\includegraphics[width=.15\linewidth]{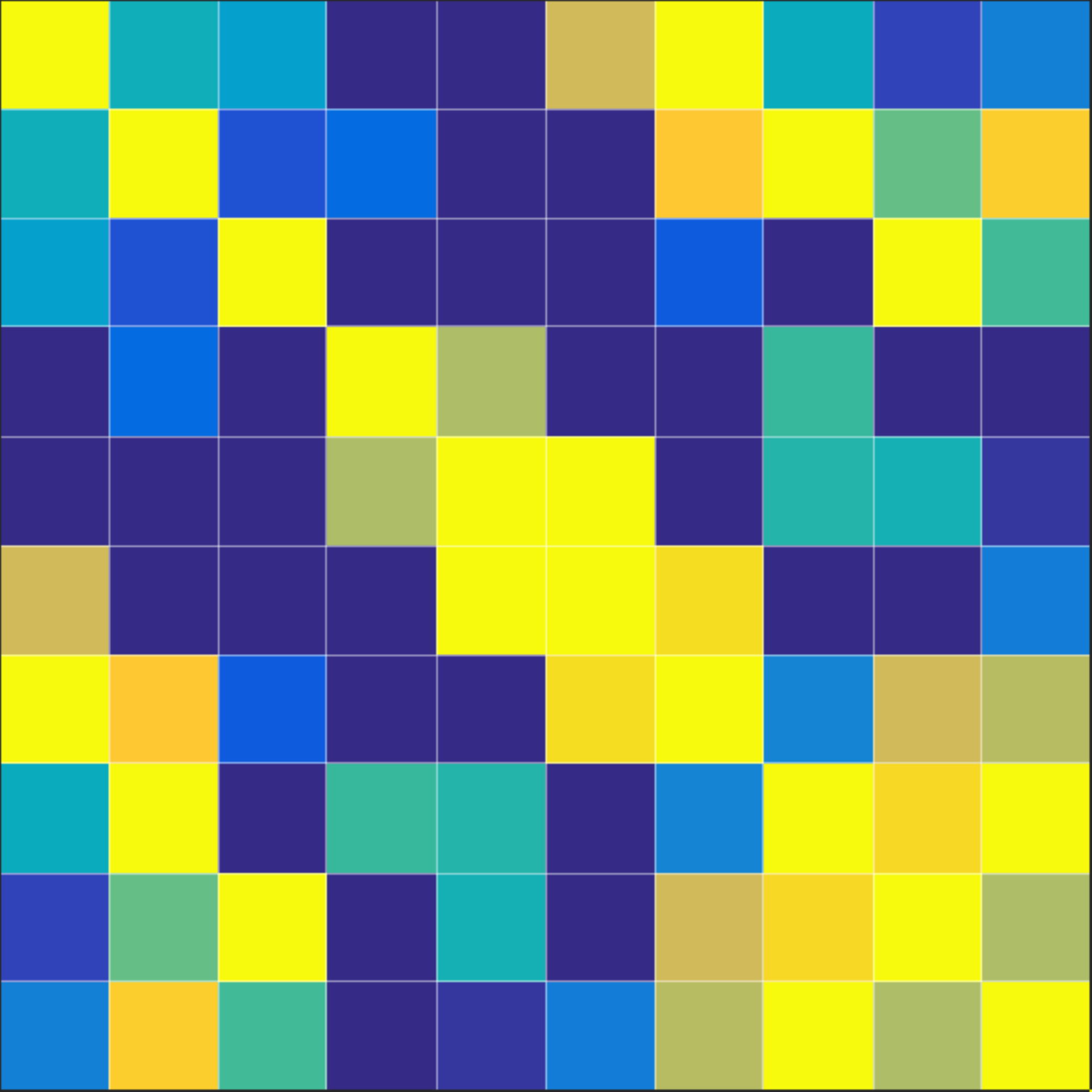}
    \label{fig:motivate.PC.single.4}}
  \caption{A motivating example based on synthetically generated
    data via the SimTB MATLAB toolbox~\protect\cite{simtb,
      Allen.cortex.14}. Fig.~\ref{fig:motivate.states} shows the
    time profile of four brain-network resting states (ten
    nodes). For each state, the functional connectivity pattern
    stays fixed
    (Figs.~\ref{fig:motivate.Conn.1}--\ref{fig:motivate.Conn.4}).
    Fig.~\ref{fig:motivate.BOLD} demonstrates a single
    realization of a single-node BOLD time series. Average
    covariance
    (Figs.~\ref{fig:motivate.Cov.1}--\ref{fig:motivate.Cov.4})
    and partial-correlation (PC)
    (Figs.~\ref{fig:motivate.PC.1}--\ref{fig:motivate.PC.4})
    matrices are obtained by sample averaging $100$ realizations
    of the covariance and PC matrices, computed from the BOLD
    time series whose length equals the time span of a network
    state. No sample averaging is considered in the computation
    of the PC matrices of
    Figs.~\ref{fig:motivate.PC.single.1}--\ref{fig:motivate.PC.single.4}.}
  \label{fig:motivate}
\end{figure}

To motivate the following discussion, consider the ten-node
resting-state brain-network (RSBN) toy example of
Fig.~\ref{fig:motivate}, with four distinct network
states/structures whose evolution over time is shown in
Fig.~\ref{fig:motivate.states}. Those states are associated with
the four functional connectivity matrices of
Figs.~\ref{fig:motivate.Conn.1}--\ref{fig:motivate.Conn.4}: nodes
of the same color are considered to be connected, while no
connection is established among nodes with different colors. For
each state, connectivity matrices stay fixed. Based on the
previous connectivity matrices, blood-oxygen-level dependent
(BOLD) time series~\cite{Ogawa.brain.90}, \eg,
Fig.~\ref{fig:motivate.BOLD}, are simulated via the SimTB MATLAB
toolbox~\cite{simtb, Allen.cortex.14}, under a generation
mechanism detailed in Sec.~\ref{sec:synthetic.data}. Examples of
features extracted from the BOLD time series are the covariance
(Figs.~\ref{fig:motivate.Cov.1}--\ref{fig:motivate.Cov.4}) and
partial-correlation matrices
(Figs.~\ref{fig:motivate.PC.1}--\ref{fig:motivate.PC.4}),
computed via correlations of the time series whose time spans are
set equal to the time span of a single state; see
Sec.~\ref{sec:PC} for a detailed description. For patterns to
emerge, Figs.~\ref{fig:motivate.Cov.1}--\ref{fig:motivate.PC.4}
suggest that sample averaging of features over many time-series
realizations is needed. On the contrary,
Figs.~\ref{fig:motivate.PC.single.1}--\ref{fig:motivate.PC.single.4}
demonstrate that partial-correlation matrices, obtained without
any sample averaging, do not offer much help in identifying the
latent connectivity structure. Since \textit{multiple}\/
realizations of BOLD time series are hard to find in practice,
rather than associating a single feature with a network state
(Figs.~\ref{fig:motivate.PC.single.1}--\ref{fig:motivate.PC.single.4}),
it would be preferable to extract a \textit{sequence}\/ of
features $(x_t)_t$ ($t$ denotes discrete time), \eg, running
averages of covariance matrices, to characterize a network
state. This is also in accordance with recent evidence showing
that brain-network resting states demonstrate dynamic attributes,
\eg, \cite{Buckner.unrest.07}. Indeed, the usual presupposition
that functional connectivity is static over relatively large
period of times has been challenged in works focusing on
time-varying connectivity patterns~\cite{Sakoglu.10,
  Allen.cortex.14, Zalesky.14, Molenaar.16, Britz.EEG.RSN.10},
shifting the fMRI/EEG paradigm to the so-called ``chronnectome''
setting, where coupling within the brain network is dynamic, and
two or more brain regions or sets of regions, all possibly
evolving in time, are coupled with connective strengths that are
also themselves explicit functions of
time~\cite{chronnectome}. Such an approach has been already
utilized to show that sleep states can be predicted via
connectivity patterns at given times~\cite{Tagliazucchi.14}, and
that schizophrenia can be correctly
identified~\cite{Damaraju.14}.

The previous discussion brings forth the following pressing
questions:
\begin{enumerate*}[label=\textbf{(\roman*)}]

\item Are there features that carve the latent network
  state/structure out of the observed network-wide time series?
  Is it possible to extract a \textit{sequence}\/ of features
  from a time series to capture a possibly dynamically evolving
  network state, as Fig.~\ref{fig:motivate} and the related
  discussion suggest?

\item Is there any model that injects geometrical arguments in
  the feature space, and is there any way to exploit that
  geometry to design a learning (in particular clustering)
  algorithm which provides state-of-the-art performance?
  
\end{enumerate*}

\subsection{Contributions of this work}\label{sec:contributions}

This paper provides answers to the previous questions. Although
the advocated methods, together with the underlying theory, apply
to any network-wide time series, this paper focuses on
brain-networks. Time-series data are processed sequentially via a
finite-size sliding window that moves along the time axis to
extract features which monitor the possibly time-varying
state/structure of the network (Fig.~\ref{fig:Flowchart};
Secs.~\ref{sec:ARMA} and \ref{sec:PC}). Two feature-extraction
schemes, novel in exploiting latent Riemannian geometry within
network-wide time series, are introduced.

First, motivated by Granger-causality arguments, which play a
prominent role in time-series analysis~\cite{Granger.69,
  Chen.frequency.Granger.06, Geweke.Granger.84,
  Boudjellaba.ARMA.Granger.92}, an auto-regressive moving average
model is proposed to extract low-rank linear vector subspaces
from the columns of appropriately defined observability
matrices. Such linear subspaces demonstrate a remarkable
geometrical property: they are points of the Grassmannian, a
well-known Riemannian manifold (Sec.~\ref{sec:ARMA}).

Second, Sec.~\ref{sec:PC} generalizes the popular
network-analytic tool of ``linear'' partial correlations
(PCs)~\cite{Kolaczyk} to ``non-linear'' PCs, via reproducing
kernel functions (\cf~Appendix~\ref{app:RKHS}), to capture the
likely non-linear dependencies among network nodes, \eg,
\cite{Karanikolas.icassp.16}. Geometry is also prominent in
Sec.~\ref{sec:PC}: Prop.~\ref{prop:kPC} demonstrates that
matrices generated by kernel-based PCs are points of the
celebrated Riemannian manifold of positive-definite
matrices.

Capitalizing on the Riemannian-geometry thread that binds the
previous feature-extraction schemes, learning, in particular
clustering, is performed in a Riemannian manifold
$\mathpzc{M}$. The key hypothesis, adopted from the very
recent~\cite{MMM.arxiv.14, MMM.15}, is the \textit{Riemannian
  multi-manifold modeling (RMMM)}\/ assumption: each cluster
constitutes a submanifold of $\mathpzc{M}$, and distinguishing
disparate time series amounts to clustering multiple Riemannian
submanifolds; \cf~Figs.~\ref{fig:Flowchart} and
\ref{fig:RMMM}. This is in contrast with the prevailing
perception of clusters in literature as ``well-concentrated''
data clouds, whose convex hulls can be (approximately) separated
by hyperplanes in the feature space, a hypothesis which lies also
beneath the success of Kmeans and
variants~\cite{Theodoridis.pattern.09}. In contrast, RMMM, as
well as the advocated clustering algorithm of Sec.~\ref{sec:GCT},
\textit{allow}\/ for clusters (submanifolds) to intersect. The
extensive numerical tests of Sec.~\ref{sec:tests} demonstrate
that the proposed framework outperforms classical and
state-of-the-art techniques in clustering brain-network
states/structures.

\subsection{Prior art}\label{sec:prior.art}

Although the majority of methods on time-series clustering
follows the ``shape-based'' approach, where clustering is applied
to raw time-series
data~\cite{Aghabozorgi.time.series.clustering.15}, fewer studies
have focused on model/feature-based approaches, such as the
present
one~\cite[Table~4]{Aghabozorgi.time.series.clustering.15}.
Study~\cite{Kalpakis.ARIMA.01} fits an auto-regressive integrated
moving average (ARIMA) model to \textit{non}-network-wide
time-series data, measures dissimilarities of patterns via the
(Euclidean) $\ell_2$-distance of cepstrum coefficients, and
applies the Kmedoids algorithm to cluster cepstum-coefficient
patterns. In~\cite{Golay.fuzzy.98}, fuzzy Cmeans is applied to
vectors comprising the Pearson's correlation coefficients of fMRI
time series, under the $\ell_2$- and a hyperbolic-distance
metric. In~\cite{Ou.HMM.15}, hierarchical clustering is applied
to functional connectivity matrices, comprising Pearson's
correlation coefficients of BOLD time series via the
$\ell_2$-distance. Once again, the $\ell_2$-distance is used
in~\cite{Leonardi.disentangling.14}, together with Kmeans and its
sparsity-cognizant K-SVD variant, in clustering functional
connectivity matrices which are formed by Pearson's correlation
coefficients, as well as low-rank matrices obtained via PCA. In
\cite{Allen.cortex.14}, Kmeans is applied to windowed correlation
matrices, under both the $\ell_1$- and $\ell_2$-distances. Kmeans
is also used in clustering brain electrical activity into
microstates in \cite{Marqui.microstates.95}.

In all of the previous cases, Kmeans and variants are predicated
on the assumption that a ``cluster center'' represents well the
``spread'' or variability of the data-cloud associated with each
cluster. Moreover, any underlying feature-space Riemannian
geometry is not exploited. This is in contrast with the RMMM
hypothesis, advocated by this paper, where clusters are modeled
as Riemannian submanifolds, allowed to intersect and to have a
``spread'' which cannot be captured by a single cluster-center
point. To highlight such a difference, Kmeans under the standard
$\ell_2$-distance will be employed in all tests in
Sec.~\ref{sec:tests}. An application of the Riemannian
(Grassmann) distance between low-rank matrices to detect
network-state transitions in fMRI time series can be found
in~\cite{Aviyente.tensor.16}. However, Grassmmanian geometry is
exploited only up to the use of the distance metric
in~\cite{Aviyente.tensor.16}, without taking advantage of the
rich first-order (tangential) information of submanifolds, as the
current study offers in Sec.~\ref{sec:GCT}. Another line of
fruitful research focuses on detecting communities within brain
networks (\eg, \cite{Aviyente.hierarchical.15}) by utilizing
powerful concepts drawn from network/graph theory, such as
modularity~\cite{Newman.modularity.06}. Due to lack of space,
such a community-detection route is not pursued in this paper,
and the related discussion is deferred to a future publication.

Regarding manifold clustering, most of the algorithms stem from
schemes developed originally for Euclidean spaces. An extension
of Kmeans to Grassmannians, with an application to non-negative
matrix factorization, was presented in
\cite{And06grassmannclustering}. The mean-shift algorithm was
also generalized to analytic manifolds in
\cite{subbarao.meer.cvpr.06, centingul_vidal09}. Geodesic
distances of product manifolds were utilized for clustering human
expressions, gestures, and actions in video sequences in
\cite{5771473}. Moreover, spectral clustering and nonlinear
dimensionality reduction techniques were extended to Riemannian
manifolds in \cite{GOH_VIDAL08}. Such schemes are quite
successful when the convex hulls of clusters are well-separated;
however, they often fail when clusters intersect or are closely
located. Clustering data-sets which demonstrate low-dimensional
structure is recently accommodated by unions of affine subspaces
or submanifold models. Submanifolds are usually restricted to
manifolds embedded in either a Euclidean space or the
sphere. Unions of affine subspace models, a.k.a.\ hybrid linear
modeling (HLM) or subspace clustering, have been recently
attracting growing interest, e.g.,
\cite{SubspaceClustering_Vidal, spectral_theory,
  lp_recovery_part2_11, soltan_candes12}. There are fewer
strategies for the union of submanifolds model, a.k.a.\ manifold
clustering \cite{higher-order11, LocalPCA, zhu08multi, Gong2012,
  Kushnir06multiscale, wang2011spectral, ElhamifarV_nips11,
  6619442, Souvenir05, Haro06}. Notwithstanding, only
higher-order spectral clustering and spectral local PCA are
theoretically guaranteed~\cite{higher-order11,
  LocalPCA}. Multiscale strategies for data on Riemannian
manifolds were reported in \cite{rahman05}. The following
discussion is based on \cite{MMM.arxiv.14, MMM.15}, where tangent
spaces and angular information of submanifolds are utilized in a
novel way. Even of a different context, the basic principles of
\cite{rahman05} share common ground with those in
\cite{MMM.arxiv.14, MMM.15}. It is worth noting that a simplified
version of the algorithm in Sec.~\ref{sec:GCT} offers theoretical
guarantees. This paper attempts, for the first time in the
network-science literature, to exploit the first-order
(tangential) information of Riemannian submanifolds in clustering
dynamic time series.

\subsection{Notation}\label{sec:prelim}

\begin{figure}[!t]
  \centering
  \subfloat[Network-wide time series]
  {\includegraphics[width=.6\linewidth]{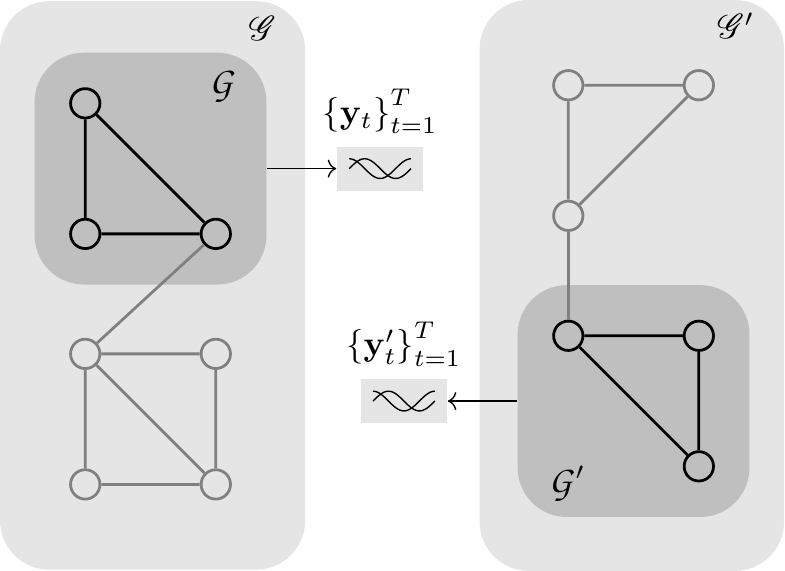}
    \label{fig:NetworkWideSignal}}\\
  \subfloat[Flowchart of the feature-extraction scheme]
  {\includegraphics[width=.95\linewidth]{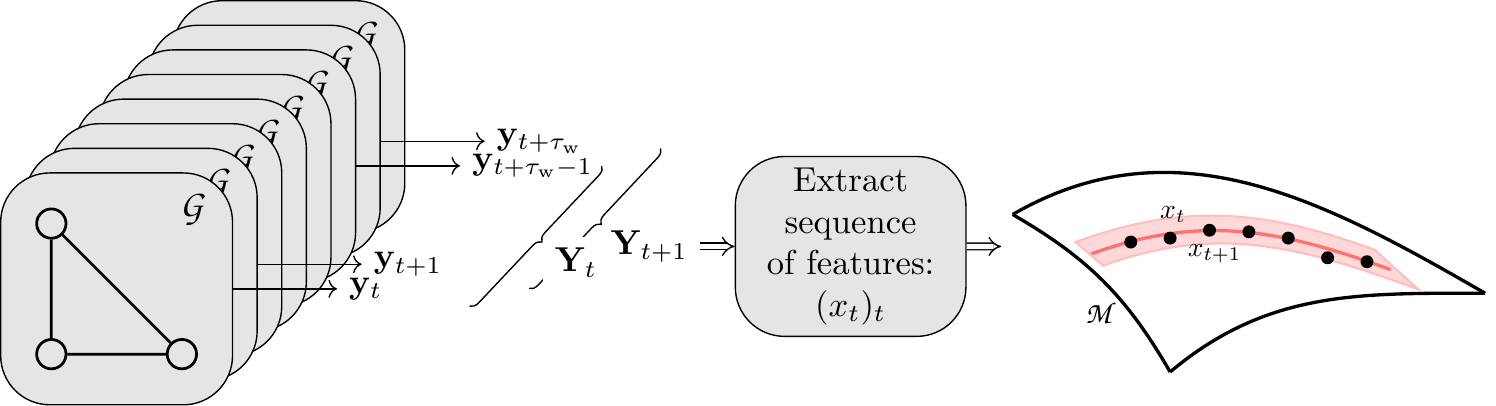}
    \label{fig:Flowchart}}
  \caption{(a) Subgraphs $\mathpzc{G}$ and $\mathpzc{G}'$ of the
    potentially different graphs $\mathscr{G}$ and
    $\mathscr{G}'$, respectively. Node $\nu$ of $\mathpzc{G}$
    emanates signal $y_{\nu t}$ (realization of a stochastic
    process) at discrete time $t$. All those values are gathered
    in the $N_{\mathpzc{G}} \times 1$ vector $\vect{y}_t$
    (snapshot of $\mathpzc{G}$ at time $t$). Such snapshots,
    observed over the time span $t\in\Set{1,2, \ldots, T}$, are
    collected into matrices
    $\vect{Y} := [\vect{y}_1, \ldots, \vect{y}_T]$ and
    $\vect{Y}' := [\vect{y}_1', \ldots, \vect{y}_T']$. The goal
    is to distinguish $\mathpzc{G}$ and $\mathpzc{G}'$ from the
    time-series information included in $\vect{Y}$ and
    $\vect{Y}'$. (b) A sliding window sequentially collects data
    $(\vect{Y}_t := [\vect{y}_{t}, \vect{y}_{t+1}, \ldots,
    \vect{y}_{t+\tau_{\text{w}}-1}])_t$ and extracts features
    (Secs.~\ref{sec:ARMA} and \ref{sec:PC}) which can be viewed
    as points on or close to a Riemannian submanifold (the
    Riemannian multi-manifold modeling hypothesis
    (RMMM)~\cite{MMM.arxiv.14,
      MMM.15}).}\label{fig:GraphModelAndFlowchart}
\end{figure}

Having $\Real$ and $\Integer$ stand for the set of all real and
integer numbers, respectively, let 
$\RealPP := (0,+\infty)$ and
$\IntegerPP := \Set{1,2,\ldots} \subset \Set{0,1,2,\ldots} =:
\IntegerP$. Column vectors and matrices are denoted by upright
boldfaced symbols, \eg, $\vect{y}$, while row vectors are denoted
by slanted boldfaced ones, \eg, $\bm{y}$. Vector/matrix
transposition is denoted by the superscript $\top$. Notation
$\vect{A}\!\!\succ\!\!(\succeq)\vect{0}$ characterizes a
symmetric positive (semi)definite [P(S)D] matrix. Consider a
(brain) network/graph
$\mathscr{G} := (\mathscr{N}, \mathscr{E})$, with sets of nodes
$\mathscr{N}$ and edges $\mathscr{E}$. In the case of fMRI data,
nodes could be defined as (contiguous) voxels belonging to either
anatomically defined or data-driven
regions~\cite{brain.tutorial.spm.13}. Each node
$\nu\in\mathscr{N}$ is annotated by a real-valued random variable
(r.v.) $Y_{\nu}$, whose realizations comprise the time series
associated with the $\nu$th node. Consider a subgraph
$\mathpzc{G} = (\mathpzc{V}, \mathpzc{E})$ of $\mathscr{G}$, with
cardinality $N_{\mathpzc{G}} := \lvert \mathpzc{V} \rvert$, \eg,
\begin{enumerate*}[label=\textbf{(\roman*)}]

\item $\mathpzc{G} = \mathscr{G}$; and

\item $\mathpzc{G}$ is a singleton $\mathpzc{G} = \Set{\nu}$,
  for some node $\nu$.

\end{enumerate*}
Realizations $\Set{y_{\nu t}}_{\nu\in\mathpzc{V}}$, or, a
snapshot of $\mathpzc{G}$ at the $t$th time instance, are
collected into the $N_{\mathpzc{G}} \times 1$ vector
$\vect{y}_t$, and form the $N_{\mathpzc{G}} \times T$ matrix
$\vect{Y} := [\vect{y}_1, \ldots, \vect{y}_T]$ over the time span
$t\in\Set{1, \ldots, T}$;
\cf~Fig.~\ref{fig:GraphModelAndFlowchart}. For subgraph
$\mathpzc{G}$, and a $\tau_{\text{w}}\in\IntegerPP$, which
represents the length of a ``sliding window'' that moves forward
along the time axis, snapshots
$(\vect{y}_{\tau})_{\tau=t}^{t + \tau_{\text{w}} -1}$ of
$\mathpzc{G}$ are gathered into the data matrix
$\vect{Y}_t := [\vect{y}_{t}, \vect{y}_{t+1}, \ldots,
\vect{y}_{t+\tau_{\text{w}}-1}]$;
\cf~Fig.~\ref{fig:Flowchart}. The following two sections
introduce two ways to capture intra-network connectivity patterns
and dynamics.

\section{ARMA Modeling}\label{sec:ARMA}

Motivated by Granger causality~\cite{Granger.69,
  Chen.frequency.Granger.06, Geweke.Granger.84,
  Boudjellaba.ARMA.Granger.92}, this section provides a scheme
for capturing spatio-temporal dependencies among network
nodes. Granger causality is built on a linear
\textit{auto-regressive (AR)}\/ model that approximates
$\vect{y}_t$ by a linear combination of the copies
$\Set{\vect{y}_{t-j}}_{j=1}^p$:
$\vect{y}_t := \sum_{j=1}^p \vect{D}_{j} \vect{y}_{t-j} +
\vect{v}_t$, for some $N_{\mathpzc{G}} \times N_{\mathpzc{G}}$
matrices $\Set{\vect{D}_j}_{j=1}^p$, $p\in\IntegerPP$, and
$\vect{v}_t$ is the r.v.\ that quantifies noise and modeling
inaccuracies. High-quality estimates of the $pN_{\mathpzc{G}}^2$
entries of $\Set{\vect{D}_j}_{j=1}^p$ require a large number of
training data, and thus an abundance of computational resources,
especially in cases of large-scale networks. The following
discussion provides a way to reduce the number of unknowns in the
previous identification task by capitalizing on the
\textit{low-rank}\/ arguments of the more general (linear)
\textit{auto-regressive moving average (ARMA)} model.

ARMA models are powerful parametric tools for spatio-temporal
series analysis with numerous applications in signal processing,
controls and machine learning \cite{Ljung.book,
  aggarwal.etal.icpr.04, Turaga+11}. ARMA modeling describes
$\vect{y}_{\tau}$ via the
$\rho\times 1\ (\rho\ll N_{\mathpzc{G}})$ latent vector
$\vect{z}_{\tau}$~\cite[\S10.6, p.~340]{Ljung.book}:
% $\forall \tau\in \Set{t, \ldots, t+\tau_{\text{w}}-1}$,
\begin{subequations}\label{arma}%
  \begin{align}
    \vect{z}_{\tau}
    & = \sum\nolimits_{j=1}^p \vect{A}_j\vect{z}_{\tau-j} +
      \vect{w}_{\tau}\,, \label{state.eq}\\ 
    \vect{y}_{\tau}
    & =\vect{C}\vect{z}_{\tau} +
      \vect{v}_{\tau}\,, \label{space.eq}
  \end{align}
\end{subequations}%
where
\begin{enumerate*}[label=\textbf{(\roman*)}]

\item \eqref{state.eq} is called the \textit{state}\/ and
  \eqref{space.eq} the \textit{space}\/ equation;

\item $\rho$ is the \textit{order}\/ of the model;

\item $\vect{C} \in\Real^{N_{\mathpzc{G}} \times \rho}$
  is the \textit{observation}\/ and
  $\Set{\vect{A}_j}_{j=1}^p \subset \Real^{\rho\times\rho}$ the
  \textit{transition}\/ matrices; and

\item $\vect{v}_{\tau}$ as well as $\vect{w}_{\tau}$ are
  realizations of zero-mean, white-noise random processes,
  uncorrelated both w.r.t.\ each other and $\vect{y}_{\tau}$.

\end{enumerate*}
% Matrix $\vect{C}$ filters and diffuses $\vect{z}_{\tau}$ which
% drives the activity in $\mathpzc{G}$, \eg, a common task
% $\vect{z}_{\tau}$, that needs to be accomplished by
% $\mathpzc{G}$, drives signal $\vect{y}_{\tau}$ in
% \eqref{space.eq}.
As in AR modeling, matrices $\Set{\vect{A}_j}_{j=1}^p$ manifest
causality throughout the process $\Set{\vect{z}_t}$.
% In the case where $\mathpzc{G}$ is a singleton, \ie,
% $\mathpzc{G} = \Set{\nu}$, \eqref{space.eq} collects
% scalar-valued time series observed at $\nu$:
% $y_{\tau} = \vect{C}\vect{z}_{\tau} + v_{\tau}$, where, now,
% $\vect{C}$ is an $1\times\rho$ vector.
The system identification problem \eqref{arma} requires 
estimation of the $N_{\mathpzc{G}}\rho + p\rho^2$ entries of
$\vect{C}$ and $\Set{\vect{A}_j}_{j=1}^p$, which are many less
than the $pN_{\mathpzc{G}}^2$ ones in the AR modeling case,
provided that $\rho\ll N_{\mathpzc{G}}$. For example, any
$0<\varpi \leq [(1+4p^2)^{1/2}-1]/(2p)$ guarantees that for
$\rho:= \varpi N_{\mathpzc{G}}$,
$N_{\mathpzc{G}}\rho + p\rho^2\leq pN_{\mathpzc{G}}^2$.

To simplify \eqref{arma}, re-define $\vect{z}_{\tau}$ and
$\vect{w}_{\tau}$ as the $p\rho\times 1$ vectors
$[\vect{z}_{\tau}^{\top}, \vect{z}_{\tau-1}^{\top}, \ldots,
\vect{z}_{\tau-p+1}^{\top}]^{\top}$ and
$[\vect{w}_{\tau}^{\top}, \vect{0}^{\top}, \ldots,
\vect{0}^{\top}]^{\top}$, respectively. Then, it can be easily
verified that there exist a $p\rho\times p\rho$ matrix
$\vect{A}_0$ and an
% \begin{align*}
%   \vect{A}_0 :=
%   \begin{bmatrix}
%     \vect{A}_1 & \ldots & \vect{A}_p\\
%     \vect{I}_{\rho} & \ldots & \vect{0}\\
%     \vdots & \ddots & \vdots\\
%     \vect{0} & \ldots & \vect{I}_{\rho}
%   \end{bmatrix}\,,
% \end{align*}
% as well as the
$N_{\mathpzc{G}}\times p\rho$ matrix $\vect{C}_0$ such that
\eqref{arma} is recast as
\begin{align}\label{arma.II}%
  \vect{z}_{\tau}
  = \vect{A}_0\vect{z}_{\tau-1} +
  \vect{w}_{\tau}\,, \qquad \vect{y}_{\tau}
  =\vect{C}_0\vect{z}_{\tau} +
  \vect{v}_{\tau} \,.
\end{align}
Further, it can be verified by \eqref{arma.II} that for any
$i\in \IntegerP$,
\begin{align*}
  \vect{y}_{t+i} = \vect{C}_0\vect{A}_0^i \vect{z}_t +
  \sum\nolimits_{j=1}^i \vect{C}_0 \vect{A}_0^{i-j} \vect{w}_{t+j} +
  \vect{v}_{t+i}\,,
\end{align*}
where $\vect{A}_0^0 := \vect{I}_{p\rho}$ and
$\sum\nolimits_{j=1}^0 \vect{C}_0 \vect{A}_0^{-j} \vect{w}_{t+j} :=
\vect{0}$. Fix now an $m\in\IntegerPP$ and define the
$m N_{\mathpzc{G}} \times 1$ vector
\begin{align}
  \bm{\mathcalboondox{y}}_{\text{f}\tau} :=
  [\vect{y}_{\tau}^{\top}, 
  \vect{y}_{\tau+1}^{\top}\ldots,
  \vect{y}_{\tau+m-1}^{\top}]^{\top}\,, \label{def.yf}
\end{align}
where sub-script $\text{f}$ stresses the fact that one moves
\textit{forward}\/ in time and utilizes data
$\Set{\vect{y}_{\tau'}}_{\tau' = \tau}^{\tau + m - 1}$ to define
$\bm{\mathcalboondox{y}}_{\text{f}\tau}$. It can be verified that
$\bm{\mathcalboondox{y}}_{\text{f}\tau} = \vect{O}^{(m)}
\vect{z}_{\tau} + \bm{\mathcalboondox{e}}_{\text{f}\tau}$, where
$\vect{O}^{(m)}$ is the $m$\textit{th-order observability}\/
matrix of size $mN_{\mathpzc{G}} \times p\rho$:
$\vect{O}^{(m)} := [\vect{C}_0^{\top}, (\vect{C}_0
\vect{A}_0)^{\top}, \ldots,
(\vect{C}_0\vect{A}_0^{m-1})^{\top}]^{\top}$, and
$\bm{\mathcalboondox{e}}_{\text{f}\tau}$ is defined as the vector
whose entries from $iN_{\mathpzc{G}} +1$ till
$(i+1)N_{\mathpzc{G}}$, for $i\in\Set{0, \ldots, m-1}$, are given
by
$\sum\nolimits_{j=1}^{i} \vect{C}_0\vect{A}_0^{i-j}
\vect{w}_{t+j} + \vect{v}_{t+i}$. Since
$\bm{\mathcalboondox{e}}_{\text{f}\tau}$ contains zero-mean noise
terms, it can be also verified that the conditional expectation
of $\bm{\mathcalboondox{y}}_{\text{f}\tau}$ given
$\vect{z}_{\tau}$ is
$\expect\Set{\bm{\mathcalboondox{y}}_{\text{f}\tau} \given
  \vect{z}_{\tau}} = \vect{O}^{(m)} \vect{z}_{\tau}$.

It is well-known that any change of basis
$\tilde{\vect{z}}_{\tau} := \vect{P}^{-1}\vect{z}_{\tau}$ in the
state space, where $\vect{P}$ is non-singular, renders
\begin{align}\label{arma.tilde}
  \tilde{\vect{z}}_{\tau}
  = \vect{P}^{-1} \vect{A}_0\vect{P} \tilde{\vect{z}}_{\tau-1} +
  \tilde{\vect{w}}_{\tau}\,, \qquad
  \vect{y}_{\tau} 
  = \vect{C}_0 \vect{P} \tilde{\vect{z}}_{\tau} + \vect{v}_{\tau} 
  \,,
\end{align}
with observation and transition matrices
$\tilde{\vect{C}}_0 := \vect{C}_0 \vect{P}$ and
$\tilde{\vect{A}}_0 := \vect{P}^{-1} \vect{A}_0 \vect{P}$,
respectively, equivalent to \eqref{arma.II} in the sense of describing
the same signal
$\vect{y}_{\tau}$~\cite[\S10.6]{Ljung.book}. % Indeed, for any integer
% $k$,
% \begin{align*}
%   \expect\Set{\tilde{\vect{y}}_{t+k} \tilde{\vect{y}}_t^{\top}}
%   & = \tilde{\vect{C}} \tilde{\vect{A}}^k
                  %                    \expect\Set{\tilde{\vect{z}}_t
            %             \tilde{\vect{z}}_t^{\top}} \tilde{\vect{C}}^{\top} + \delta_{k0}
            %             \expect\Set{\vect{v}_{t+k} \vect{v}_t^{\top}} \\ 
    %           & = \vect{C} \vect{A}^k \expect\Set{\vect{z}_t
                  %                   \vect{z}_t^{\top}} \vect{C}^{\top} + \delta_{k0}
                  %                   \expect\Set{\vect{v}_{t+k} \vect{v}_t^{\top}}\\
    %           & = \expect\Set{\vect{y}_{t+k}\vect{y}_t^{\top}} \,,
                  %   \end{align*}
                  %                   where $\delta_{k0}$ stands for the Kronecker delta.
The observability matrix of \eqref{arma.tilde} satisfies
$\tilde{\vect{O}}^{(m)} = \vect{O}^{(m)} \vect{P}$. Remarkably,
due to the non-singularity of $\vect{P}$, even if
$\tilde{\vect{O}}^{(m)}\neq \vect{O}^{(m)}$, their columns span
the \textit{same}\/ linear subspace.

Given the previous ambiguity of ARMA modeling w.r.t.\ $\vect{P}$,
to extract features that uniquely characterize \eqref{arma.II},
it is preferable to record the column space of $\vect{O}^{(m)}$,
instead of $\vect{O}^{(m)}$ itself. To this end, notice that for
small values of $p\rho$, it is often the case in practice to have
$mN_{\mathpzc{G}}\gg p\rho$, which renders the ``tall''
$\vect{O}^{(m)}$ full-column rank, with high probability. The
``column space'' of $\vect{O}^{(m)}$ becomes a
$(p\rho)$-dimensional linear subspace of
$\Real^{mN_{\mathpzc{G}}}$, or equivalently, a point in the
\textit{Grassmannian}\/
$\text{Gr}(mN_{\mathpzc{G}}, p\rho) := \{$all $(p\rho)$-rank
linear subspaces of $\Real^{m N_{\mathpzc{G}}}\}$. Apparently,
$\text{Gr}(m N_{\mathpzc{G}}, p\rho)$ is a (smooth) Riemannian
manifold of dimension
$p\rho(m N_{\mathpzc{G}} -p\rho)$~\cite{doCarmo.book.92,
  Tu.book.08}. The Grassmannian formulation removes the previous
$\vect{P}$-similarity-transform ambiguity in \eqref{arma.tilde}:
since any linear subspace possesses an orthonormal basis, it can
be easily verified that
$\text{Gr}(m N_{\mathpzc{G}}, p\rho) = \Set{[\vect{U}] \given
  \vect{U}\in\Real^{m N_{\mathpzc{G}} \times p\rho};
  \vect{U}^{\top} \vect{U} = \vect{I}_{p\rho}}$, where given the
orthogonal $\vect{U}$, point
$[\vect{U}]\in \text{Gr}(mN_{\mathpzc{G}}, p\rho)$ stands for
$[\vect{U}] := \Set{\vect{UP} \given \vect{P}\in
  \Real^{p\rho\times p\rho}\ \text{is non-singular}}$, \ie,
$[\vect{U}]$ gathers \textit{all}\/ bases for the column space of
$\vect{U}$.

Fix now a $\tau_{\text{f}}\in\IntegerPP$ and define the
$m N_{\mathpzc{G}}\times \tau_{\text{f}}$ matrices
\begin{align}
  \bm{\mathcalboondox{Y}}_{\text{f}\tau}
  & := [\bm{\mathcalboondox{y}}_{\text{f}\tau},
    \bm{\mathcalboondox{y}}_{\text{f},\tau+1}, \ldots,
    \bm{\mathcalboondox{y}}_{\text{f},\tau+\tau_{\text{f}}-1}] 
    \,, \label{def.Yf}\\
  \bm{\mathcalboondox{E}}_{\text{f}\tau} 
  & := [\bm{\mathcalboondox{e}}_{\text{f}\tau},
    \bm{\mathcalboondox{e}}_{\text{f},\tau+1}, \ldots,
    \bm{\mathcalboondox{e}}_{\text{f}, \tau+ \tau_{\text{f}}-1}]\,,
    \notag
\end{align}
as well as the $p\rho\times \tau_{\text{f}}$ matrix
$\vect{Z}_{\tau} := [\vect{z}_{\tau}, \ldots, \vect{z}_{\tau+
  \tau_{\text{f}}-1}]$. Then,
\begin{align}
  \bm{\mathcalboondox{Y}}_{\text{f}\tau} = \vect{O}^{(m)}
  \vect{Z}_{\tau} + \bm{\mathcalboondox{E}}_{\text{f}\tau}
  \,. \label{model.O} 
\end{align}
To obtain high-quality estimates of $\vect{O}^{(m)}$ from
\eqref{model.O}, choose a $\tau_{\text{b}}\in\IntegerPP$, and
define as in~\cite[\S10.6]{Ljung.book} the
$\tau_{\text{b}} N_{\mathpzc{G}} \times 1$ vector%
\begin{subequations}\label{def.backward}%
  \begin{align}
    \bm{\mathcalboondox{y}}_{\text{b}\tau} := \left[\vect{y}_{\tau}^{\top},
    \vect{y}_{\tau-1}^{\top}\ldots,
    \vect{y}_{\tau-\tau_{\text{b}}+1}^{\top}\right]^{\top}\,, \label{def.yb}
  \end{align}
  where, as opposed to \eqref{def.yf}, one moves $\tau_{\text{b}}$ steps
  \textit{backward} in time to define
  $\bm{\mathcalboondox{y}}_{\text{b}\tau}$. Let also the
  $\tau_{\text{b}} N_{\mathpzc{G}}\times \tau_{\text{f}}$ matrix
  \begin{align}
    \bm{\mathcalboondox{Y}}_{\text{b}\tau}
    := [\bm{\mathcalboondox{y}}_{\text{b}\tau},
    \bm{\mathcalboondox{y}}_{\text{b},\tau+1},
    \ldots, \bm{\mathcalboondox{y}}_{\text{b},\tau+\tau_{\text{f}}-1}
    ]\,. \label{def.Yb}
  \end{align}%
\end{subequations}
By \eqref{model.O},
\begin{align}
  & \tfrac{1}{\tau_{\text{f}}}
    \bm{\mathcalboondox{Y}}_{\text{f}, t+\tau_{\text{b}}}
    \bm{\mathcalboondox{Y}}_{\text{b}, t+\tau_{\text{b}}-1}^{\top} \notag\\
  % & = \vect{O}^{(m)} \tfrac{1}{\tau_{\text{f}}}
  %   \vect{Z}_{t+\tau_{\text{b}}} 
  %   \bm{\mathcalboondox{Y}}_{\text{b}, t+\tau_{\text{b}}-1}^{\top} +
  %   \tfrac{1}{\tau_{\text{f}}} 
  %   \bm{\mathcalboondox{E}}_{\text{f},t+\tau_{\text{b}}}
  %   \bm{\mathcalboondox{Y}}_{\text{b}, t+\tau_{\text{b}}-1}^{\top}
  %   \notag\\
  & = \vect{O}^{(m)} \tfrac{1}{\tau_{\text{f}}}
    \vect{Z}_{t+\tau_{\text{b}}} 
    \bm{\mathcalboondox{Y}}_{\text{b}, t+\tau_{\text{b}}-1}^{\top} +
    \tfrac{1}{\tau_{\text{f}}} \smashoperator{\sum_{\tau =
    t+\tau_{\text{b}}}^{t + \tau_{\text{b}} + \tau_{\text{f}} - 1}}
    \bm{\mathcalboondox{e}}_{\text{f}, \tau}
    \bm{\mathcalboondox{y}}_{\text{b}, \tau-1}^{\top} \,.
    \label{model.4.regression}
\end{align}
To avoid any confusion regarding time indices, it is required
that
$\tau_{\text{w}} \geq \tau_{\text{f}} + \tau_{\text{b}} + m
-1$. Notice also that
$\sum_{\tau} \bm{\mathcalboondox{e}}_{\text{f}, \tau}
\bm{\mathcalboondox{y}}_{\text{b}, \tau-1}^{\top}$ comprises
terms that result from the cross-correlations of
$\vect{y}_{\tau}$ with noise vectors $\vect{w}_{\tau'}$ and
$\vect{v}_{\tau''}$, recorded at time instants $\tau'$ and
$\tau''$ that lie ahead of $\tau$, and for which, according to
the initial modeling assumptions, $\vect{y}_{\tau}$ is
uncorrelated with $\vect{w}_{\tau'}$ and $\vect{v}_{\tau''}$. If
$\tau_{\text{f}}$ is set to be large, the law of large numbers
suggests that the sample correlations in
$(1/\tau_{\text{f}}) \sum_{\tau}
\bm{\mathcalboondox{e}}_{\text{f}\tau}
\bm{\mathcalboondox{y}}_{\text{b}, \tau-1}^{\top}$ approximate
well the ensemble ones, which, as previously stated, are zero.

Motivated by \eqref{model.4.regression}, the estimation task of the
observability matrix becomes as follows:
\begin{align}
  \left(\hat{\vect{O}}^{(m)}_t, \hat{\bm{\Pi}}_t\right)
  \in \Argmin_{\substack{\vect{O}\in\Real^{mN_{\mathpzc{G}}
  \times p\rho}\\
  \bm{\Pi}\in\Real^{p\rho\times \tau_{\text{b}} N_{\mathpzc{G}}}}}
  \norm*{\tfrac{1}{\tau_{\text{f}}}
  \bm{\mathcalboondox{Y}}_{\text{f},t+\tau_{\text{b}}}
  \bm{\mathcalboondox{Y}}_{\text{b}, t+\tau_{\text{b}}-1}^{\top} -
  \vect{O} \bm{\Pi}}_{\text{F}}^2 \,. \label{estimate.O}
\end{align}
If $r$ denotes the rank of
$({1}/{\tau_{\text{f}}}) \bm{\mathcalboondox{Y}}_{\text{f},
  t+\tau_{\text{b}}} \bm{\mathcalboondox{Y}}_{\text{b},
  t+\tau_{\text{b}}-1}^{\top}$, then its \textit{thin}\/ SVD is
$({1}/{\tau_{\text{f}}}) \bm{\mathcalboondox{Y}}_{\text{f},
  t+\tau_{\text{b}}} \bm{\mathcalboondox{Y}}_{\text{b},
  t+\tau_{\text{b}}-1}^{\top} = \vect{U} \bm{\Sigma}
\vect{V}^{\top}$, where
$\vect{U} \in \Real^{mN_{\mathpzc{G}}\times r}$ and
$\vect{V}\in\Real^{\tau_{\text{b}} N_{\mathpzc{G}} \times r}$ are
orthogonal matrices, \ie,
$\vect{U}^{\top} \vect{U} = \vect{I}_{r} = \vect{V}^{\top}
\vect{V}$, and $\bm{\Sigma}$ is the $r\times r$ diagonal matrix
whose diagonal elements gather, in descending order, the non-zero
singular values of
$({1}/{\tau_{\text{f}}}) \bm{\mathcalboondox{Y}}_{\text{f},
  t+\tau_{\text{b}}} \bm{\mathcalboondox{Y}}_{\text{b},
  t+\tau_{\text{b}}-1}^{\top}$. Assuming that $p\rho\leq r$, the
celebrated Schmidt-Mirsky-Eckart-Young theorem~\cite{Ben.Israel}
suggests that a solution to \eqref{estimate.O} is given by
$\hat{\vect{O}}^{(m)}_t = \vect{U}_{:,1:p\rho}$, where
$\vect{U}_{:,1:p\rho}$ gathers the first $p\rho$ columns of
$\vect{U}$, and
$\hat{\bm{\Pi}}_t = \bm{\Sigma}_{1:p\rho, 1:p\rho}
\vect{V}^{\top}_{:,1:p\rho}$. The previous procedure of
extracting a sequence of features
$\Set{x_t := [\hat{\vect{O}}^{(m)}_t]}_t$ in the Grassmannian
$\text{Gr}(mN_{\mathpzc{G}},p\rho)$ is summarized in
Alg.~\ref{algo:ARMA.observability}. The dependence of the
estimate $\hat{\vect{O}}^{(m)}_t$ on $t$ as well as its
on-the-fly computation allow also for the application of the
previous framework to \textit{dynamical}\/ ARMA models verbatim,
\ie, the case where matrices $\vect{A}_0 := \vect{A}_{0t}$ and
$\vect{C}_0 := \vect{C}_{0t}$ are not fixed but are functions of
time in \eqref{arma.II}.

\begin{algorithm}[!t]
  \begin{algorithmic}[1]
    \renewcommand{\algorithmicindent}{1em}

    \Require\parbox[t]{\dimexpr\linewidth-3em}%
    {Data $\vect{Y} = [\vect{y}_1, \ldots, \vect{y}_T]$; window
      size $\tau_{\text{w}}$; ARMA-model order $\rho$,
      observability-matrix order $m$; parameters
      $\tau_{\text{f}}, \tau_{\text{b}}$ s.t.\
      $\tau_{\text{w}} \geq \tau_{\text{f}} + \tau_{\text{b}} + m
      -1$.}

    \Ensure\parbox[t]{\dimexpr\linewidth-3em}% 
    {Sequence $(x_t)_{t=1}^{T-\tau_{\text{w}}+1}$ in
      $\text{Gr}(mN_{\mathpzc{G}},p\rho)$.}

    \For{$t= 1, \ldots, T-\tau_{\text{w}}+1$}

    \State\parbox[t]{\dimexpr\linewidth-\algorithmicindent}%
    {Consider data $\vect{Y}_t := [\vect{y}_t, \ldots,
      \vect{y}_{t+\tau_{\text{w}}-1}]$.} 
    
    \State\parbox[t]{\dimexpr\linewidth-\algorithmicindent}%
    {Form $\bm{\mathcal{Y}}_{\text{f}, t+\tau_{\text{b}}}$ and
      $\bm{\mathcal{Y}}_{\text{b}, t+\tau_{\text{b}}-1}$ by
      \eqref{def.Yf} and \eqref{def.Yb}, respectively.}

    \State\parbox[t]{\dimexpr\linewidth-\algorithmicindent}%
    {Compute the SVD
      $(1/\tau_{\text{f}}) \bm{\mathcal{Y}}_{\text{f}, t+\tau_{\text{b}}}
      \bm{\mathcal{Y}}_{\text{b}, t+\tau_{\text{b}}-1}^{\top} =
      \vect{U} \bm{\Sigma} \vect{V}^{\top}$.}

    \State\parbox[t]{\dimexpr\linewidth-\algorithmicindent}%
    {Define $x_t := [\hat{\vect{O}}^{(m)}_t] := [\vect{U}_{:,1:p\rho}]$
      in $\text{Gr}(mN_{\mathpzc{G}},p\rho)$.}

    \EndFor
    
  \end{algorithmic}
  \caption{Extracting features $(x_t)_t$ in
    $\text{Gr}(mN_{\mathpzc{G}},p\rho)$.}\label{algo:ARMA.observability}
\end{algorithm}

\section{Kernel-Based Partial 
  Correlations}\label{sec:PC}

Partial correlation (PC) will be used as a measure of similarity
among nodes of $\mathpzc{G}$ since it is both intuitively well
suited to the task, and has well-documented merits in
network-connectivity studies~\cite{Kolaczyk,
  smith.net.models.fmri.11, Karanikolas.icassp.16}. Given data
$\vect{Y} := [\vect{y}_1, \ldots, \vect{y}_T]$, form
$\tilde{\vect{Y}} := [\tilde{\vect{y}}_1, \ldots,
\tilde{\vect{y}}_T] := [\vect{y}_1-\bm{\mu}, \ldots,
\vect{y}_T-\bm{\mu}]$ to remove from data the sample averages or
offsets $\bm{\mu} := (1/T) \sum_{t=1}^T \vect{y}_t$. Along the
lines of Sec.~\ref{sec:ARMA}, consider
$\tilde{\vect{Y}}_{t} := [\tilde{\vect{y}}_t,
\tilde{\vect{y}}_{t+1}, \ldots, \tilde{\vect{y}}_{t +
  \tau_{\text{w}} -1}]$ for some $\tau_{\text{w}} \in\IntegerPP$.

Let $\tilde{\bm{y}}_{\nu t}$ denote the $\nu$th \textit{row}\/
vector of $\tilde{\vect{Y}}_t$, or in other words, the time
profile of the $\nu$th node of $\mathpzc{G}$ over time
$\Set{t, t+1, \ldots, t+\tau_{\text{w}}-1}$. Consider also a pair
of nodes $(i,j)\in \mathpzc{V}^2$, while
$\mathpzc{V}_{-ij} := \mathpzc{V}\setminus \Set{i,j}$. Rows
$\Set{\tilde{\bm{y}}_{\nu t}}_{\nu\in \mathpzc{V}_{-ij}}$ form
the matrix $\tilde{\vect{Y}}_{-ij,t}$, where subscript $-ij$
stresses the fact that $\tilde{\vect{Y}}_{-ij,t}$ is obtained
after the $i$th $\tilde{\bm{y}}_{it}$ and $j$th
$\tilde{\bm{y}}_{jt}$ rows are removed from
$\tilde{\vect{Y}}_t$. Let, now, $\hat{\tilde{\bm{y}}}_{it}$ and
$\hat{\tilde{\bm{y}}}_{jt}$ be the least-squares (LS) estimates
of $\tilde{\bm{y}}_{it}$ and $\tilde{\bm{y}}_{jt}$, respectively,
w.r.t.\ $\tilde{\vect{Y}}_{-ij,t}$, \ie,
$\hat{\tilde{\bm{y}}}_{lt} := \tilde{\bm{y}}_{lt}
\tilde{\vect{Y}}_{-ij,t}^{\dagger} \tilde{\vect{Y}}_{-ij,t}$,
$l\in\Set{i,j}$, with $\dagger$ denoting the Moore-Penrose
pseudoinverse of a matrix~\cite{Ben.Israel}, and
$\tilde{\vect{Y}}_{-ij,t}^{\dagger} \tilde{\vect{Y}}_{-ij,t}$
stands for the (orthogonal) projection operator onto the linear
span of
$\Set{\tilde{\bm{y}}_{\nu t}}_{\nu\in \mathpzc{V}_{-ij}}$. Upon
defining the residual
$\tilde{\bm{r}}_{lt} := \tilde{\bm{y}}_{lt} -
\hat{\tilde{\bm{y}}}_{lt}$, and provided that
$\tilde{\bm{r}}_{lt}\neq \bm{0}$, $l\in\Set{i,j}$, the (sample)
PC of the pair of nodes $(i,j)$ w.r.t.~$\mathpzc{V}_{-ij}$ is
defined as~\cite{Kolaczyk}
\begin{align}
  \hat{\varrho}_{ij,t}
  := \tilde{\bm{r}}_{it}
  \tilde{\bm{r}}_{jt}^{\top} / (\norm{\tilde{\bm{r}}_{it}}_2 \cdot
  \norm{\tilde{\bm{r}}_{jt}}_2) \,. \label{PC}
\end{align}
In the case where one of $\Set{\tilde{\bm{r}}_{it}, \tilde{\bm{r}}_{jt}}$
is zero, then $\hat{\varrho}_{ij,t}$ is also defined to be zero. In other
words, $\hat{\varrho}_{ij,t}$ measures the correlation between nodes $i$
and $j$, after removing the ``influence'' that nodes $\mathpzc{V}_{-ij}$
have on $(i,j)$. Notice that the numerator in \eqref{PC} is a
dot-vector product, since $\tilde{\bm{r}}_{lt}$, $l\in\Set{i,j}$, are
row vectors.

To capture possible non-linear dependencies among nodes, and
motivated by the success of reproducing kernel functions $\kappa$
in modeling non-linearities (\cf~Appendix~\ref{app:RKHS}), define
the $N_{\mathpzc{G}}\times N_{\mathpzc{G}}$ kernel matrix
$\vect{K}_t$ whose $(\nu, \nu')$th entry is
\begin{align}
  [\vect{K}_t]_{\nu\nu'} := \kappa(\tilde{\bm{y}}_{\nu t},
  \tilde{\bm{y}}_{\nu' t})\,. \label{def.kernel.matrix}
\end{align}
Further, define the following submatrices of $\vect{K}_t$:
\begin{align}
  \bm{k}_{-ij,i} 
  & : \text{$i$th row of $\vect{K}_t$ w.o.\ $i$th and $j$th
    entries}\,, \notag\\
  \bm{k}_{-ij,j} 
  & : \text{$j$th row of $\vect{K}_t$ w.o.\ $i$th and $j$th
    entries}\,, \notag\\
  \vect{K}_{-ij,t}
  & : \text{$\vect{K}_t$ w.o.\ $i$th and $j$th rows and
    columns}\,. \label{K-ij}
\end{align}
Moreover, define $\bm{\varphi}(\tilde{\vect{Y}}_{-ij,t})$ as the
$(N_{\mathpzc{G}}-2)\times \dim\mathpzc{H}$ vector, whose $\nu$th
entry ($\nu\in \mathpzc{V}_{-ij}$) is the element
$\varphi(\tilde{\bm{y}}_{\nu t})$ of space $\mathpzc{H}$
(\cf~Appendix~\ref{app:RKHS}). Then, the LS estimate
$\hat{\varphi}(\tilde{\bm{y}}_{it})$ of
$\varphi(\tilde{\bm{y}}_{it})$ w.r.t.\
$\Set{\varphi(\tilde{\bm{y}}_{\nu t}) \given \nu\in
  \mathpzc{V}_{-ij}}$ is given by
(\cf~Appendix~\ref{app:prop.kPC})
\begin{align}
  \hat{\varphi}(\tilde{\bm{y}}_{it})
  & = \bm{k}_{-ij,i}^{\top} \vect{K}_{-ij,t}^{\dagger}
    \bm{\varphi}(\tilde{\vect{Y}}_{-ij,t})\,. \label{LS.estimate}
\end{align}
As in \eqref{PC}, upon defining the LS-residual as
$\prescript{}{\kappa}{\tilde{r}}_{lt} :=
\varphi(\tilde{\bm{y}}_{lt}) -
\hat{\varphi}(\tilde{\bm{y}}_{lt})$, $l\in\Set{i,j}$, and
provided that both
$\Set{\prescript{}{\kappa}{\tilde{r}}_{it},
  \prescript{}{\kappa}{\tilde{r}}_{jt}}$ are non-zero, the
\textit{kernel (k)PC}\/ is defined as
\begin{align}
  \prescript{}{\kappa}{\hat{\varrho}}_{ij,t}
  := \innerp{\prescript{}{\kappa}{\tilde{r}}_{it}}
  {\prescript{}{\kappa}{\tilde{r}}_{jt}}_{\mathpzc{H}} / 
  (\norm{\prescript{}{\kappa}{\tilde{r}}_{it}}_{\mathpzc{H}} \cdot
  \norm{\prescript{}{\kappa}{\tilde{r}}_{jt}}_{\mathpzc{H}}) \,.
  \label{kPC}
\end{align}
In the case where one of
$\Set{\prescript{}{\kappa}{\tilde{r}}_{it},
  \prescript{}{\kappa}{\tilde{r}}_{jt}}$ is zero, then
$\prescript{}{\kappa}{\hat{\varrho}}_{ij,t}$ is defined to be zero. 

\begin{prop}\label{prop:kPC}
  \begin{subequations}
    Define the \textit{generalized Schur complement}\/ 
    $\vect{K}_t/\vect{K}_{-ij,t}$ of $\vect{K}_{-ij,t}$ in
    $\vect{K}_t$ as the following $2\times 2$ matrix
    \begin{align}
      \vect{K}_t/\vect{K}_{-ij,t} :=
      & \left[\begin{smallmatrix}
          [\vect{K}_t]_{ii} & [\vect{K}_t]_{ij} \\
          [\vect{K}_t]_{ji} & [\vect{K}_t]_{jj}
        \end{smallmatrix}\right] \notag\\
      & - \left[\begin{smallmatrix}
          \bm{k}_{-ij,i}\\
          \bm{k}_{-ij,j}
        \end{smallmatrix}\right] \vect{K}_{-ij,t}^{\dagger}
      \left[
      \begin{smallmatrix}
        \bm{k}_{-ij,i}^{\top} & \bm{k}_{-ij,j}^{\top}
      \end{smallmatrix}\right] \,. \label{gSC}
    \end{align}
    Then, the $(i,j)$th kPC is given by
    \begin{align}
      \prescript{}{\kappa}{\hat{\varrho}}_{ij,t} =
      \frac{[\vect{K}_t/ \vect{K}_{-ij,t}]_{12}}{\sqrt{[\vect{K}_t/
      \vect{K}_{-ij,t}]_{11}\cdot
      [\vect{K}_t/\vect{K}_{-ij,t}]_{22}}}\,. \label{prop:kPC.eq1}
    \end{align}
    If $\vect{K}_t$ is non-singular, then
    \begin{align}
      \prescript{}{\kappa}{\hat{\varrho}}_{ij,t} =
      \frac{-[\vect{K}_t^{-1}]_{ij}}
      {\sqrt{[\vect{K}_t^{-1}]_{ii} 
      [\vect{K}_t^{-1}]_{jj}}} \,. \label{prop:kPC.eq2}
    \end{align}
  \end{subequations}
\end{prop}

\begin{IEEEproof}
  See Appendix~\ref{app:prop.kPC}.
\end{IEEEproof}

According to \eqref{prop:kPC.eq2}, information about PCs is
contained in the positive definite (PD) matrix
$\bm{\Gamma}_t := (\diag \vect{K}_t^{-1})^{-1/2} \vect{K}_t^{-1}
(\diag \vect{K}_t^{-1})^{-1/2}$, where $\diag \vect{K}_t^{-1}$ is
the diagonal matrix whose main diagonal coincides with that of
$\vect{K}_t^{-1}$. It is well-known that the set of all
$N_{\mathpzc{G}}\times N_{\mathpzc{G}}$ PD matrices, denoted by
$\text{PD}(N_{\mathpzc{G}})$, is a (smooth) Riemannian manifold
of dimension $N_{\mathpzc{G}}(N_{\mathpzc{G}}+1)/2$. Assuming
that the dynamics of the network vary slowly w.r.t.~time, it is
conceivable that $\Set{x_t:= \bm{\Gamma}_t}$ constitute smooth
``trajectories'' in $\mathpzc{M}:= \text{PD}(N_{\mathpzc{G}})$ as
in Figs.~\ref{fig:Flowchart} and \ref{fig:RMMM}. Of course, there
are several other choices for points $x_t$ in $\mathpzc{M}$, \eg,
$\vect{K}_t$ or $\vect{K}_t^{-1}$, or the
$N_{\mathpzc{G}}\times N_{\mathpzc{G}}$ matrix $\vect{R}_t$,
whose $(\nu,\nu')$th entry is defined to be
$\kappa(\bm{y}_{\nu t}, \bm{y}_{\nu' t})$, with $\bm{y}_{\nu t}$
being the $\nu$th row of the data matrix $\vect{Y}$. In the case
where $\vect{K}_t$ is PSD, diagonal loading can be used to render
the matrix PD, \ie, $\vect{K}_t$ is re-defined as
$\vect{K}_t + \epsilon\vect{I}_{N_{\mathpzc{G}}}$, for some
$\epsilon\in\RealPP$. All the
previous choices for $x_t$ will be explored in
Sec.~\ref{sec:tests}.

\subsection{Designing the kernel matrix}\label{sec:design.kernel}

\subsubsection{Single kernel function}\label{sec:single.kernel}

There are numerous choices for the reproducing kernel function
$\kappa$, with the more popular ones being the linear, Gaussian,
and polynomial kernels (\cf~Appendix~\ref{app:RKHS}). Since
$\vect{K}_t$ is a Gram matrix, it is non-singular iff the
$(\dim\mathpzc{H})$-dimensional vectors
$\Set{\varphi(\tilde{\bm{y}}_{\nu t})}_{\nu=1}^{N_{\mathpzc{G}}}$
are linearly independent~\cite{Luenberger.Optim.Book}. The larger
$\dim\mathpzc{H}$ is, the more likely is for
$\Set{\varphi(\tilde{\bm{y}}_{\nu t})}_{\nu=1}^{N_{\mathpzc{G}}}$
to be linearly independent. This last remark justifies the choice
of a Gaussian kernel (yields an infinite-dimensional RKHS space;
\cf~Appendix~\ref{app:RKHS}) in the numerical tests of
Sec.~\ref{sec:tests}.

\subsubsection{Multiple kernel functions}\label{sec:multi.kernel}

For any user-defined set of reproducing kernel functions
$\Set{\kappa_l}_{l=1}^L$, with associated RKHSs
$\Set{\mathpzc{H}_l}_{l=1}^L$, and any set of positive weights
$\Set{\alpha_l}_{l=1}^L$, it can be verified that the kernel
function $\kappa := \sum_{l=1}^L \alpha_l \kappa_l$ is
reproducing, and induces an RKHS $\mathpzc{H}$ which is a linear
subspace of $\sum_{l=1}^L \mathpzc{H}_l$. Such a construction is
beneficial in cases where prior knowledge on the data does not
provide information on choosing adequately a single kernel
function that models data well. For example, whenever an adequate
variance $\sigma^2$ for a single Gaussian kernel
$\kappa_{\sigma}$ cannot be identified, then choosing the kernel
$\kappa := (1/L)\sum_{l=1}^L \kappa_{\sigma_l}$, for a set of
variances $\Set{\sigma_l}_{l=1}^L$ that cover the range of
interest, alleviates the problems that a designer faces due to
lack of prior information.

\subsubsection{Semidefinite embedding (SDE)}\label{sec:SDE}

In SDE the kernel matrix $\vect{K}_t$ becomes also part of the
data-driven learning process~\cite{SDE.06}. For convenience, the
discussion in Appendix~\ref{app:SDE} highlights SDE's key-points,
demonstrating that SDE can be cast as a convex-optimization task
over the set of PSD matrices.

\begin{algorithm}[!t]
  \begin{algorithmic}[1]
    \renewcommand{\algorithmicindent}{1em}

    \Require\parbox[t]{\dimexpr\linewidth-3em}%
    {Data $\vect{Y} = [\vect{y}_1, \ldots, \vect{y}_T]$; window size
      $\tau_{\text{w}}$; $\epsilon\in\RealPP$.}

    \Ensure\parbox[t]{\dimexpr\linewidth-3em}%
    {Sequence $(x_t)_{t=1}^{T-\tau_{\text{w}}+1}$ in
      $\text{PD}(N_{\mathpzc{G}})$.}

    \State{Form
      $\tilde{\vect{Y}} := [\tilde{\vect{y}}_1, \ldots,
      \tilde{\vect{y}}_T] := [\vect{y}_1-\bm{\mu}, \ldots,
      \vect{y}_T-\bm{\mu}]$, where $\bm{\mu} := (1/T) \sum_{t=1}^T
      \vect{y}_t$.}

    \For{$t= 1, \ldots, T-\tau_{\text{w}}+1$}

    \State\parbox[t]{\dimexpr\linewidth-\algorithmicindent}%
    {Consider the rows
      $\Set{\tilde{\bm{y}}_{\nu t}}_{\nu=1}^{N_{\mathpzc{G}}}$ of
      $\tilde{\vect{Y}}_t := [\tilde{\vect{y}}_t, \ldots,
      \tilde{\vect{y}}_{t+\tau_{\text{w}}-1}]$.}

    \State\parbox[t]{\dimexpr\linewidth-\algorithmicindent}%
    {Construct the kernel matrix $\vect{K}_t$ by using any of the
      methods demonstrated in Secs.~\ref{sec:single.kernel},
      \ref{sec:multi.kernel}, or
      \ref{sec:SDE}.}\label{alg:design.kernel.matrix} 

    \If{$\vect{K}_t$ is singular}
    \State{Re-define $\vect{K}_t$ as
      $\vect{K}_t+\epsilon\vect{I}_{N_{\mathpzc{G}}}$.} \label{alg:diagonal.loading}
    \EndIf

    \State\parbox[t]{\dimexpr\linewidth-\algorithmicindent}%
    {Define $x_t := (\diag \vect{K}_t^{-1})^{-1/2} \vect{K}_t^{-1}
      (\diag \vect{K}_t^{-1})^{-1/2}$.}\label{alg:kPC}

    \EndFor
    
  \end{algorithmic}
  \caption{Extracting features $(x_t)_t$ in
    $\text{PD}(N_{\mathpzc{G}})$}\label{algo:kPC}
\end{algorithm}

\section{Clustering Algorithm}\label{sec:GCT}

After features have been extracted from the network-wide time
series and mapped into a Riemannian feature space
(\cf.~Fig.~\ref{fig:Flowchart}), clustering is performed to
distinguish the disparate time series. To this end, a very short
introduction on Riemannian geometry will facilitate the following
discussion. For more details, the interested reader
is referred to~\cite{doCarmo.book.92, Tu.book.08}.

\subsection{Elements of manifold
  theory}\label{sec:prelim.manifolds}

\begin{figure}[!t]
  \centering
  \subfloat[Riemannian multi-manifold modeling (RMMM)]
  {\includegraphics[scale=.8]{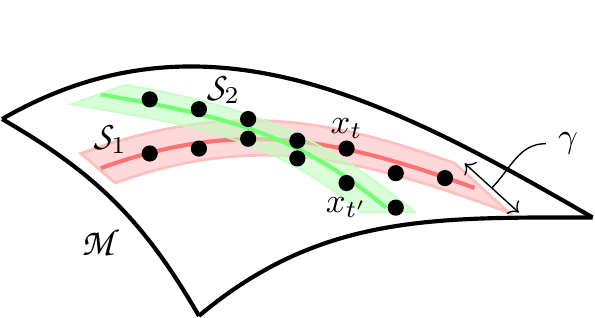}\label{fig:RMMM}}\\
  \subfloat[Logarithm and exponential map]
  {\includegraphics[scale=.8]{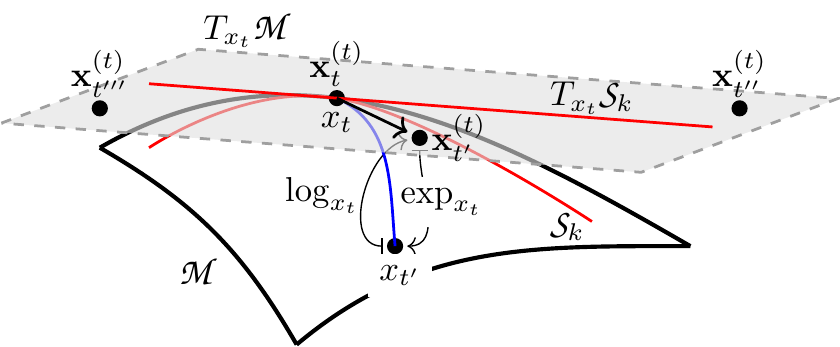}\label{fig:LogExp}}\\
  \subfloat[Estimating tangent spaces]
  {\includegraphics[scale=.8]{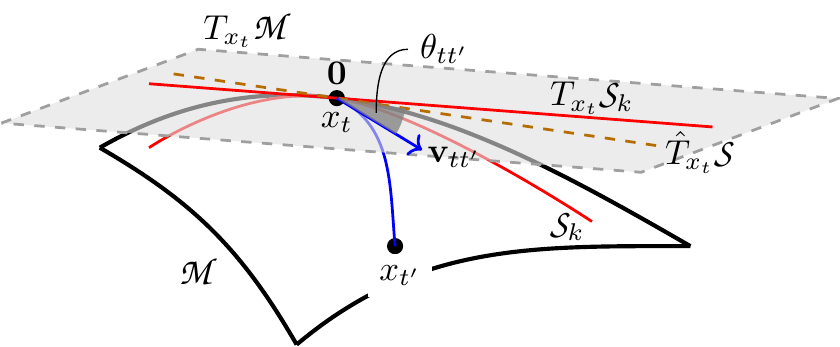}\label{fig:EstTangent}}
  \caption{Two ($K=2$) submanifolds/clusters
    $\mathpzc{S}_1, \mathpzc{S}_2$ and their tubular
    neighborhoods on the Riemannian manifold $\mathpzc{M}$, as
    well as the associated exponential and logarithm maps. In
    contrast to classical Kmeans, clusters are allowed here to
    have non-empty intersection.}
\end{figure}

Consider a $D$-dimensional Riemannian manifold $\mathpzc{M}$ with
metric $g$. Based on $g$, the (Riemannian) distance function
$\dist_g(x,y)$ between points $x, y\in \mathpzc{M}$ is
well-defined, and a geodesic is the (locally) distance-minimizing
curve in $\mathpzc{M}$ connecting $x$ and $y$. Loosely speaking,
geodesics generalize ``straight lines'' in Euclidean spaces to
shortest paths in the ``curved'' $\mathpzc{M}$ one. The RMMM
hypothesis, which this paper advocates, postulates that the
acquired data-points $\Set{x_t}$ are located on or ``close'' to
$K$ submanifolds (clusters) $\Set{\mathpzc{S}_k}_{k=1}^K$ of
$\mathpzc{M}$, with possibly different dimensionalities. In
contrast to the prevailing hypothesis for Kmeans, clusters in
RMMM are allowed to have \textit{non-empty intersection.} To
accommodate noise and mis-modeling errors, data $\Set{x_t}$ are
considered to lie within the following $\gamma$-width
($\gamma\in\RealPP$) tubular neighborhood
$\Set{x\in \mathpzc{M} \given \exists (s,k)\in \mathpzc{M}\times
  \Set{1, \ldots, K}\ \text{s.t.}\ s \in \mathpzc{S}_k\
  \text{and}\ \dist_g(x,s)< \gamma}$; see Fig.~\ref{fig:RMMM}.
If $T_{x_t}\mathpzc{M}$ denotes the tangent space of
$\mathpzc{M}$ at $x_t$ (a $D$-dimensional Euclidean space; see
Fig.~\ref{fig:LogExp}), and assuming that $x_t$ is located on a
submanifold $\mathpzc{S}_k$, then $T_{x_t}\mathpzc{S}_k$ stands
for the tangent space of the $d_k$-dimensional ($d_k<D$)
submanifold $\mathpzc{S}_k$ at $x_t$. Loosely speaking, the
exponential map $\exp_{x_t}(\cdot)$ maps a $D$-dimensional
tangent vector $\vect{v}\in T_{x_t}\mathpzc{M}$ to a point
$\exp_{x_t}(\vect{v})\in \mathpzc{M}$. If $\mathpzc{S}_k$ is
geodesic, \ie, it contains the geodesic defined by any two of its
points, then $\mathpzc{S}_k$ becomes the image of
$T_{x_t}\mathpzc{S}_k$ under $\exp_{x_t}$. The functional inverse
of $\exp_{x_t}$ is the logarithm map
$\log_{x_t}: \mathpzc{M}\to T_{x_t} \mathpzc{M}$, which maps
$x_t$ to the origin $\vect{0}$ of $T_{x_t}\mathpzc{M}$. Let
$\vect{x}_{t'}^{(t)}$ denote the image of a data point $x_{t'}$
via the logarithm map at $x_t$, \ie,
$\vect{x}_{t'}^{(t)} := \log_{x_{t}}(x_{t'})$. Having the number
of clusters/submanifolds $K$ known, the goal is to cluster
data-set $\mathpzc{X} := \Set{x_t}_{t\in\mathpzc{T}}$
($\mathpzc{T} = \Set{1, \ldots, T-\tau_{\text{w}}+1}$ in the
context of Secs.~\ref{sec:ARMA} and \ref{sec:PC}) into $K$ groups
$\Set{\mathpzc{X}_k}_{k=1}^K \subset \mathpzc{M}$ s.t.\ points in
$\mathpzc{X}_k$ are associated with the submanifold
$\mathpzc{S}_k$. Note that if $\mathpzc{M}$ is a Euclidean space,
and submanifolds are affine subspaces, then RMMM boils down to
the subspace-clustering
modeling~\cite{SubspaceClustering_Vidal}.

\subsection{Algorithm}\label{sec:algo}

Since the submanifold $\mathpzc{S}_k$, that point $x_t$ belongs
to, is unknown, so is $T_{x_t}\mathpzc{S}_k$. To this end, an
estimate of $T_{x_t}\mathpzc{S}_k$, denoted by
$\hat{T}_{x_t} \mathpzc{S}$, is associated with each point $x_t$
of the data-set. Given a user-defined parameter
$N_{\text{NN}}^{\text{GCT}}\in\IntegerPP$, let the neighborhood
\begin{align}\label{N.neighb}
  \mathpzc{T}_{\text{NN},t}^{\text{GCT}} := \Set*{t'\in\mathpzc{T}
  \given
  \begin{aligned}
    & x_{t'}\ \text{is one of the}\ N_{\text{NN}}^{\text{GCT}}\\
    & \text{nearest neighbors of}\ x_t
  \end{aligned}}\,,
\end{align}
where closeness is measured via $\dist_g(\cdot,\cdot)$, and
define $\hat{\vect{C}}_{x_t}$ as the ``local'' sample correlation
matrix
\begin{align}\label{corr.matrix}
  \hat{\vect{C}}_{x_t} := \tfrac{1}{N_{\text{NN}}^{\text{GCT}} -1} 
  \sum\nolimits_{t'\in \mathpzc{T}_{\text{NN},t}^{\text{GCT}}}
  \vect{x}_{t'}^{(t)}
  \vect{x}_{t'}^{(t)}{}^{\top}\,.
\end{align}
Moreover, let
$\norm{\hat{\vect{C}}_{x_t}} =
\lambda_{\max}(\hat{\vect{C}}_{x_t})$ denote the spectral norm of
$\hat{\vect{C}}_{x_t}$ as the maximum eigenvalue of the PSD
$\hat{\vect{C}}_{x_t}$. Assuming that $x_t$ lies close (in the
Riemannian-distance sense) to submanifold $\mathpzc{S}_k$,
estimates of the dimension $d_k$ of $\mathpzc{S}_k$, or
equivalently, of $T_{x_t}\mathpzc{S}_k$, can be obtained by
identifying a principal eigenspace $\hat{T}_{x_t}\mathpzc{S}$ of
$\hat{\vect{C}}_{x_t}$ via PCA arguments. Any method of
estimating a principal eigenspace can be employed here; \eg,
define $\hat{T}_{x_t}\mathpzc{S}$ as the linear subspace spanned
by the eigenvalues larger than or equal to
$\eta \lambda_{\max}(\hat{\vect{C}}_{x_t})$, for a user-defined
parameter $\eta\in(0,1)$ (\cf.~\cite{LocalPCA}). An illustration
of $\hat{T}_{x_t}\mathpzc{S}$ can be found in
Fig.~\ref{fig:EstTangent}. If $l(x_t,x_{t'})$ denotes the
(shortest) geodesic connecting $x_t$ and $x_{t'}$ in
$\mathpzc{M}$, and upon defining the tangent vector
$\vect{v}_{tt'} := \log_{x_t}(x_{t'})$, standing as the
``velocity'' of $l(x_t,x_{t'})$ at $x_t$, let the (empirical
geodesic) angle $\theta_{tt'}$ be defined as the angle between
$\vect{v}_{tt'}$ and the estimated linear subspace
$\hat{T}_{x_t}\mathpzc{S}$ of $T_{x_t}\mathpzc{M}$.

\begin{algorithm}[!t]
  \begin{algorithmic}[1]
    \renewcommand{\algorithmicindent}{1em}
    \setstretch{1}

    \Require Manifold $\mathpzc{M}$; number of clusters $K$; dataset
    $\Set{x_t}_{t\in\mathpzc{T}}$; the number of nearest neighbors
    $N_{\text{NN}}^{\text{GCT}}$; distance parameter $\sigma_d$
    (default $\sigma_d=1$); angle parameter $\sigma_a$ (default
    $\sigma_a=1$); eigenvalue threshold $\eta\in(0,1)$.

    \Ensure Data-cluster associations.

    \For{$t=1,\ldots,\lvert\mathpzc{T}\rvert$}
    
    \State\parbox[t]{\dimexpr\linewidth-\algorithmicindent}{
      Define neighborhood $\mathpzc{T}_{\text{NN},t}^{\text{GCT}}$
      [cf.~\eqref{N.neighb}].}

    \State\parbox[t]{\dimexpr\linewidth-\algorithmicindent}{Compute
      $\vect{x}_{t'}^{(t)} = \log_{x_t}(x_{t'})$, $\forall
      t'\in \mathpzc{T}_{\text{NN},t}^{\text{GCT}}$.}\label{alg:log}
    
    \State\parbox[t]{\dimexpr\linewidth-\algorithmicindent}{%
      \textbf{(Local sparse coding:)} Identify weights
      $\{\alpha_{tt'}\}_{t'\in\mathpzc{T}}$ via
      \eqref{sparse.coding}.}\label{alg:sparse.coding}
    
    \State\parbox[t]{\dimexpr\linewidth-\algorithmicindent}{Compute
      the sample correlation matrix $\hat{\vect{C}}_{x_t}$ by
      \eqref{corr.matrix}.}
    
    \State\parbox[t]{\dimexpr\linewidth-\algorithmicindent}{\textbf{(Local
        PCA:)} Identify the eigenvalues which are larger than or equal
      to $\eta\lambda_{\max}(\hat{\vect{C}}_{x_t})$, and call the
      eigenspace spanned by the associated eigenvalues
      $\hat{T}_{x_t}\mathcal{S}$.}\label{alg:PCA}
    
    \State\parbox[t]{\dimexpr\linewidth-\algorithmicindent}{\textbf{(Angular
        information:)} Compute the empirical geodesic angles
      $\{\theta_{tt'}\}_{t'\in\mathpzc{T}}$.}\label{alg:angles}
    \EndFor

    \State \setlength\abovedisplayskip{2.5pt}%
    \setlength\belowdisplayskip{2.5pt}%
    \setlength\abovedisplayshortskip{2.5pt}%
    \setlength\belowdisplayshortskip{2.5pt}%
    Form the $\lvert\mathpzc{T}\rvert \times \lvert\mathpzc{T}\rvert$
    affinity matrix $\vect{W}:= [w_{tt'}]_{(t,t')\in\mathpzc{T}^2}$ as
    \begin{align}\label{equ:sparseweights}
      w_{tt'} := \exp(|\alpha_{tt'}| + |\alpha_{t't}|) \cdot
      \exp\left(-\tfrac{\theta_{tt'} + \theta_{t't}}{\sigma_a}\right)
      \,.
    \end{align}
    
    \State Apply spectral clustering~\cite{luxburg_tutorial} to $\vect{W}$
    to identify data-cluster associations.\label{alg:affinity.matrix}
    
  \end{algorithmic}
  \caption{Geodesic clustering by tangent spaces (GCT).}\label{algo:GCT}
\end{algorithm}

Motivated by a very recent line of research~\cite{MMM.arxiv.14,
  MMM.15}, this paper advocates the \textit{geodesic clustering by
  tangent spaces (GCT)}\/ algorithm, detailed in Alg.~\ref{algo:GCT},
to solve the clustering task at hand. Key-points of GCT are the local
sparse coding of step~\ref{alg:sparse.coding}, local PCA of
step~\ref{alg:PCA}, and the extraction of the angular information at
step~\ref{alg:angles}. Regarding the sparse-coding step, after mapping
data-points $\Set{x_t}_{t\in\mathpzc{T}}$ to vectors
$\Set{\vect{x}_{t'}^{(t)}}_{t'\in\mathpzc{T}}$ in the tangent space
$T_{x_t} \mathpzc{M}$ at $x_t$, and motivated by the affine geometry
of $T_{x_t} \mathpzc{M}$ (\cf~Fig.~\ref{fig:LogExp}), ``relations''
between data within neighborhood $\mathpzc{T}_{\text{NN},t}^{\text{GCT}}$,
centered at $\vect{x}_{t}^{(t)}$, are captured by the amount that
neighbors
$\Set{\vect{x}_{t'}^{(t)}}_{t'\in
  \mathpzc{T}_{\text{NN},t}^{\text{GCT}}\setminus\Set{t}}$
($\vect{x}_{t'}^{(t)}$, $\vect{x}_{t''}^{(t)}$ and
$\vect{x}_{t'''}^{(t)}$ in Fig.~\ref{fig:LogExp}, for example)
contribute in the description of $\vect{x}_{t}^{(t)}$ via affine
combinations:
\begin{align}
  \min_{\Set{\alpha_{tt'}}_{t'\in
  \mathpzc{T}_{\text{NN},t}^{\text{GCT}} \setminus\Set{t}}}\
  & \overbrace{\norm*{\vect{x}_t^{(t)} - \sum\nolimits_{t'\in
    \mathpzc{T}_{\text{NN},t}^{\text{GCT}} \setminus\Set{t}} \alpha_{tt'} 
    \vect{x}_{t'}^{(t)}}_2^2}^{\text{Data-fit term}} \notag\\
  & + \underbrace{\sum_{t'\in
    \mathpzc{T}_{\text{NN},t}^{\text{GCT}} \setminus\Set{t}}
    \exp\left(\tfrac{\norm{\vect{x}_t^{(t)}-\vect{x}_{t'}^{(t)}}_2}{\sigma_d}
    \right) \lvert\alpha_{tt'}\rvert}_{\text{Sparsity-promoting term}} \notag\\
  & \text{s.to}\ \sum\nolimits_{t'\in \mathpzc{T}_{\text{NN},t}^{\text{GCT}}
    \setminus\Set{t}} \alpha_{tt'} =1 \,,
    \label{sparse.coding}
\end{align}
where the constraint in \eqref{sparse.coding} manifests that
neighbors should cooperate affinely to describe
$\vect{x}_{t}^{(t)}$ in the data-fit term. The regularization
term in \eqref{sparse.coding} enforces sparsity in the previous
representation by penalizing, thus eliminating, contributions
from neighbors which are located far from $\vect{x}_{t}^{(t)}$
via the weights
$\exp(\norm{\vect{x}_t^{(t)}-\vect{x}_{t'}^{(t)}}_2 / \sigma_d)$:
the larger the distance of $\vect{x}_{t'}^{(t)}$ from
$\vect{x}_t^{(t)}$ in the tangent space $T_{x_t} \mathpzc{M}$,
the larger the penalty on the modulus of the affine coefficient
$\alpha_{tt'}$. Moreover, no relations are established between
$\vect{x}_t^{(t)}$ and data points
$\Set{\vect{x}_{t'}^{(t)}}_{t'\in \mathpzc{T}\setminus
  \mathpzc{T}_{\text{NN},t}^{\text{GCT}}}$ which do not belong to
neighborhood $\mathpzc{T}_{\text{NN},t}^{\text{GCT}}$, by setting
$\alpha_{tt'} :=0$ for any
$t'\in \mathpzc{T}\setminus
\mathpzc{T}_{\text{NN},t}^{\text{GCT}}$. All information
collected in weights $\Set{\alpha_{tt'}}$ and
$\Set{\theta_{tt'}}$ are gathered in the affinity matrix
$\vect{W}$ (step~\ref{alg:affinity.matrix} of
Alg.~\ref{algo:GCT}) that is fed in any spectral clustering (SC)
algorithm that provides data-cluster associations. The
contribution of GCT~\cite{MMM.arxiv.14, MMM.15} in clustering on
Riemannian surfaces is the novel way of extraction and
incorporation of the angular information $\Set{\theta_{tt'}}$ in
an SC affinity matrix. A performance analysis, with guarantees on
the clustering accuracy and the number of mis-classified
data-points, has been already provided for a simplified version
of GCT, where submanifolds are considered to be ``geodesic,''
justifying thus the name GCT, the sparse-coding scheme of
step~\ref{alg:sparse.coding} in Alg.~\ref{algo:GCT} is not
employed, and the affinity matrix of
step~\ref{alg:affinity.matrix} becomes a binary one, with entries
either $1$ or $0$, depending on whether conditions on the
dimensions of the estimated tangent subspaces, the angular
information $\Set{\theta_{tt'}}$ and the Riemannian distance
between data-points are satisfied or not~\cite{MMM.arxiv.14}.

\subsection{Computational complexity}\label{sec:complexity}

A major part of GCT computations take place within the
neighborhood $\mathpzc{T}_{\text{NN},t}^{\text{GCT}}$. The
complexity for computing the $N_{\text{NN}}^{\text{GCT}}$
(typically $\leq 100$ in all numerical tests) nearest neighbors
of $x_t$ is
$(\lvert\mathpzc{T}\rvert \mathcal{C}_{\dist} +
N_{\text{NN}}^{\text{GCT}} \log\lvert\mathpzc{T}\rvert)$, where
$\mathcal{C}_{\dist}$ denotes the cost of computing the
Riemannian distance between any two points,
$\lvert\mathpzc{T}\rvert \mathcal{C}_{\dist}$ refers to the
complexity of computing $\lvert\mathpzc{T}\rvert-1$ distances,
and $N_{\text{NN}}^{\text{GCT}} \log\lvert\mathpzc{T}\rvert$
refers to the effort of identifying the
$N_{\text{NN}}^{\text{GCT}}$ nearest neighbors of $x_t$. Notice
that once the logarithm map $\log_{x_{t}}(x_{t'})$ is computed,
under complexity $\mathcal{C}_{\log}$
(\cf.~Appendix~\ref{app:LogMap}), then
$\mathcal{C}_{\dist} = \mathcal{O}(\dim\mathpzc{M})$. If
$\mathpzc{M}$ is the set $\text{PD}(N_{\mathpzc{G}})$, then
$\mathcal{C}_{\log} =
\mathcal{O}[\sqrt{(N_{\mathpzc{G}}(N_{\mathpzc{G}}+1)/2)^3}]$,
while
$\mathcal{C}_{\log} = \mathcal{O}(p^2\rho^2 mN_{\mathpzc{G}})$ if
$\mathpzc{M}$ is the Grassmannian
$\text{Gr}(mN_{\mathpzc{G}}, p\rho)$.

Step~\ref{alg:sparse.coding} of Alg.~\ref{algo:GCT} requires
solving the sparsity-promoting optimization task of
\eqref{sparse.coding}. Notice that due to $\norm{}_2$, only inner
products of Euclidean vectors are necessary to form the loss
function in \eqref{sparse.coding}, which entails a complexity of
order $\mathcal{O}(\dim\mathpzc{M})$. Given that only
$N_{\text{NN}}^{\text{GCT}}$ vectors are involved,
\eqref{sparse.coding} is a small-scale convex-optimization task
that can be determined efficiently (let $\mathcal{C}_{\text{sc}}$
denote that complexity) by any off-the-shelf
solver~\cite{Bauschke.Combettes.book}. Step~\ref{alg:PCA} of
Alg.~\ref{algo:GCT} involves the computation of the top
eigenvectors of the sample covariance matrix
$\hat{\vect{C}}_{x_t}$, under complexity of
$\mathcal{O}[\dim\mathpzc{M} +
(N_{\text{NN}}^{\text{GCT}})^3]$. Finally, to compute the
empirical geodesic angles,
$\mathcal{O}(\lvert\mathpzc{T}\rvert \mathcal{C}_{\log} +
\lvert\mathpzc{T}\rvert \dim\mathpzc{M})$ operations are
necessary. Spectral clustering is invoked in
step~\ref{alg:affinity.matrix} of Alg.~\ref{algo:GCT} on the
$\lvert\mathpzc{T}\rvert \times \lvert\mathpzc{T}\rvert$ affinity
matrix $\vect{W}$. Its main computational burden is to identify
$K$ eigenvectors ($K$ is the number of clusters) of $\vect{W}$,
which entails complexity of order
$\mathcal{O}(K\lvert\mathpzc{T}\rvert^2)$. To summarize, the
complexity of GCT is
$\mathcal{O}[\lvert\mathpzc{T}\rvert^2 (\mathcal{C}_{\dist} +
\mathcal{C}_{\log} + \dim\mathpzc{M} + K) +
N_{\text{NN}}^{\text{GCT}}
\lvert\mathpzc{T}\rvert\log\lvert\mathpzc{T}\rvert +
\lvert\mathpzc{T}\rvert \mathcal{C}_{\text{sc}} +
\lvert\mathpzc{T}\rvert \dim\mathpzc{M} + \lvert\mathpzc{T}\rvert
(N_{\text{NN}}^{\text{GCT}})^3]$.

\section{Numerical tests}\label{sec:tests}

To assess performance, the proposed GCT algorithm is compared
with the following methods:
\begin{enumerate}[label = \textbf{(\roman*)}]

\item Sparse manifold clustering (SMC)~\cite{ElhamifarV_nips11,
    6619442}. SMC was introduced in~\cite{ElhamifarV_nips11} for
  clustering submanifolds within Euclidean spaces, and it was
  later modified in~\cite{6619442} for clustering submanifolds on
  the sphere. SMC is adapted here, according to our needs, to
  cluster submanifolds in a Riemannian manifold, and still
  referred to as SMC. SMC's basic idea is as follows: Per each
  data-point $x$, a local neighborhood is mapped to the tangent
  space $T_x\mathpzc{M}$ by the logarithm map
  (\cf~step~\ref{alg:log} of Alg.~\ref{algo:GCT}), and a
  sparse-coding task (\cf~step~\ref{alg:sparse.coding} of
  Alg.~\ref{algo:GCT}) is solved in $T_x\mathpzc{M}$ to provide
  weights for an SC similarity matrix.
  
\item Spectral clustering~\cite{luxburg_tutorial} equipped with
  Riemannian metric (SCR)
  of~\cite{GOH_VIDAL08}. SCR~\cite{GOH_VIDAL08} utilizes SC under
  the weighted affinity matrix
  $[\vect{W}]_{tt'} := \exp[-\dist_g^2(x_t,x_{t'})/(2\sigma^2)]$,
  where the Riemannian distance metric $\dist_g(\cdot,\cdot)$ is
  used to quantify affinity among data-points~\cite{GOH_VIDAL08}.

\item Kmeans, where data lying in the Riemannian manifold are
  embedded into a Euclidean space, and then the classical Kmeans,
  under the classical (Euclidean) $\ell_2$-distance metric, is
  applied to the embedded dataset. In particular, Grassmannian
  manifolds are embedded into Euclidean spaces by the isometric
  embedding~\cite{machado1985grassmannian,
    Basri_nearest_subspace11}, and $\text{PD}(N_{\mathpzc{G}})$
  is embedded into $\Real^{N_{\mathpzc{G}}(N_{\mathpzc{G}}+1)/2}$
  by vectorizing the triangular upper part of the elements of
  $\text{PD}(N_{\mathpzc{G}})$. This set of tests stands as a
  representative of all schemes that do not exploit the
  underlying Riemannian geometry, as detailed in
  Sec.~\ref{sec:prior.art}.

\end{enumerate}
Unlike GCT, none of the previous methods utilizes the underlying
submanifold tangential information (Kmeans is even
Riemannian-geometry agnostic). In contrast to the prevailing
hypothesis of Kmeans and variants, that clusters are not
closely located to each other, RMMM allows for non-empty
intersections of submanifolds (\cf~Fig.~\ref{fig:RMMM}).

The ground-truth labels of clusters are available in each
experiment, and assessment is done via the notion of
\textit{clustering accuracy,} defined as ``($\#$ of points with
cluster labels equal to the ground-truth ones) $/$ ($\#$ of total
points).'' Signal-to-noise ratio (SNR) is set to be $10$dB for
all experiments. Tests are run for a number of $50$ realizations,
and average clustering accuracies, as well as standard
deviations, are depicted in the subsequent figures.

\subsection{Synthetically generated time
  series}\label{sec:synthetic.data}

This section refers to the setting of
Fig.~\ref{fig:motivate}. Per state, there are up to three
tasks/events/modules that need to be accomplished through the
cooperation of nodes. Each node contributes to a specific task by
sharing a common signal with other nodes assigned to the same
task. Nodes that share a common task are considered to be
connected to each other. Per node, the previous common signal is
linearly combined with a signal characteristic of the
node, and with a first-order auto-regressive (AR) process, with
time-varying AR coefficient, contributing to the dynamics of
the task-specific signal. The AR signal is described by the
recursion
$y_{\nu t, \text{AR}} := \cos\theta_t \cdot y_{\nu (t-1),
  \text{AR}} + \sqrt{1-\cos^2\theta_t}\cdot v_t$, where $v_t$ is
a zero-mean and unit-variance normal r.v., and
$\theta_{t} := \theta_{t-1} + \Delta\theta$, for some
user-defined parameters $\theta_0$ and $\Delta\theta$. The linear
combination of all the previous time series is filtered by the
model of \cite{simtb} to yield the BOLD data
$\Set{\vect{y}_t}_{t=1}^T$. % States and a realization of the
% resultant BOLD time series are depicted in the first two rows of
% Fig.~\ref{fig:motivate}, while the connectivity matrices for the
% four distinct states can be found in the third row of
% Fig.~\ref{fig:motivate}. As the hardly distinguishable sample
% correlation matrices in the fourth row of Fig.~\ref{fig:motivate}
% suggest, computing correlations along the entire time course of a
% single state does not offer any help in identifying clusters.

Regarding Alg.~\ref{algo:ARMA.observability} of
Sec.~\ref{sec:ARMA}, parameters are set as follows:
$N_{\mathpzc{G}} := 10$, $m = 3$, $p=1$, $\rho = 3$,
$\tau_{\text{f}} = 20$, $\tau_{\text{b}} = 20$, and
$\tau_{\text{w}} \in\Set{50,70,80}$. Results pertaining to the
observability-matrix features of Sec.~\ref{sec:ARMA} are denoted
by the ``OB'' tag in the legents of all subsequent figures.

Regarding the methodology of Sec.~\ref{sec:PC}, several features
are explored in the numerical tests. More specifically, with
reference to \eqref{def.kernel.matrix}, point
$x_t\in \text{PD}(N_{\mathpzc{G}})$ takes the following values:
\begin{enumerate*}[label=\textbf{(\roman*)}]

\item
  $(\diag \vect{K}_t^{-1})^{-1/2} \vect{K}_t^{-1} (\diag
  \vect{K}_t^{-1})^{-1/2}$ from step~\ref{alg:kPC} of
  Alg.~\ref{algo:kPC}, denoted by the tag ``kPC'' in the
  subsequent figures;

\item $\vect{K}_t$ from step~\ref{alg:diagonal.loading} of
  Alg.~\ref{algo:kPC}, denoted by tag ``Cov'';

\item $\vect{K}_t^{-1}$, denoted by tag ``ICov''; and

\item $\bm{\Lambda}_t$, where
  $[\bm{\Lambda}_t]_{\nu\nu'} := \kappa(\bm{y}_{\nu t},
  \bm{y}_{\nu't})$, with
  $\Set{\bm{y}_{\nu t}}_{\nu=1}^{N_{\mathpzc{G}}}$ being the rows
  of $\vect{Y}_{t} := [\vect{y}_t, \vect{y}_{t+1}, \ldots,
  \vect{y}_{t + \tau_{\text{w}} -1}]$, and denoted by tag
  ``Corr''.

\end{enumerate*}

Constructing a reproducing kernel function $\kappa$, or the
sequence of kernel matrices $\Set{\vect{K}_t}$ in
step~\ref{alg:design.kernel.matrix} of Alg.~\ref{algo:kPC}, plays
a principal role in the methodology of Sec.~\ref{sec:PC}. To this
end and along the lines of Sec.~\ref{sec:design.kernel}, four
ways of designing the kernel matrices are explored:
\begin{enumerate}[label=\textbf{(\roman*)}]

\item \textbf{Linear kernel function:} By choosing
  $\kappa_{\text{l}}$ of Appendix~\ref{app:RKHS} as the kernel
  function, the feature space $\mathpzc{H}$ becomes nothing but
  the input Euclidean $\Real^{\tau_{\text{w}}}$ one, with
  $\kappa_{\text{l}} (\bm{y}, \bm{y}') = \bm{y}
  \bm{y}'{}^{\top}$, for any
  $\bm{y}, \bm{y}'\in \Real^{\tau_{\text{w}}}$. As such, the
  previously met $\vect{K}_t$ and $\bm{\Lambda}_t$ become the
  classical covariance and correlation matrices, respectively. As
  Figs.~\ref{fig:synth.linear.kernel.50.12}--%
  \ref{fig:synth.linear.kernel.80.16} demonstrate, the larger the
  values of the sliding window $\tau_{\text{w}}$ and the number
  of nearest neighbors $N_{\text{NN}}^{\text{GCD}}$ are, the
  better \textit{all}\/ methods perform. However, GCT exhibits
  the best performance even for small values of those parameters,
  particularly for the advocated features of kPC and
  observability matrices (``OB''). Further, focusing on these two
  features, it can be seen that ``OB'' outperforms kPC in almost
  all scenarios.

  \begin{figure}[!t]
    \centering
    \includegraphics[width=.8\linewidth]{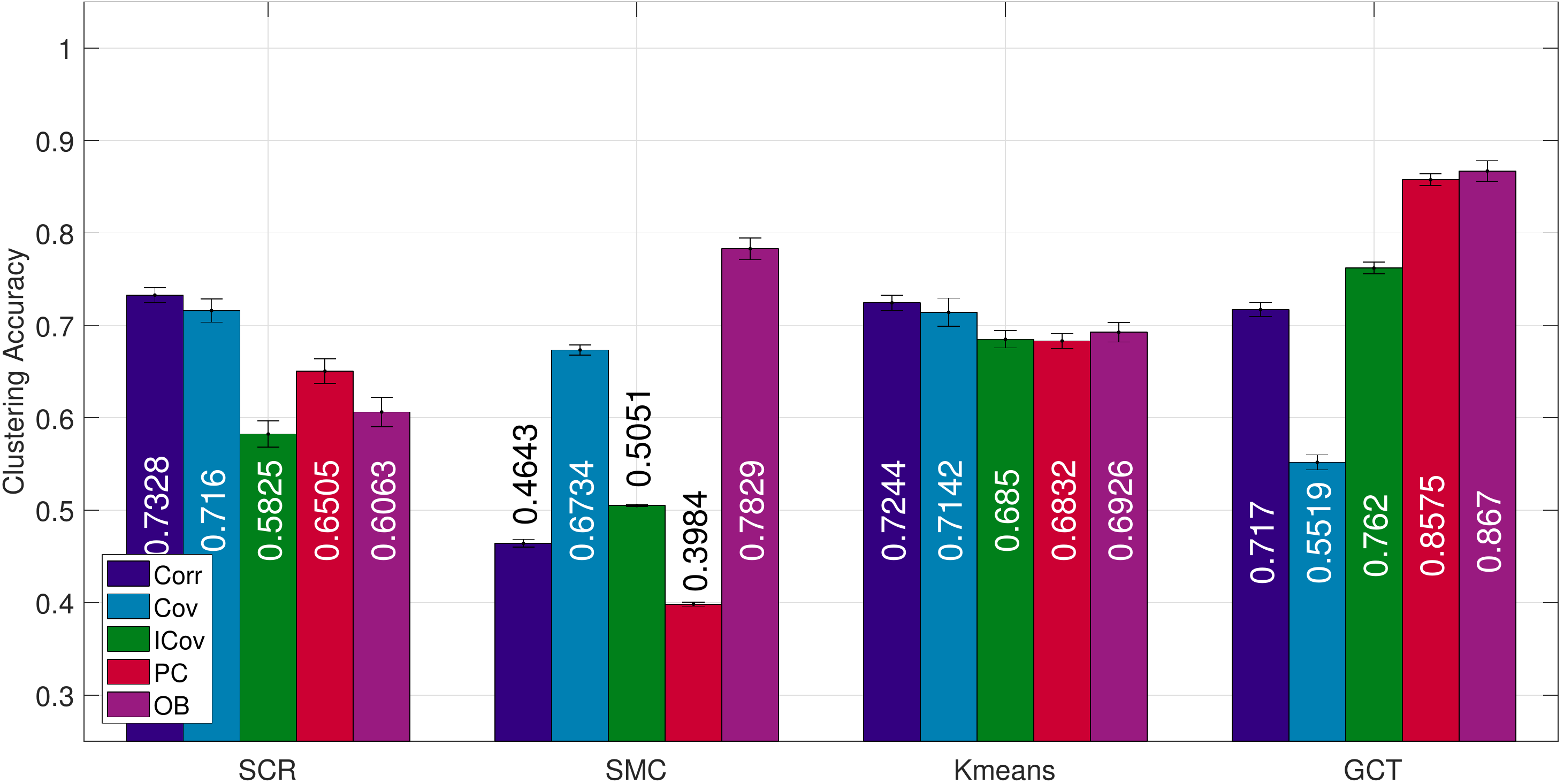}
    \caption{Linear kernel: $\tau_{\text{w}} = 50$;
      $N_{\text{NN}}^{\text{GCD}} =
      12$.}\label{fig:synth.linear.kernel.50.12} 
  \end{figure}

  \begin{figure}[!t]
    \centering
    \includegraphics[width=.8\linewidth]{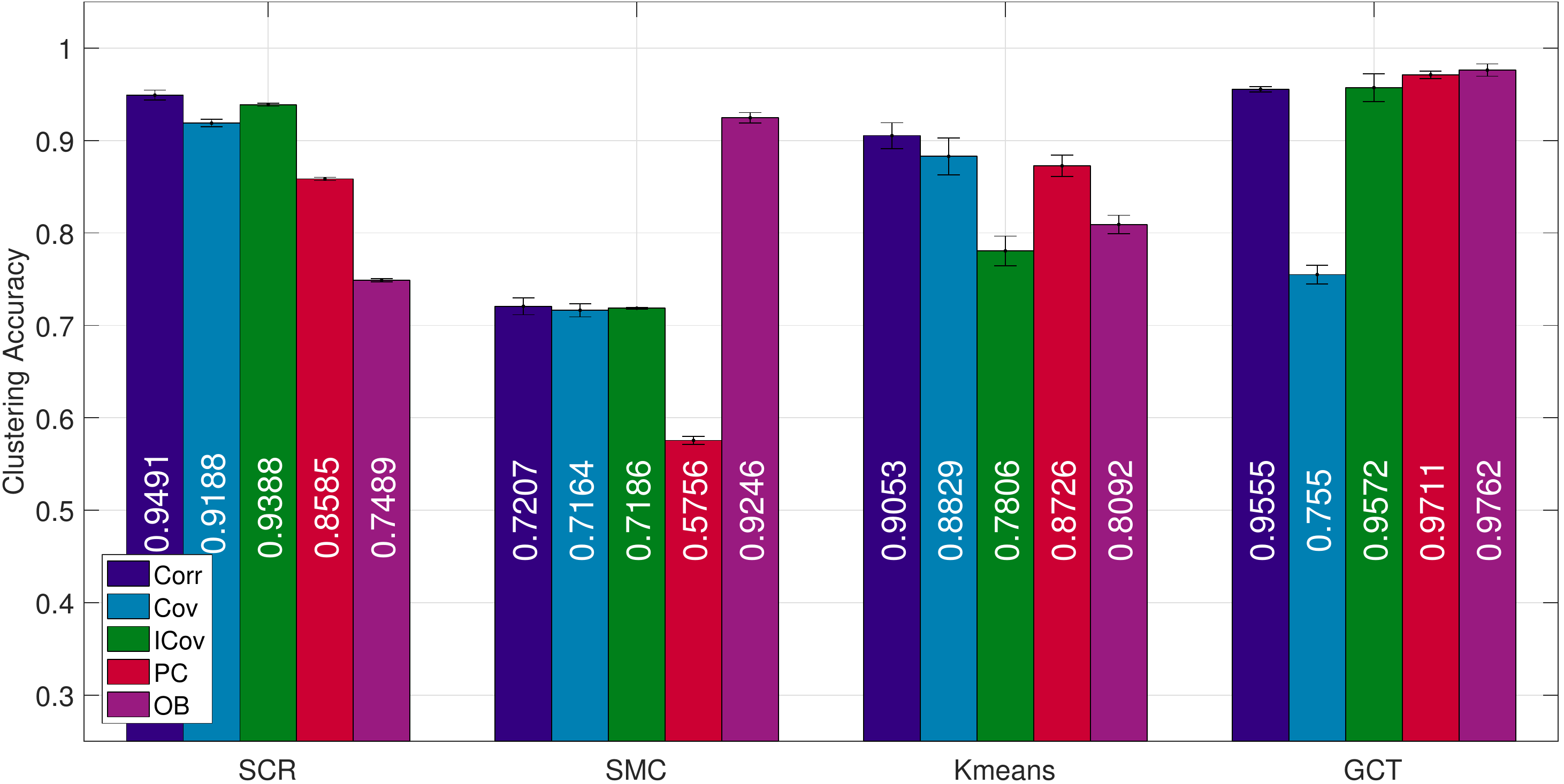}
    \caption{Linear kernel: $\tau_{\text{w}} = 70$;
      $N_{\text{NN}}^{\text{GCT}} =
      12$.}\label{fig:synth.linear.kernel.70.12} 
  \end{figure}

  \begin{figure}[!t]
    \centering
    \includegraphics[width=.8\linewidth]{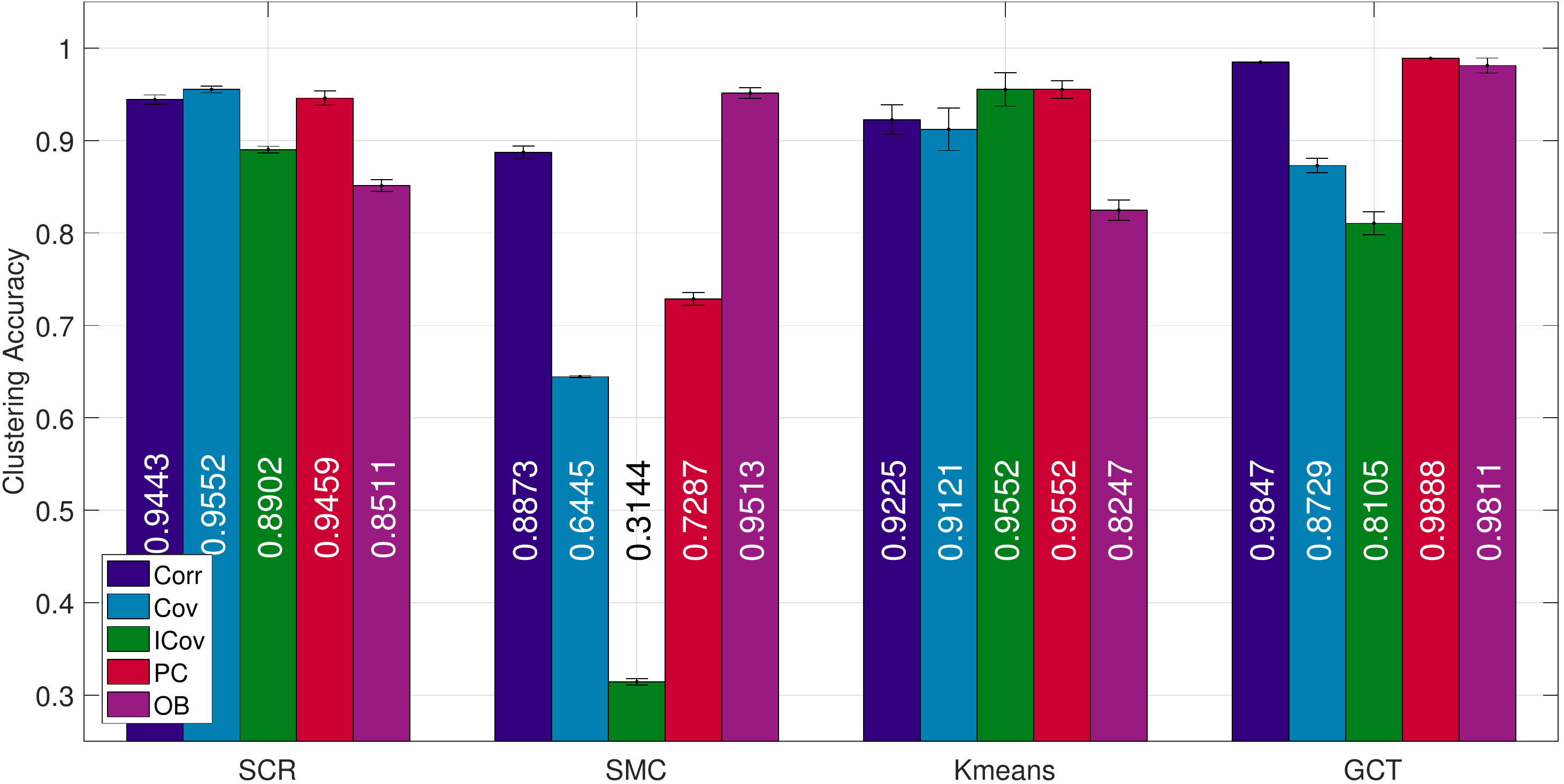}
    \caption{Linear kernel: $\tau_{\text{w}} = 80$;
      $N_{\text{NN}}^{\text{GCT}} = 8$.}
    \label{fig:synth.linear.kernel.80.8}
  \end{figure}

  \begin{figure}[!t]
    \centering
    \includegraphics[width=.8\linewidth]{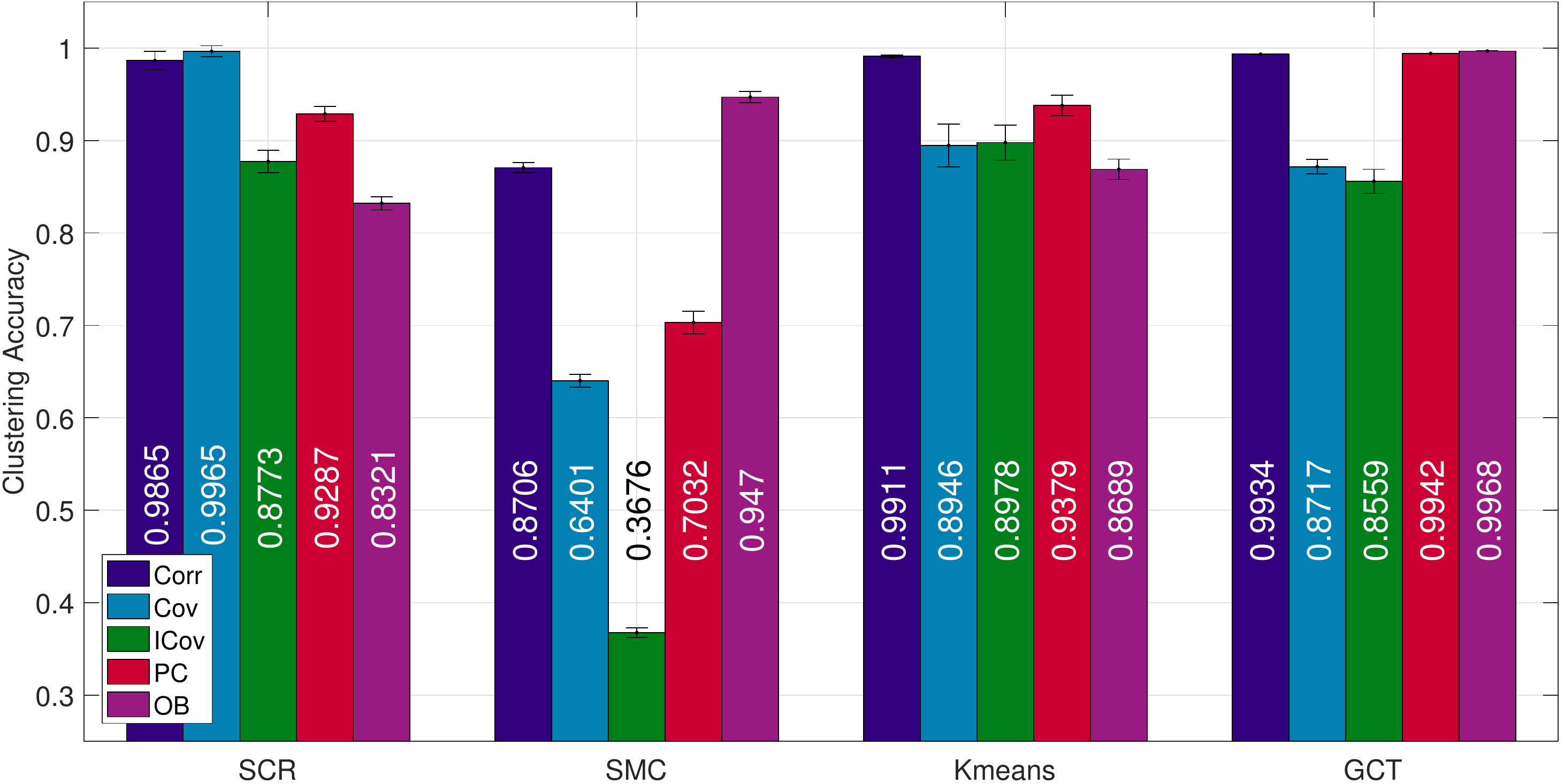}
    \caption{Linear kernel: $\tau_{\text{w}} = 80$;
      $N_{\text{NN}}^{\text{GCT}} = 12$.}
    \label{fig:synth.linear.kernel.80.12}
  \end{figure}

  \begin{figure}[!t]
    \centering
    \includegraphics[width=.8\linewidth]{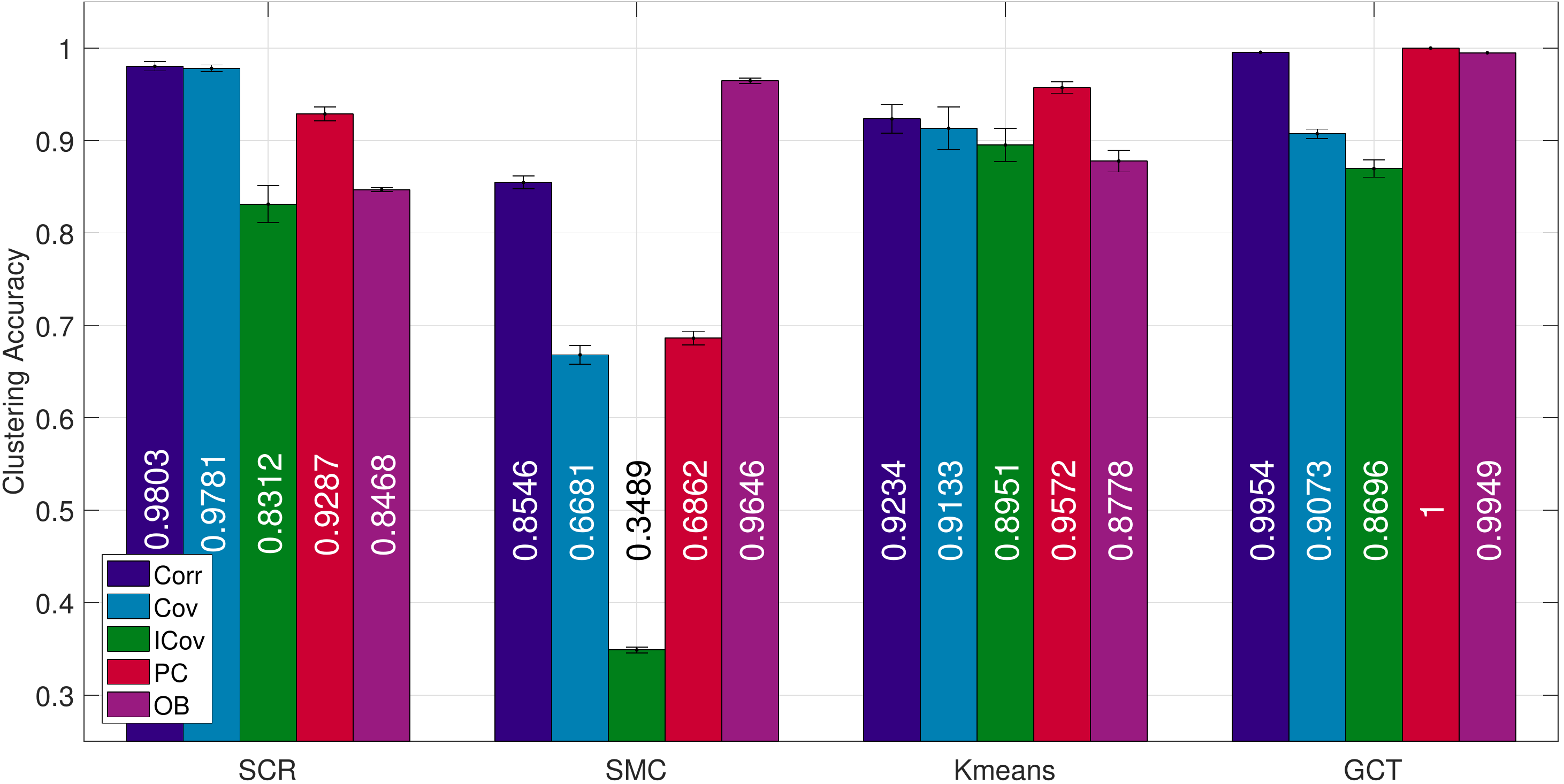}
    \caption{Linear kernel: $\tau_{\text{w}} =80$;
      $N_{\text{NN}}^{\text{GCT}} = 16$.}
    \label{fig:synth.linear.kernel.80.16}
  \end{figure}

\item \textbf{Single Gaussian kernel function:} The Gaussian
  (reproducing) kernel function $\kappa_{\sigma}$ of
  Appendix~\ref{app:RKHS} is used here, with variance values
  $\sigma^2\in \Set{0.5, 1, 2}$. As Appendix~\ref{app:RKHS}
  suggests, the feature space $\mathpzc{H}_{\sigma}$ becomes an
  infinite-dimensional functional space. Notice that the ``OB''
  tag is not included in
  Figs.~\ref{fig:synth.single.kernel.50.16.05}--%
  \ref{fig:synth.single.kernel.80.16.2}, since the methodology of
  Sec.~\ref{sec:ARMA} does not include any kernel-based
  arguments. As the relevant figures demonstrate, \textit{all}
  methods appear to be sensitive to the choice of the kernel's
  variance value: the less the value is, the worse the clustering
  accuracies become. Still, under such a uniform behavior, GCT
  exhibits the best performance among employed methods.

  \begin{figure}[!t]
    \centering
    \includegraphics[width=.8\linewidth]{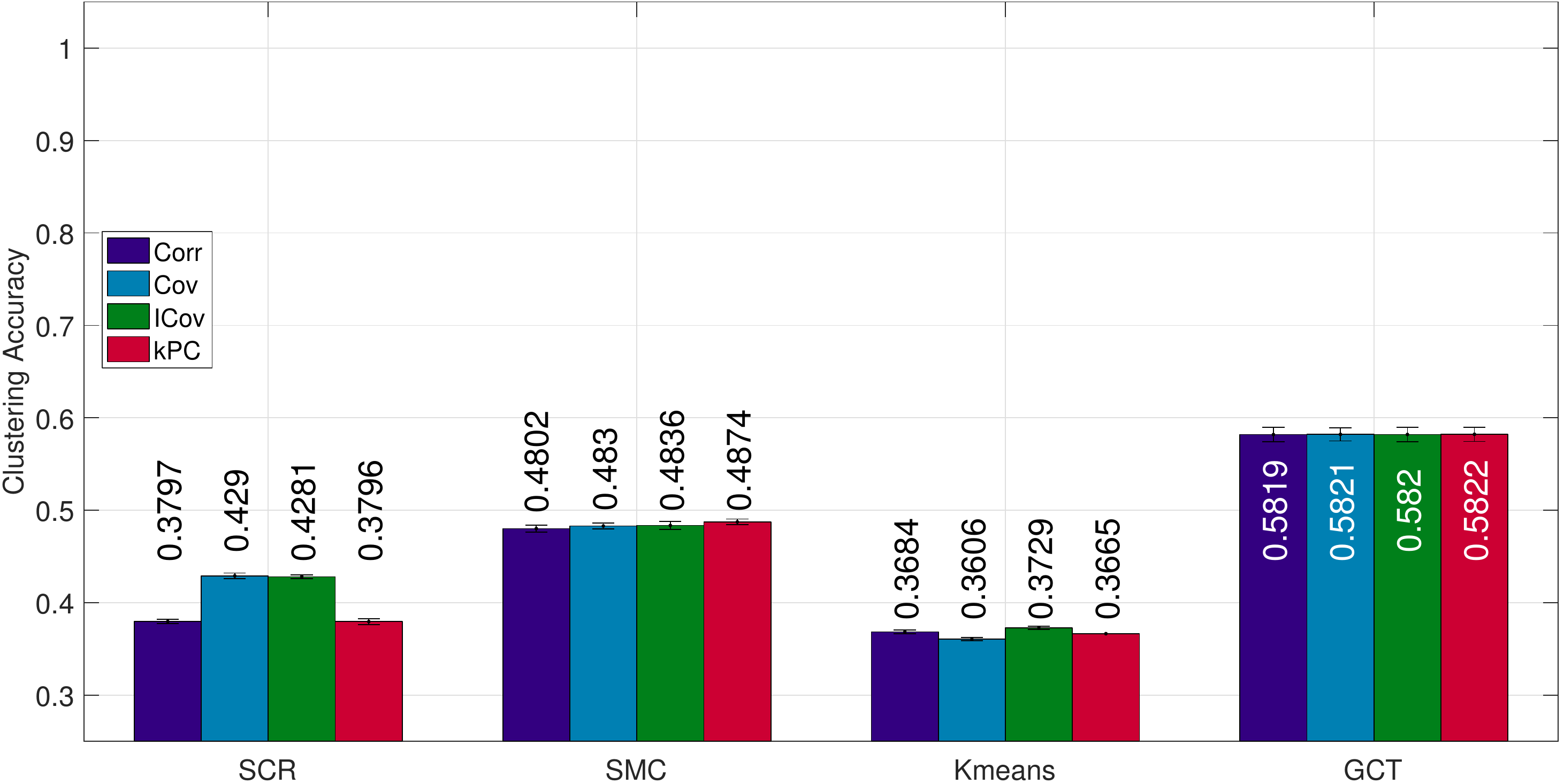}
    \caption{Single Gaussian kernel: $\sigma^2=0.5$;
      $\tau_{\text{w}} = 50$; $N_{\text{NN}}^{\text{GCT}} = 16$.}
    \label{fig:synth.single.kernel.50.16.05}
  \end{figure}

  \begin{figure}[!t]
    \centering
    \includegraphics[width=.8\linewidth]{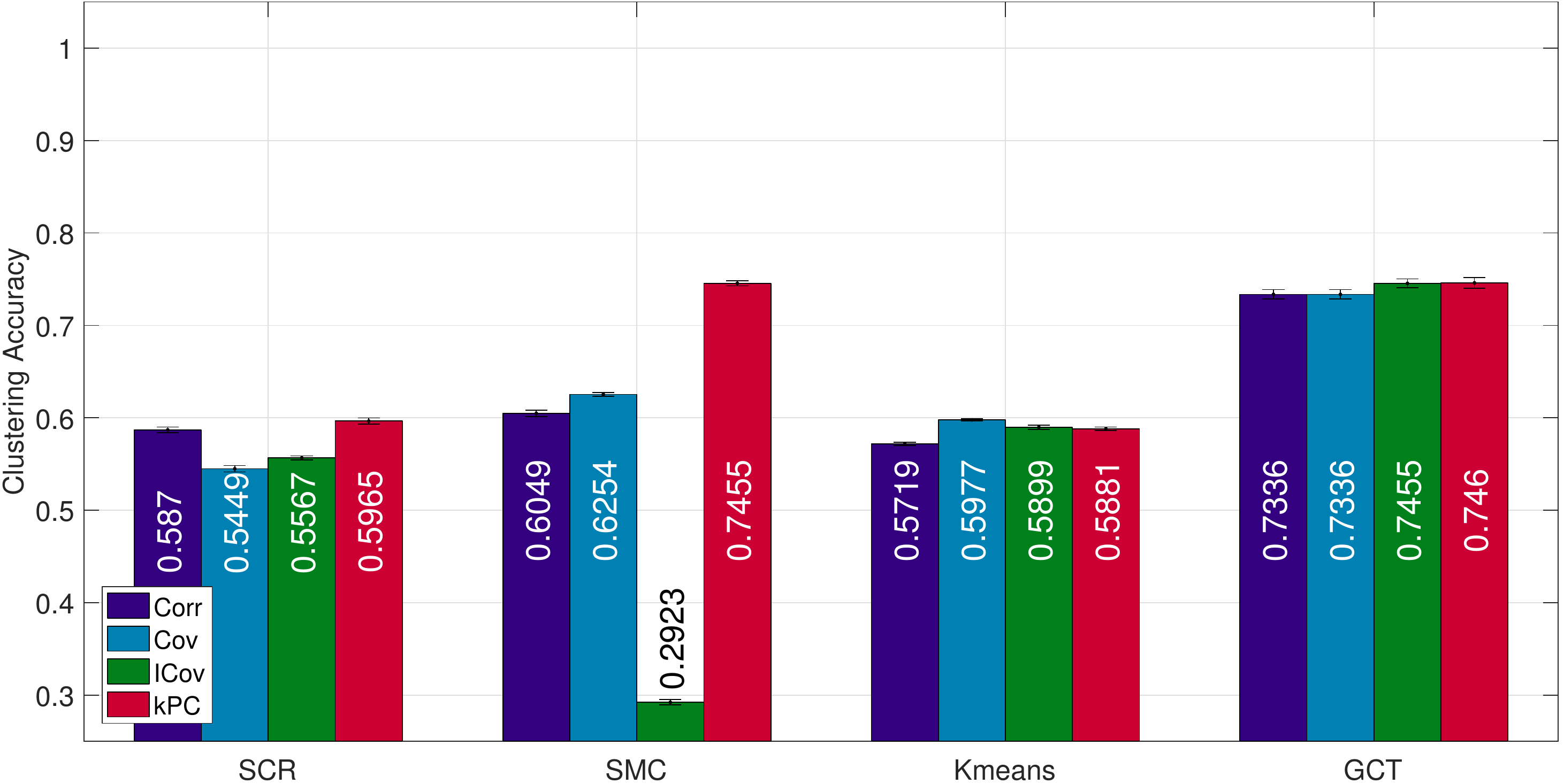}
    \caption{Single Gaussian kernel: $\sigma^2=1$;
      $\tau_{\text{w}} = 50$; $N_{\text{NN}}^{\text{GCT}} = 16$.}
    \label{fig:synth.single.kernel.50.16.1}
  \end{figure}

  \begin{figure}[!t]
    \centering
    \includegraphics[width=.8\linewidth]{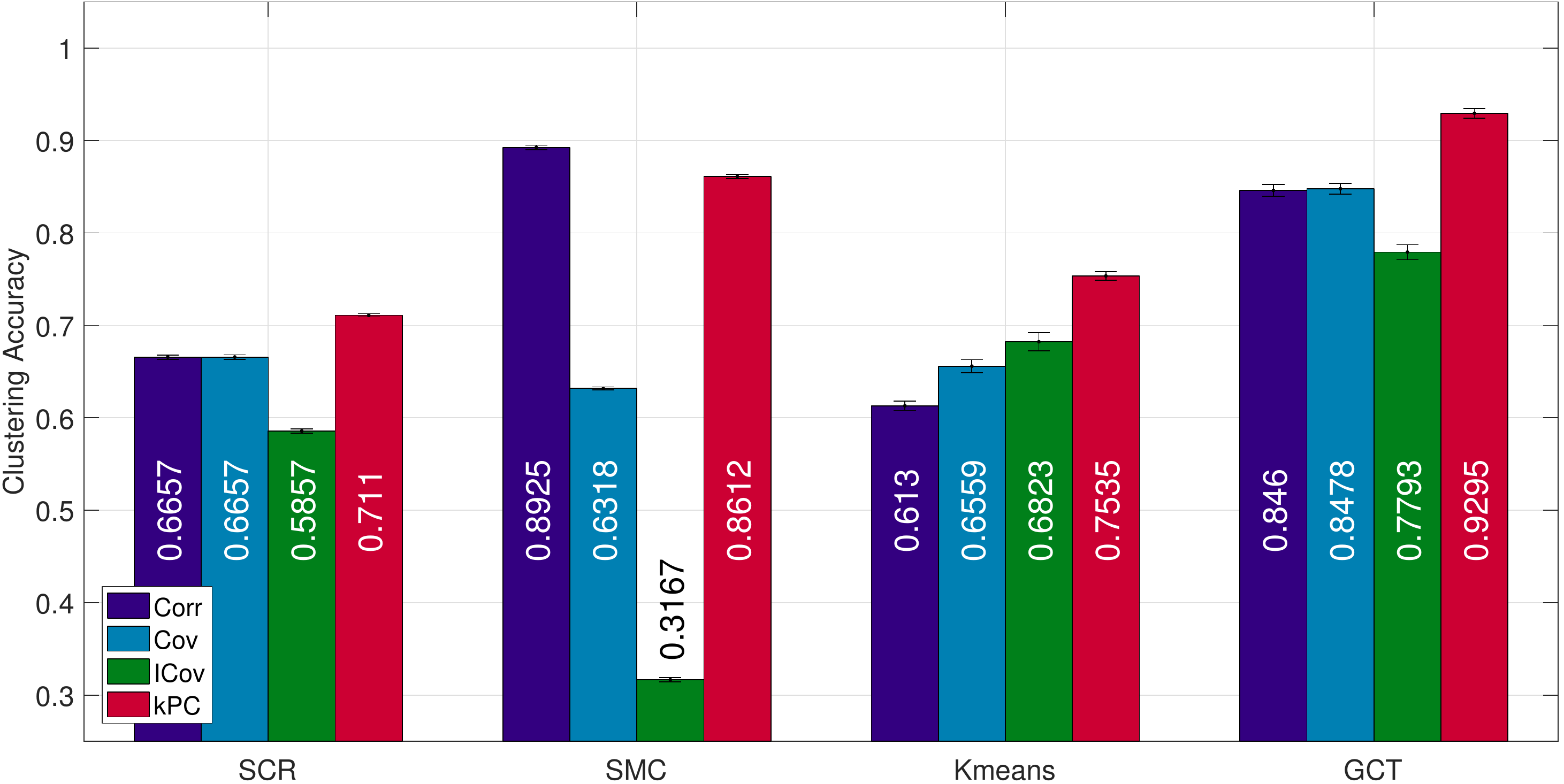}
    \caption{Single Gaussian kernel: $\sigma^2=2$;
      $\tau_{\text{w}} = 50$; $N_{\text{NN}}^{\text{GCT}} = 16$.}
    \label{fig:synth.single.kernel.50.16.2}
  \end{figure}

  \begin{figure}[!t]
    \centering
    \includegraphics[width=.8\linewidth]{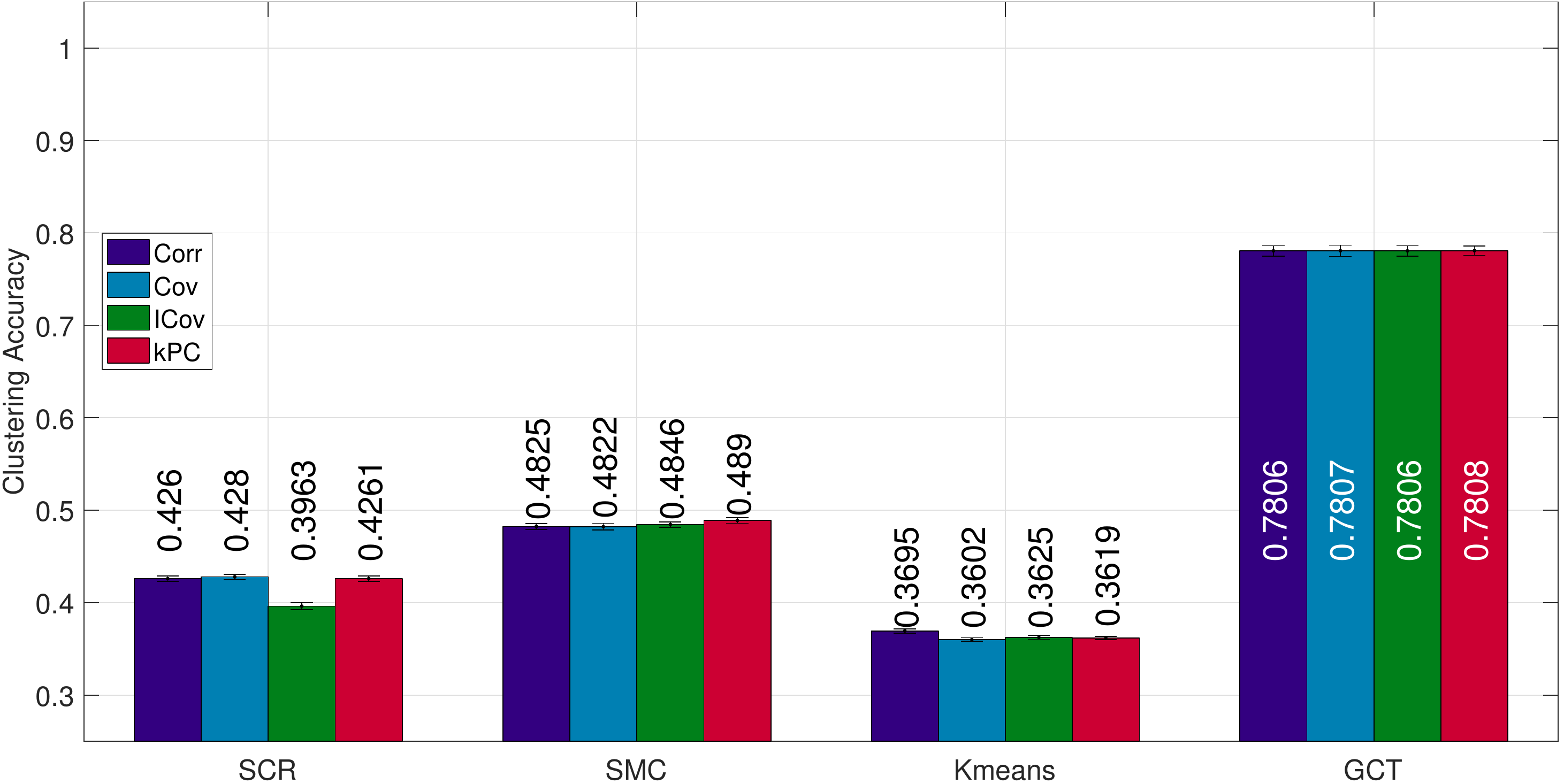}
    \caption{Single Gaussian kernel: $\sigma^2=0.5$;
      $\tau_{\text{w}} = 80$; $N_{\text{NN}}^{\text{GCT}} = 16$.}
    \label{fig:synth.single.kernel.80.16.05}
  \end{figure}

  \begin{figure}[!t]
    \centering
    \includegraphics[width=.8\linewidth]{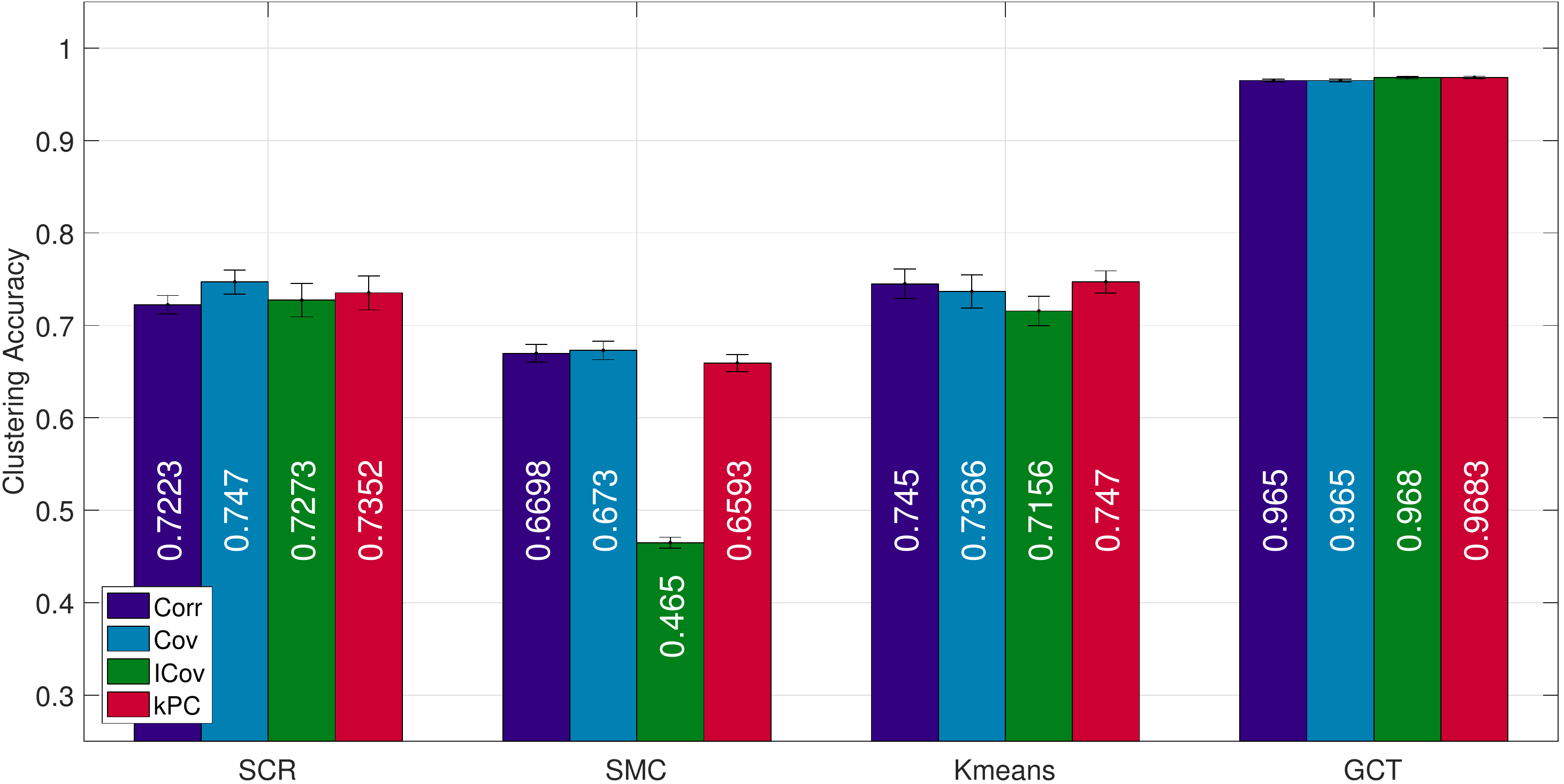}
    \caption{Single Gaussian kernel: $\sigma^2=1$;
      $\tau_{\text{w}} = 80$; $N_{\text{NN}}^{\text{GCT}} = 16$.}
    \label{fig:synth.single.kernel.80.16.1}
  \end{figure}

  \begin{figure}[!t]
    \centering
    \includegraphics[width=.8\linewidth]{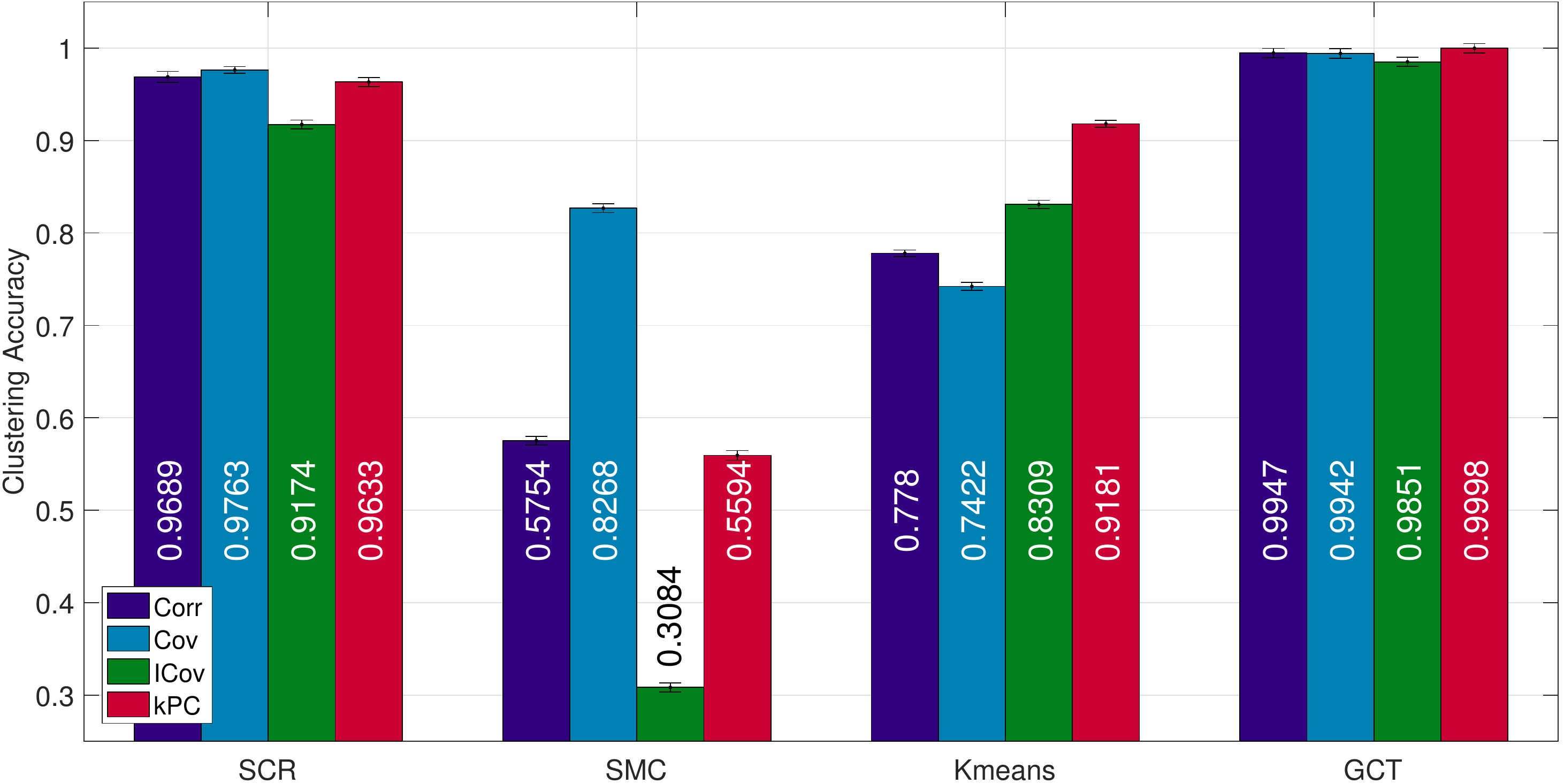}
    \caption{Single Gaussian kernel: $\sigma^2=2$;
      $\tau_{\text{w}} = 80$; $N_{\text{NN}}^{\text{GCT}} = 16$.}
    \label{fig:synth.single.kernel.80.16.2}
  \end{figure}

\item \textbf{Multi-kernel function:} As
  Figs.~\ref{fig:synth.single.kernel.50.16.05}--%
  \ref{fig:synth.single.kernel.80.16.2} demonstrate, the choice
  of the value of variance of a Gaussian kernel hinders the
  clustering-accuracy performance of all employed techniques. To
  this end, a multi-kernel-function approach is adopted here to
  robustify all methods:
  $\kappa := (1/I) \sum_{i=1}^I \kappa_{\sigma_i}$, where the
  values of variances cover the wide range
  $\sigma_i\in \Set{0.25 + 0.01(i-1)}_{i=1}^{I}$, with $I:= 376$,
  $\sigma_1 = 0.25$, and $\sigma_I = 4$. It can be easily
  verified that $\kappa$ is a reproducing kernel
  (\cf~Appendix~\ref{app:RKHS}). Moreover, the resulting feature
  space $\mathpzc{H}$ is an infinite-dimensional functional
  space. Needless to say that there are numerous ways of defining
  similar multi-kernel functions, such as the incorporation of
  polynomial or linear kernels in $\kappa$. Since this study is
  not meant to be exhaustive, such a path is not
  pursued. Figs.~\ref{fig:synth.multi.kernel.50.12}--%
  \ref{fig:synth.multi.kernel.80.16} show results for several
  values of sliding-window length $\tau_{\text{w}}$ and
  $N_{\text{NN}}^{\text{GCT}}$. As expected, multi-kernel
  functions enhance performance of all methods, with GCT
  exhibiting the best performance among employed
  techniques.

  \begin{figure}[!t]
    \centering
    \includegraphics[width=.8\linewidth]{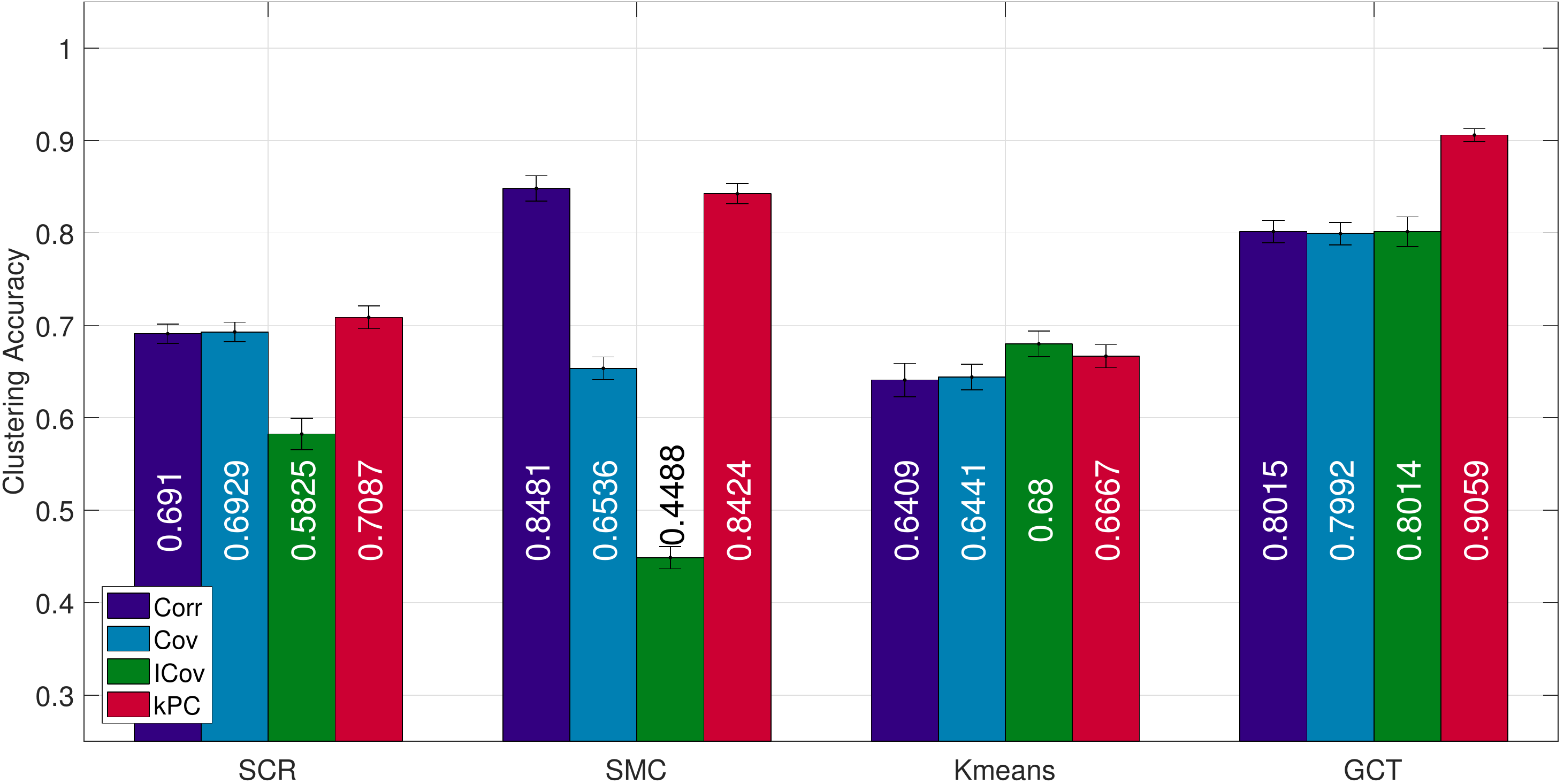}
    \caption{Multi-kernel: $\tau_{\text{w}} = 50$;
      $N_{\text{NN}}^{\text{GCT}} = 12$.} 
    \label{fig:synth.multi.kernel.50.12}
  \end{figure}

  \begin{figure}[!t]
    \centering
    \includegraphics[width=.8\linewidth]{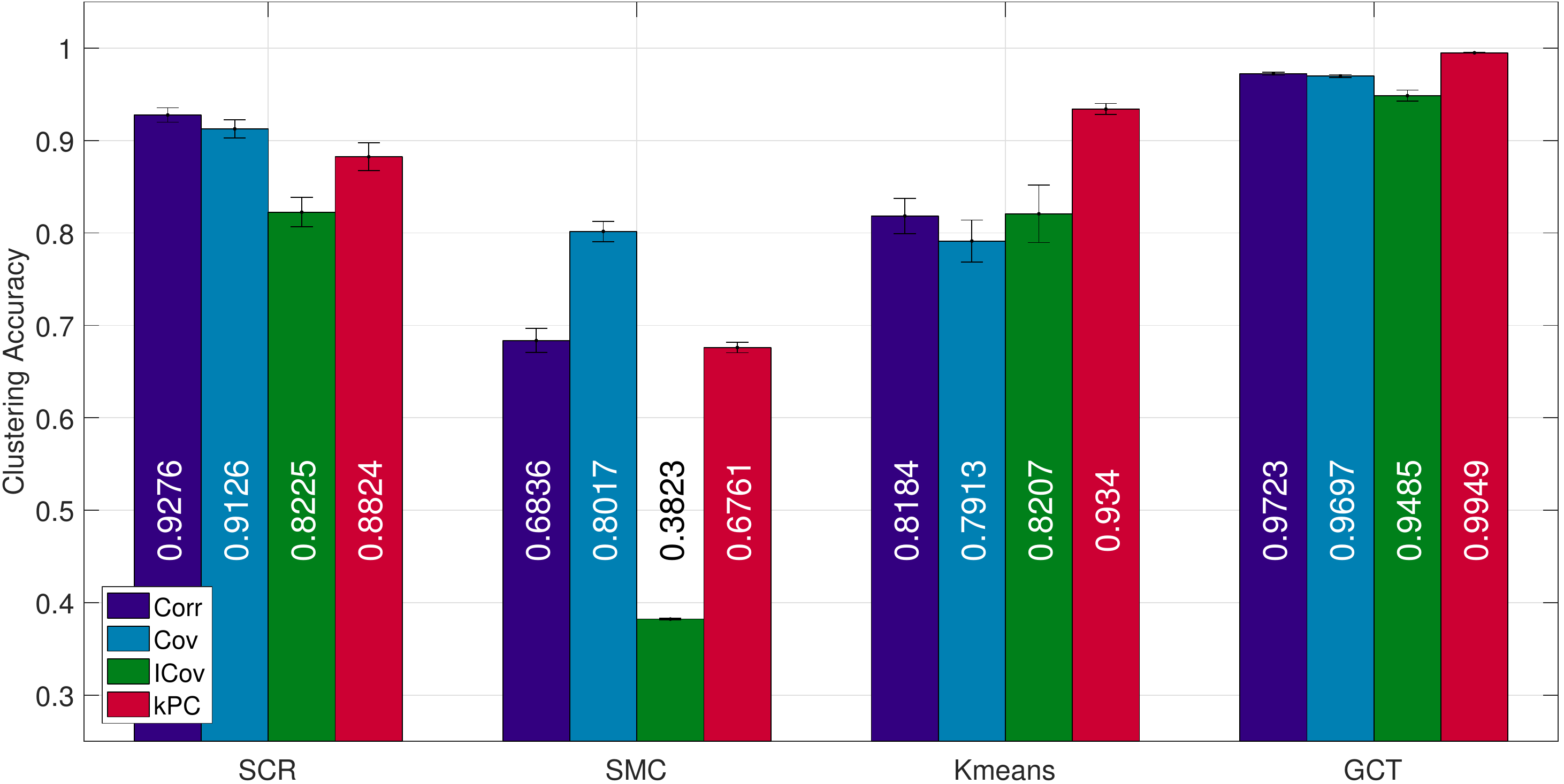}
    \caption{Multi-kernel: $\tau_{\text{w}} = 70$;
      $N_{\text{NN}}^{\text{GCT}} = 12$.} 
    \label{fig:synth.multi.kernel.70.12}
  \end{figure}

  \begin{figure}[!t]
    \centering
    \includegraphics[width=.8\linewidth]{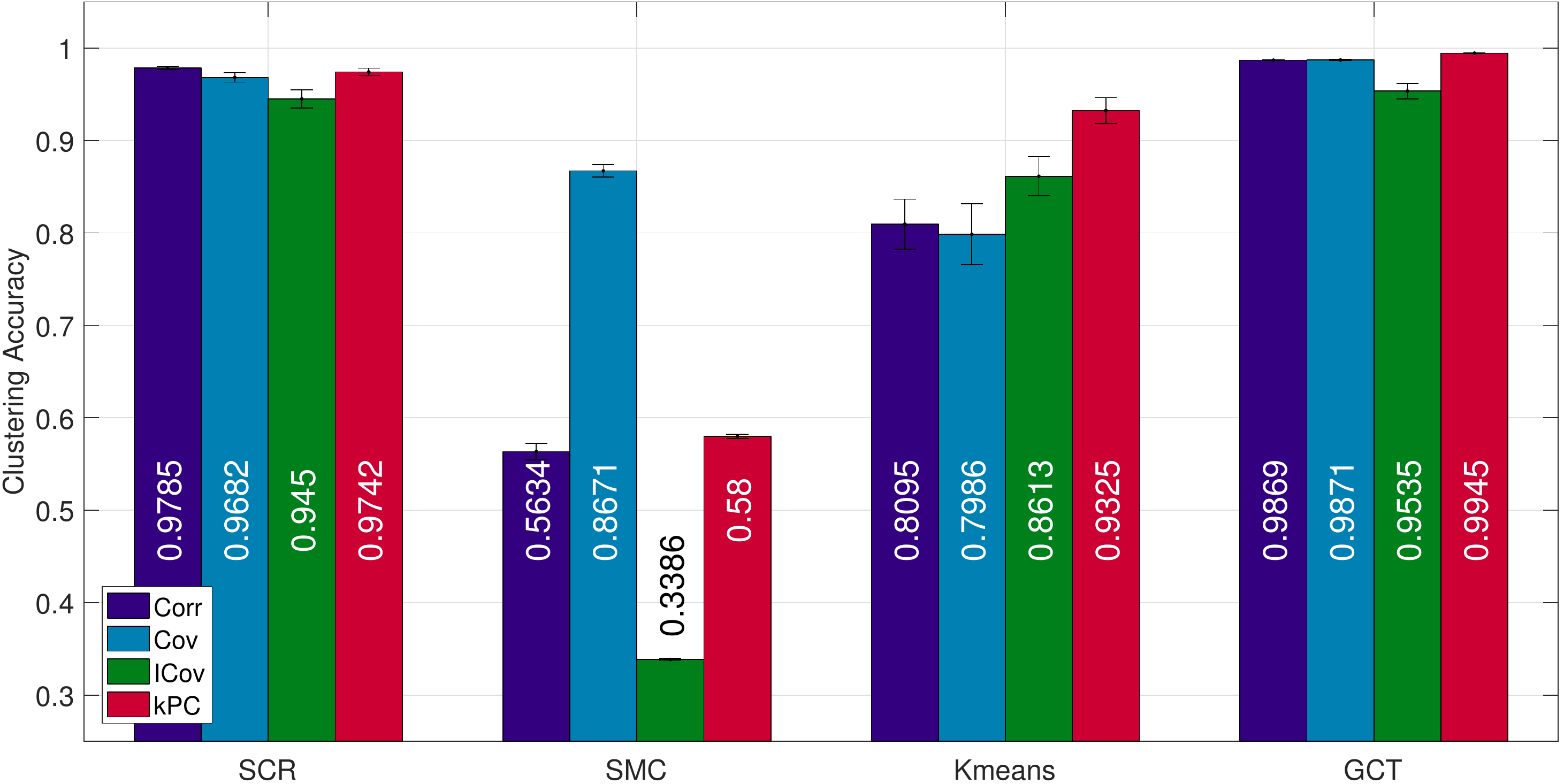}
    \caption{Multi-kernel: $\tau_{\text{w}} = 80$;
      $N_{\text{NN}}^{\text{GCT}} = 8$.} 
    \label{fig:synth.multi.kernel.80.8}
  \end{figure}

  \begin{figure}[!t]
    \centering
    \includegraphics[width=.8\linewidth]{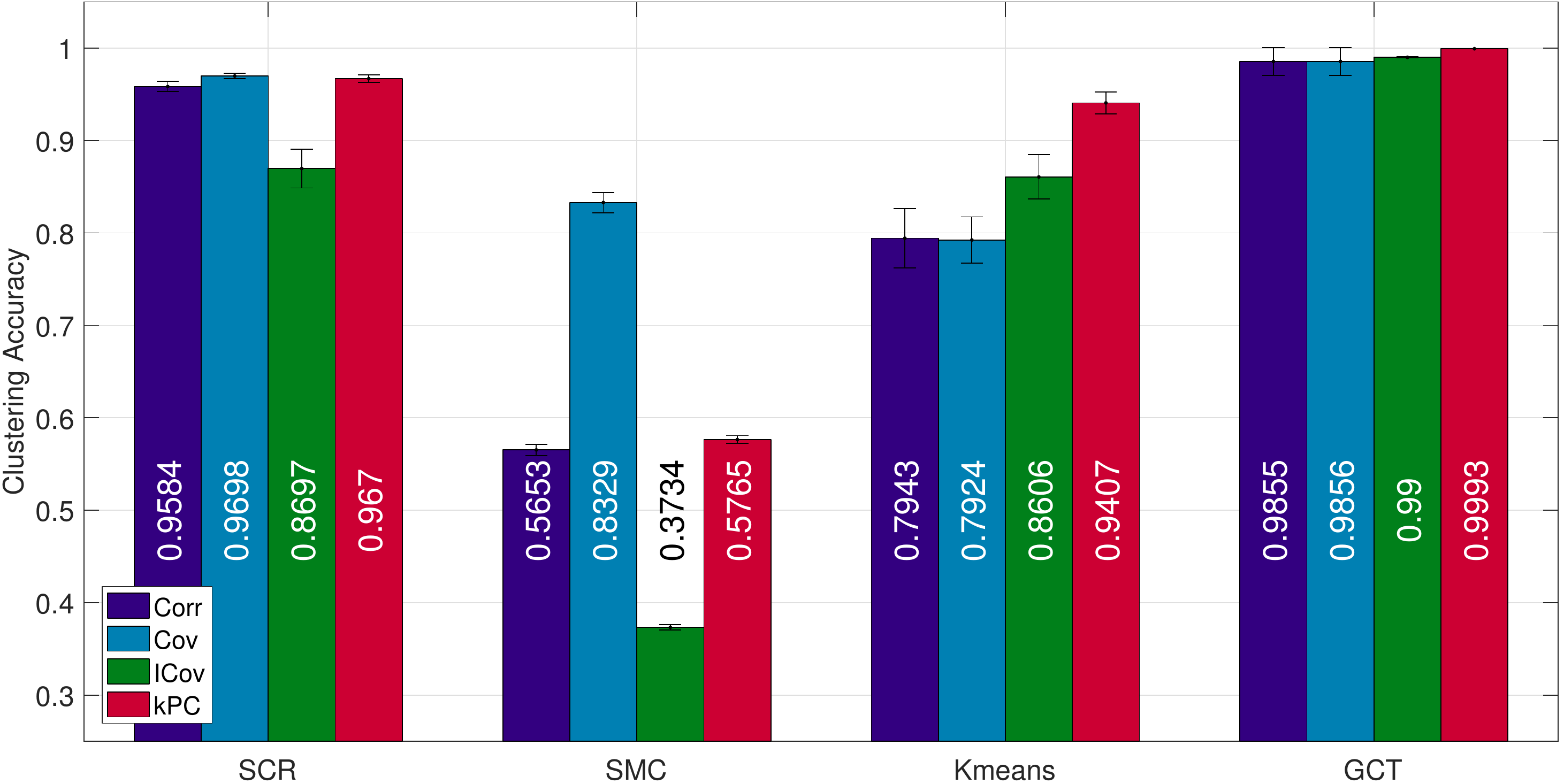}
    \caption{Multi-kernel: $\tau_{\text{w}} = 80$;
      $N_{\text{NN}}^{\text{GCT}} = 12$.} 
    \label{fig:synth.multi.kernel.80.12}
  \end{figure}

  \begin{figure}[!t]
    \centering 
    \includegraphics[width=.8\linewidth]{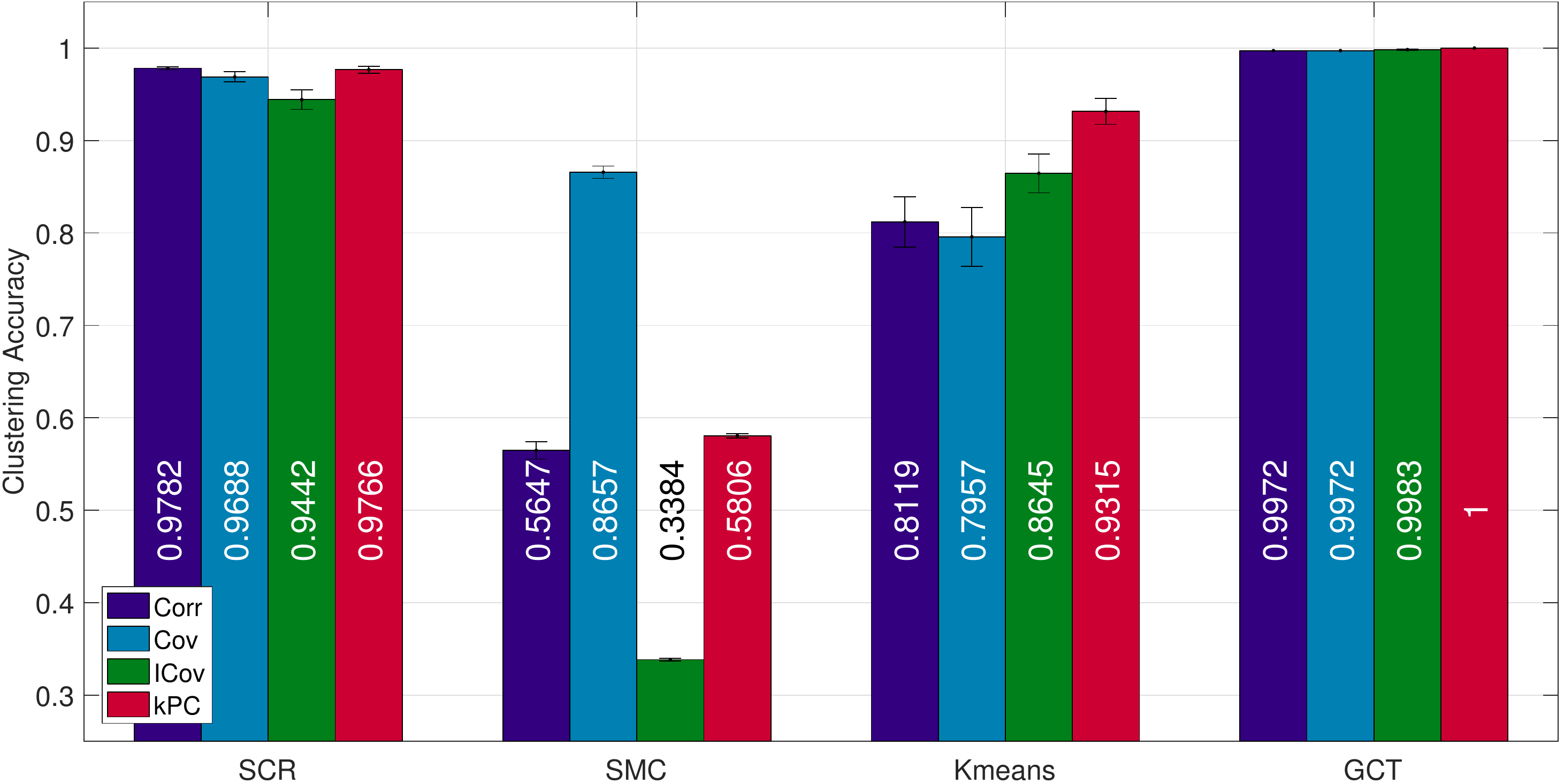}
    \caption{Multi-kernel: $\tau_{\text{w}} = 80$;
      $N_{\text{NN}}^{\text{GCT}} = 16$.} 
    \label{fig:synth.multi.kernel.80.16}
  \end{figure}

\item \textbf{SDE:} Here, the kernel function is designed via the
  data-driven approach of Sec.~\ref{sec:SDE}, and results are
  demonstrated in Figs.~\ref{fig:synth.SDE.10.3}--%
  \ref{fig:synth.SDE.50.20}. To be able to vary meaningfully the
  neighborhood sizes $\Set{N^{\text{SDE}}_{\nu t}}$, needed as
  parameters in SDE, the brain-network size $N_{\mathpzc{G}}$
  took the values of $10$ and $50$. As
  Figs.~\ref{fig:synth.SDE.10.3}%
  --\ref{fig:synth.SDE.50.20} exhibit, the larger the network and
  SDE-neighborhood size are, the better SDE performs.

  \begin{figure}[!t]
    \centering
    \includegraphics[width=.8\linewidth]{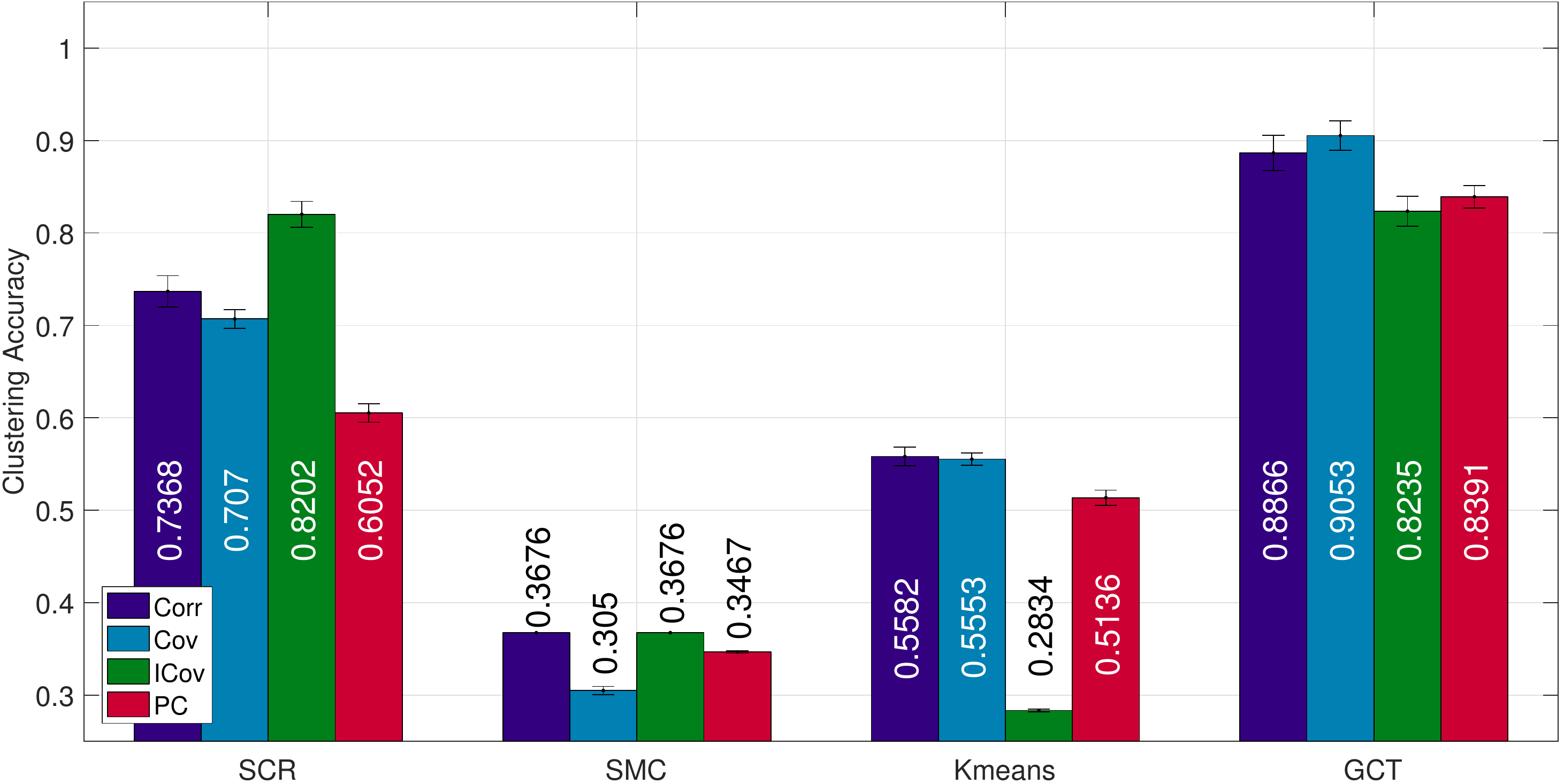}
    \caption{SDE: $N^{\text{SDE}}_{\nu t} = 3$;
      $N_{\mathpzc{G}} = 10$; $N_{\text{NN}}^{\text{GCT}}=16$.}
    \label{fig:synth.SDE.10.3}
  \end{figure}

  \begin{figure}[!t]
    \centering
    \includegraphics[width=.8\linewidth]{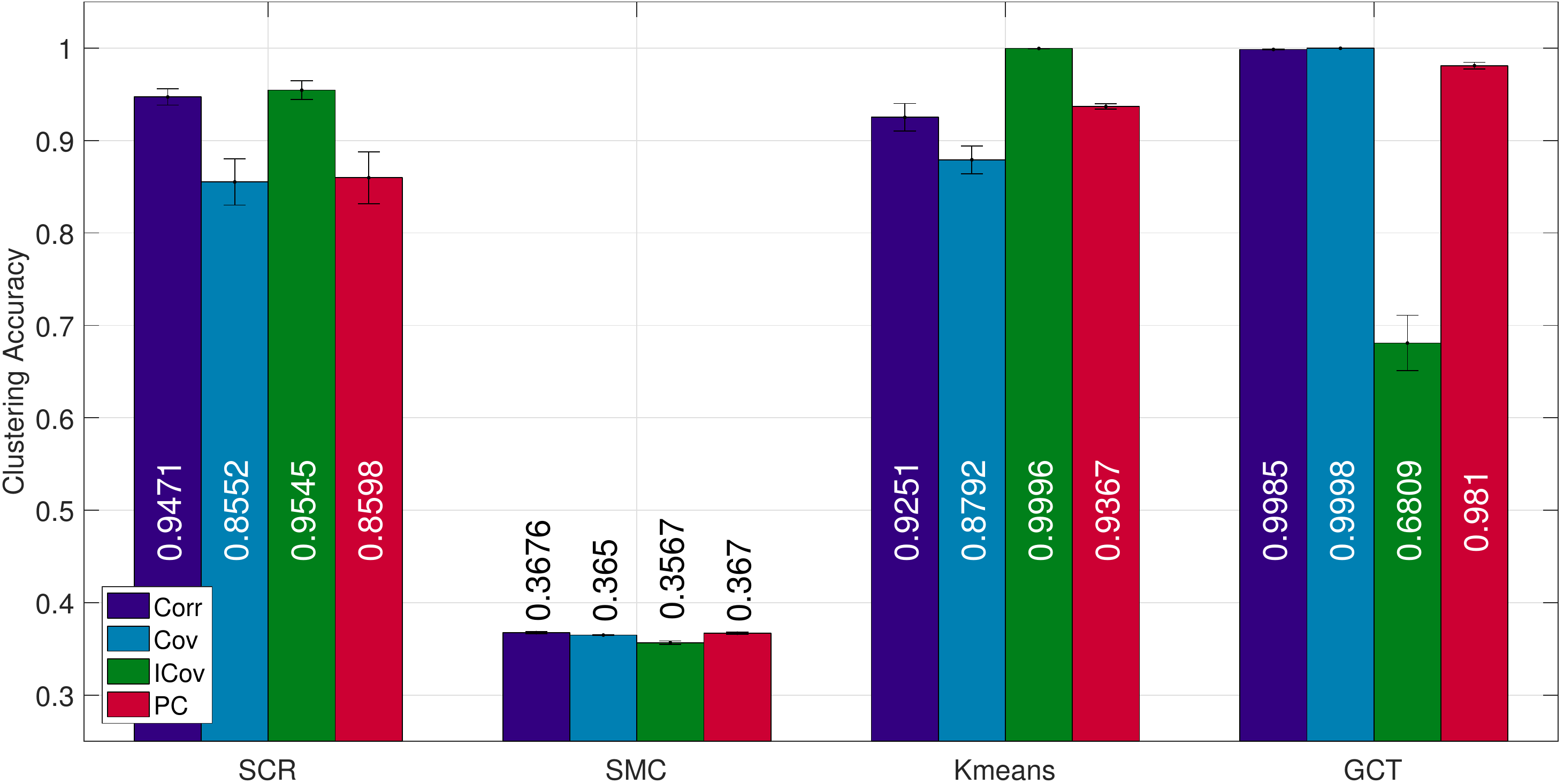}
    \caption{SDE: $N^{\text{SDE}}_{\nu t} = 10$;
      $N_{\mathpzc{G}} = 50$; $N_{\text{NN}}^{\text{GCT}} = 16$.}
    \label{fig:synth.SDE.50.10}
  \end{figure}

  \begin{figure}[!t]
    \centering
    \includegraphics[width=.8\linewidth]{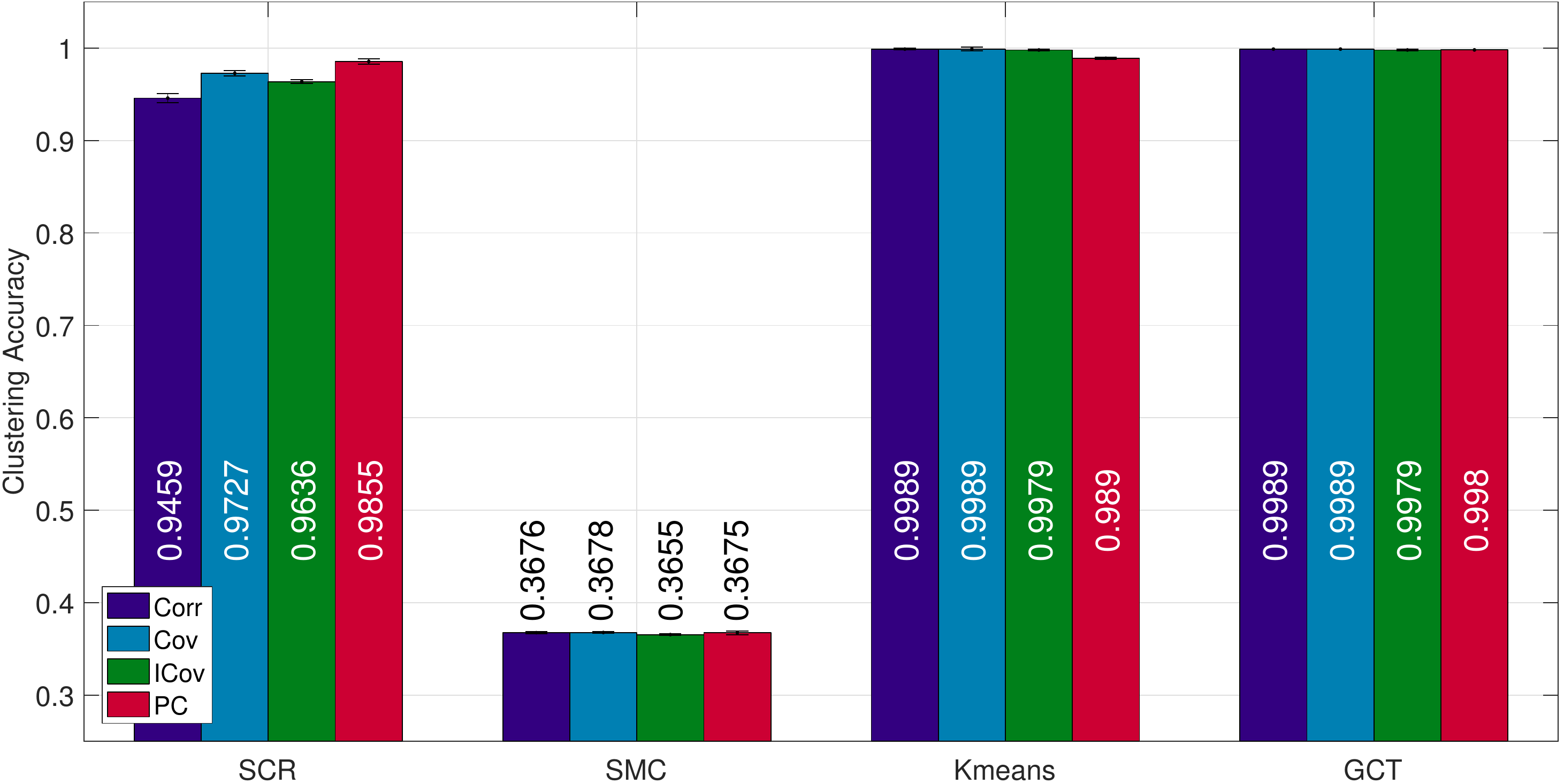}
    \caption{SDE: $N^{\text{SDE}}_{\nu t} = 20$;
      $N_{\mathpzc{G}} = 50$; $N_{\text{NN}}^{\text{GCT}} = 16$.}
    \label{fig:synth.SDE.50.20}
  \end{figure}

\end{enumerate}

The best clustering-accuracy result among
Figs.~\ref{fig:synth.linear.kernel.50.12}--%
\ref{fig:synth.SDE.50.20} is recorded for GCT in
Figs.~\ref{fig:synth.linear.kernel.80.16} and
\ref{fig:synth.multi.kernel.80.16}, with a value of $1$ for the
``kPC'' feature.

\subsection{Real-data-driven time series}\label{sec:real.data}

The brain activity analyzed in this section was obtained by the
spatially embedded nonlinear model
of~\cite{Muldoon.stimulation.16}, and the structural brain
networks derived from diffusion spectrum imaging (DSI) of the
data collected from $4$ healthy adult subjects. All subjects
volunteered with informed consent in writing and in accordance
with the Institutional Review Board/Human Subjects Committee,
Univ.~of~California, Santa~Barbara.

\begin{figure}[!t]
  \centering
  \subfloat[Subject 1]
  {\includegraphics[width=.3\linewidth]{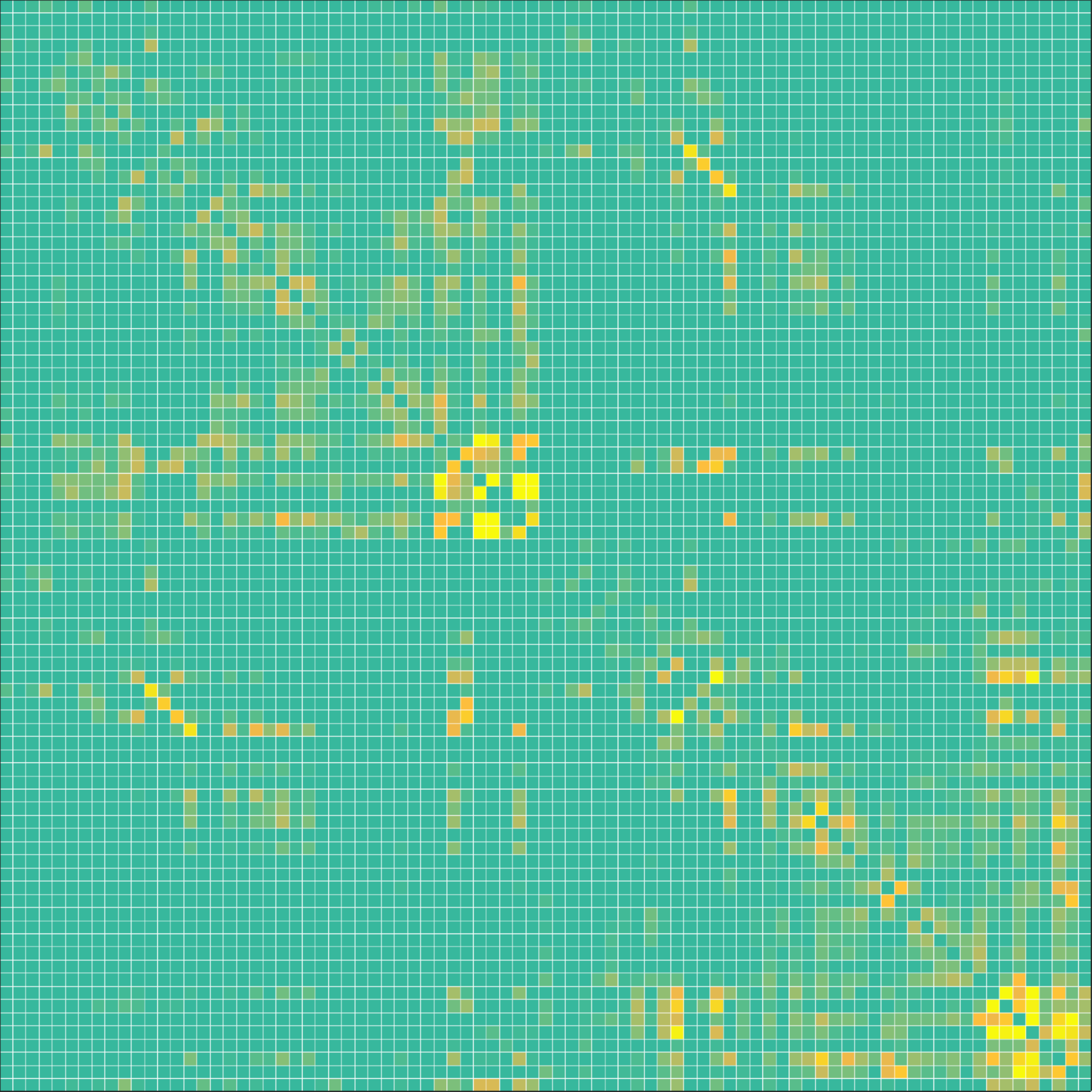}
    \label{fig:USAF.Conn.1}}
  \subfloat[Subject 2]
  {\includegraphics[width=.3\linewidth]{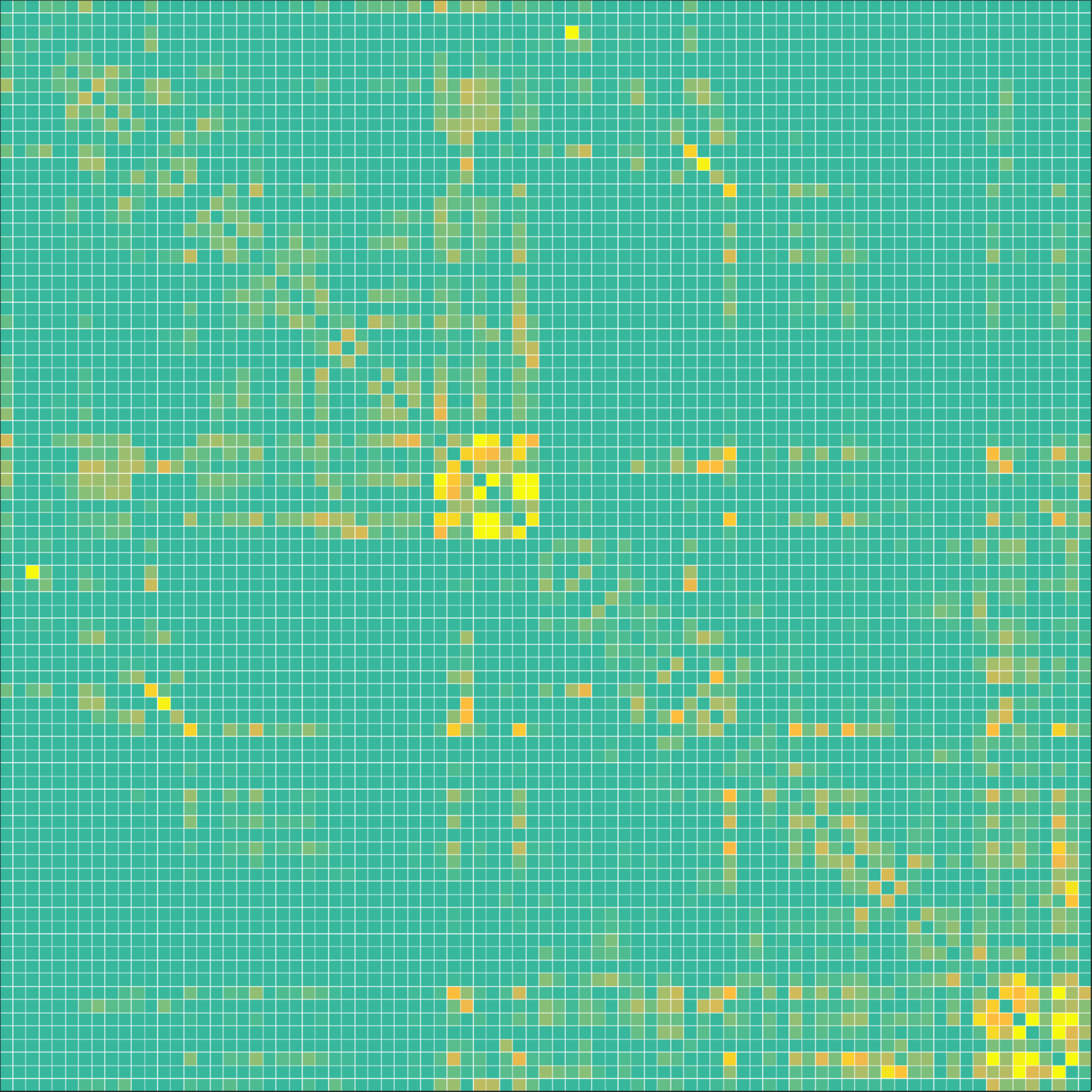}
    \label{fig:USAF.Conn.2}}\\
  \subfloat[Subject 3]
  {\includegraphics[width=.3\linewidth]{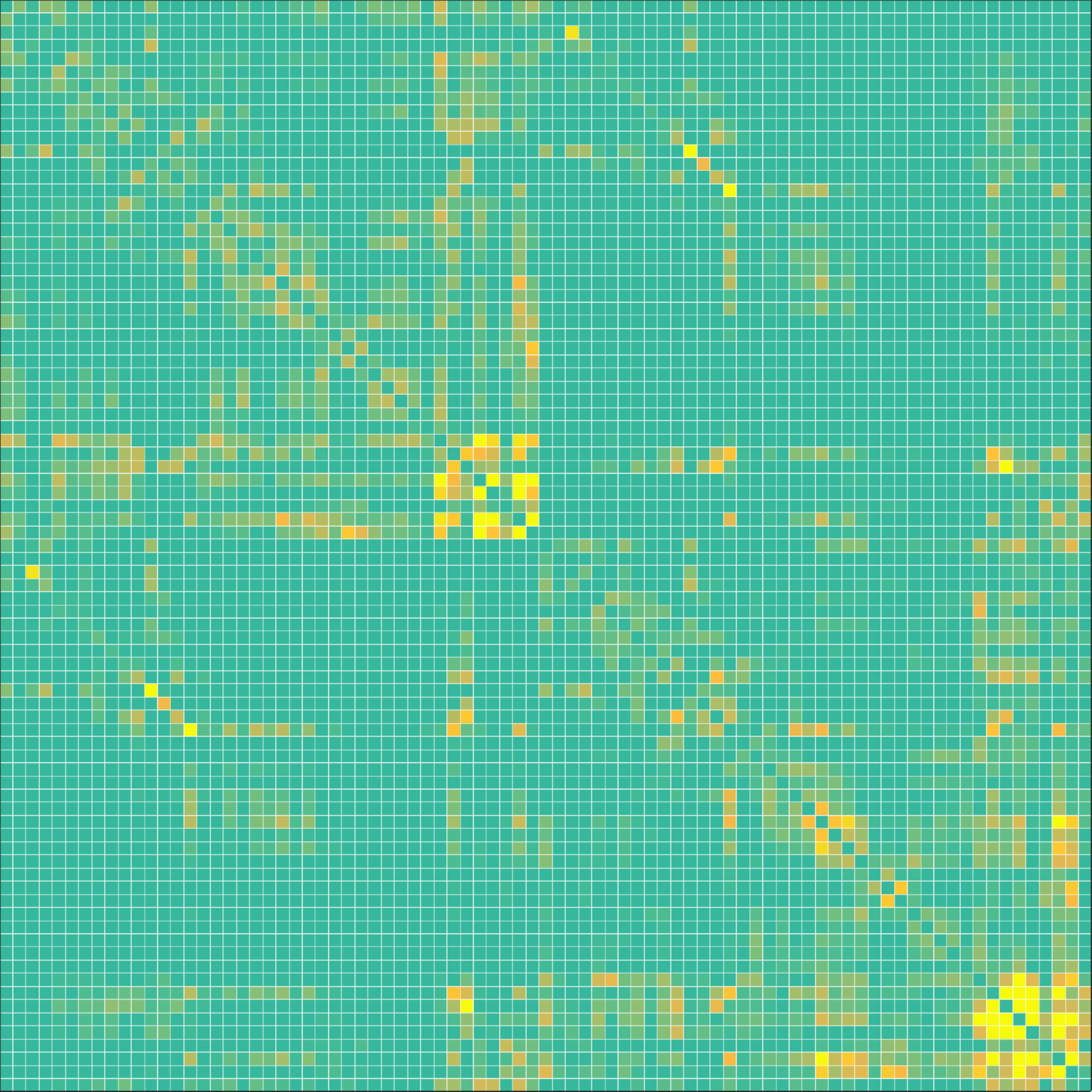}
    \label{fig:USAF.Conn.3}}
  \subfloat[Subject 4]
  {\includegraphics[width=.3\linewidth]{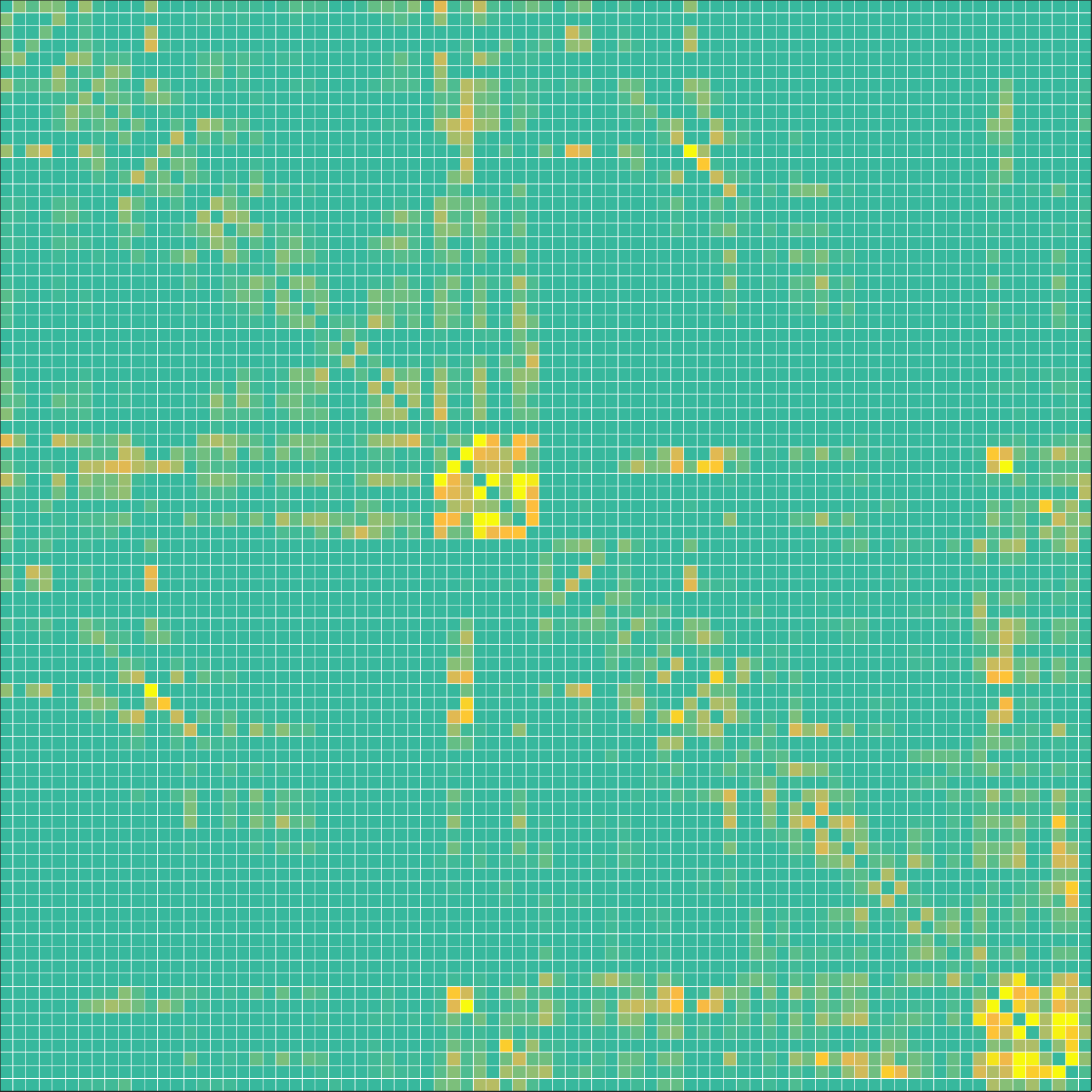}
    \label{fig:USAF.Conn.4}}
  \caption{Real-data structural weighted adjacency
    matrices.}\label{fig:USAF.Conn.Mtrx}
\end{figure}

As described fully in~\cite{Muldoon.stimulation.16}, diffusion
tractography was used to estimate the number of streamlines
linking a number $N_{\mathpzc{G}} = 83$ of large-scale cortical
and subcortical regions extracted from the Lausanne
atlas~\cite{Hagmann.mapping.08}. The number of streamlines
connecting two regions was normalized by the sum of the volumes
of the regions, resulting in the weighted adjacency matrix
$\vect{B} = [b_{\nu\nu'}]$, where $b_{\nu\nu'}$ reflects the density of
streamlines connecting the $\nu$th and $\nu'$th brain regions
(Fig.~\ref{fig:USAF.Conn.Mtrx}). Additionally, the spatial
distance between two brain regions was used to estimate the
signal transmission time, assuming a signal propagation speed of
$8\text{m}/\text{sec}$.

Regional brain activity (EEG-type time series) was modeled using
biologically motivated nonlinear Wilson-Cowan
oscillators~\cite{Wilson.excitatory.72,
  Wilson.math.73}. Wilson-Cowan oscillators represent the
mean-field dynamics of a spatially localized population of
neurons, modeled through equations governing the firing rate of
excitatory, $y_{\nu t}$, and inhibitory, $x_{\nu t}$, neuronal
populations. As in \cite{Muldoon.stimulation.16}, single
Wilson-Cowan oscillators are linked as follows, via the
individual's adjacency and delay matrices which are unique for
each of the four subjects:
\begin{subequations}\label{Sarah-Henry.model}
  \begin{align}
    \tfrac{dy_{\nu t}}{dt}
    && = & -\alpha y_{\nu t} + \eta_{yt} \notag\\
    &&& +\tfrac{0.9945-y_{\nu t}}{8} f_{y} \Bigl( \gamma_1 y_{\nu
        t} - \gamma_2 x_{\nu t} \notag\\
    &&& \hphantom{\frac{0.9945-y_{\nu t}}{8} f_{y}} + \gamma_5
        \sum\nolimits_{\nu'=1}^{N_{\mathpzc{G}}} b_{\nu\nu'}
        y_{\nu'}(t-d_{\nu\nu'}) \notag\\
    &&& \hphantom{\frac{0.9945-y_{\nu t}}{8} f_{y}}
        + \mu_{\nu t}\Bigr)\,, \label{eq:exc}\\
    \tfrac{dx_{\nu t}}{dt}
    && = & -\alpha x_{\nu t} + \eta_{xt} \notag\\
    &&& + \tfrac{0.9994-x_{\nu t}}{8}
        f_{x} \bigl(\gamma_3 y_{\nu t} -\gamma_4 x_{\nu t}
        \bigr)\,, \label{eq:inh}\\
    f_{z} (q)
    && := & \tfrac{1}{1+e^{-\zeta_z(q -
            \theta_z)}} - \tfrac{1}{1+e^{\zeta_z \theta_z}},
            \quad z \in \Set{x,y}\,, \label{eq:fn}
  \end{align}
\end{subequations}
where $\eta_{zt}$ is a realization of a Gaussian random variables
with mean $0$ and variance $\sigma^2$, per $t$ and
$z\in\Set{x,y}$. The external stimulation input is set equal to
$\mu_{\nu t} := 1.25$, if $\nu = 1$, and $\mu_{\nu t} := 0$, if
$\nu \neq 1$, $\forall t$. Parameters
$(\alpha, \gamma_1, \ldots, \gamma_5, \sigma^2, \zeta_x,
\theta_x, \zeta_y, \theta_y)$ are set equal to
$(1/8, 16, 12, 15, 3, 1.1, 10^{-10}, 1.3, 4, 2, 3.7)$, similarly
to~\cite{Wilson.excitatory.72, Muldoon.stimulation.16}. Node
dynamics are measured using the firing rate of the excitatory
population $\Set{y_{\nu t}}$. Simulated data were generated by
Matlab using Heun's method under a sampling rate of
$1\text{msec}$ in order to obtain $5\text{sec}$ ($5,000$ samples)
of simulated brain activity per subject. For each subject, the
simulated brain activity resulted in $N_{\mathpzc{G}} = 83$
time series. Each subject's brain activity represents a unique
state and the results of clustering are compared to this ground
truth.

\begin{figure}[!t]
  \centering
  \includegraphics[width=.75\linewidth]{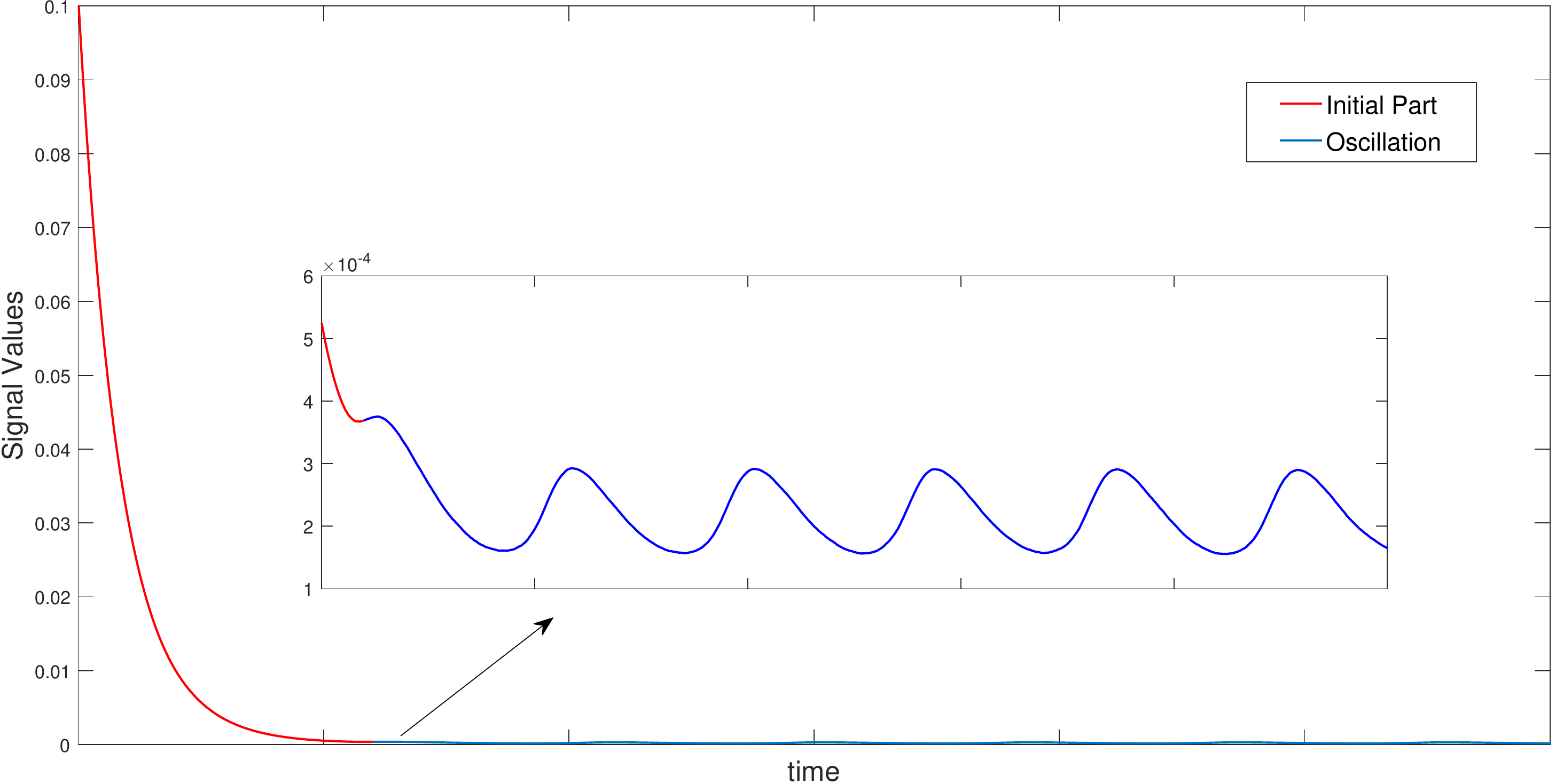}
  \caption{Single regional-brain-activity signal generated by the
    structural connectivity matrices of
    Fig.~\ref{fig:USAF.Conn.Mtrx} and model
    \eqref{Sarah-Henry.model}.}\label{fig:Real.Signal}
\end{figure}

\begin{figure}[!t]
  \centering
  \includegraphics[width=.8\linewidth]{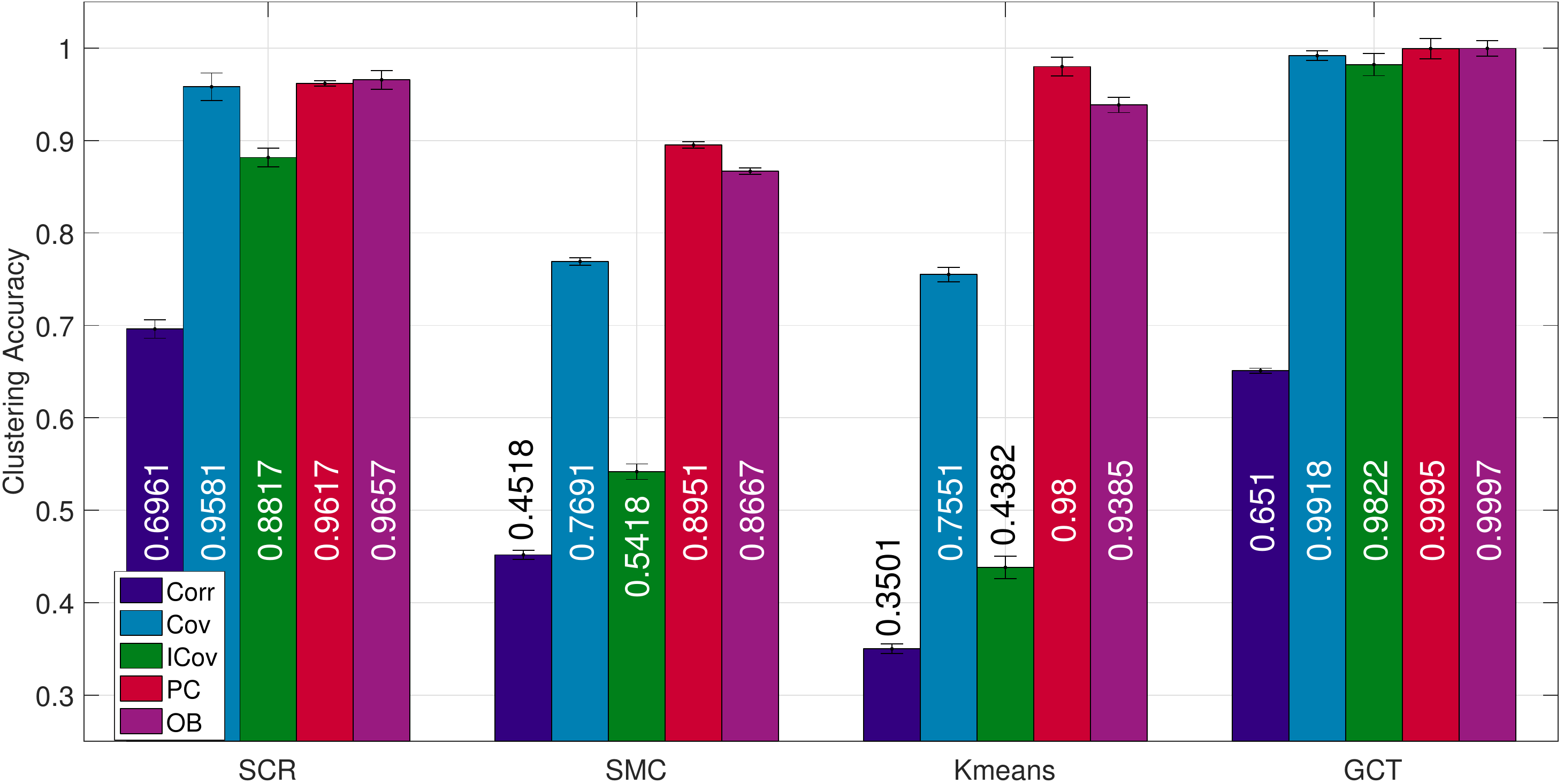}
  \caption{Linear kernel: $N_{\text{NN}}^{\text{GCT}} = 16$.} \label{fig:real.linear.kernel}
\end{figure}

\begin{figure}[!t]
  \centering
  \includegraphics[width=.8\linewidth]{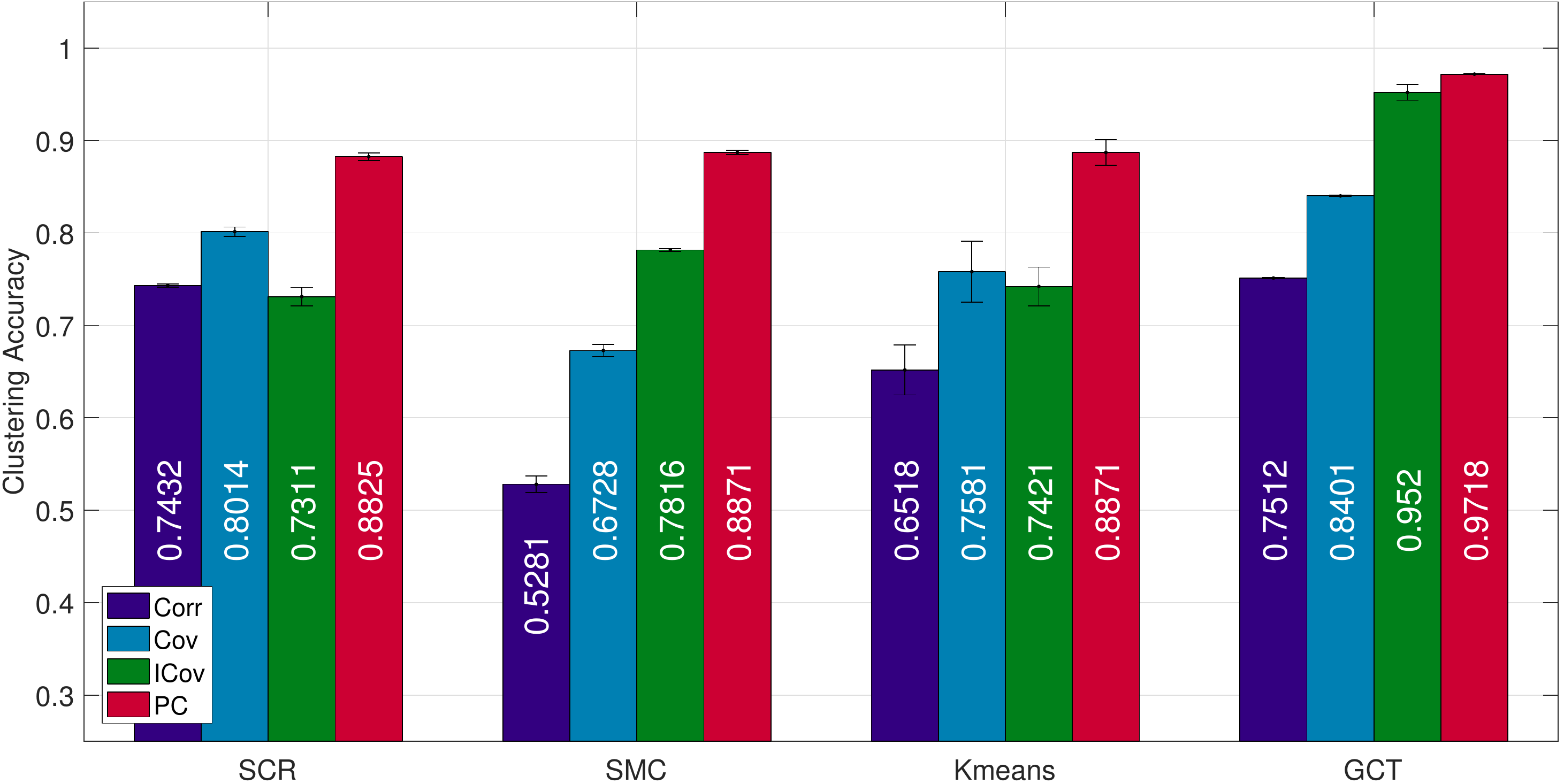}
  \caption{SDE: $N^{\text{SDE}}_{\nu t}= 25$; $N_{\mathpzc{G}} =
    83$; $N_{\text{NN}}^{\text{GCT}} =16$.}\label{fig:real.SDE}
\end{figure}

An example of the time series $(y_{\nu t})_t$, for a single
subject and a specific node, is shown in
Fig.~\ref{fig:Real.Signal}. As noted in the figure, there is an
initial and an oscillation mode of the time series. Per node
$\nu$, $500$ samples from the initial phase and $500$ ones from
the oscillation phase of the signal comprise the time series
$(y_{\nu t})_{t=1}^{1,000}$. The sliding-window lengths
$\tau_{\text{w}} \in\Set{500, 600, 700, 900}$ were tested. Length
$\tau_{\text{w}} = 700$ produced better results than those of
$500$ and $600$, for all clustering methods, while there was no
significant improvement by setting $\tau_{\text{w}}$ equal to
$900$. For this reason, only results for $\tau_{\text{w}} = 700$
are shown here. As in Sec.~\ref{sec:synthetic.data}, both the
methodologies of Secs.~\ref{sec:ARMA} and \ref{sec:PC} are
applied to this set of data, under choices of the linear, single
Gaussian, and the multi-kernel functions, as well as the SDE
approach. In the multi-kernel case, a weighted average of
Gaussian kernels is used, \ie,
$\kappa := (1/I) \sum_{i=1}^I \kappa_{\sigma_i}$, where
$\sigma_i\in \Set{0.25 + 0.01(i-1)}_{i=1}^{I}$, with $I:= 76$,
$\sigma_1 = 0.25$, and $\sigma_I =
1$. Figs.~\ref{fig:real.linear.kernel}--%
\ref{fig:real.SDE} show that GCT exhibits the most robust
performance among all methods. The best clustering-accuracy
result among Figs.~\ref{fig:real.linear.kernel}--%
\ref{fig:real.SDE} is recorded for GCT in
Fig.~\ref{fig:real.linear.kernel}, with a value of $0.9997$ for
the ``OB'' feature.

\begin{figure}[!t]
  \centering
  \includegraphics[width=.8\linewidth]{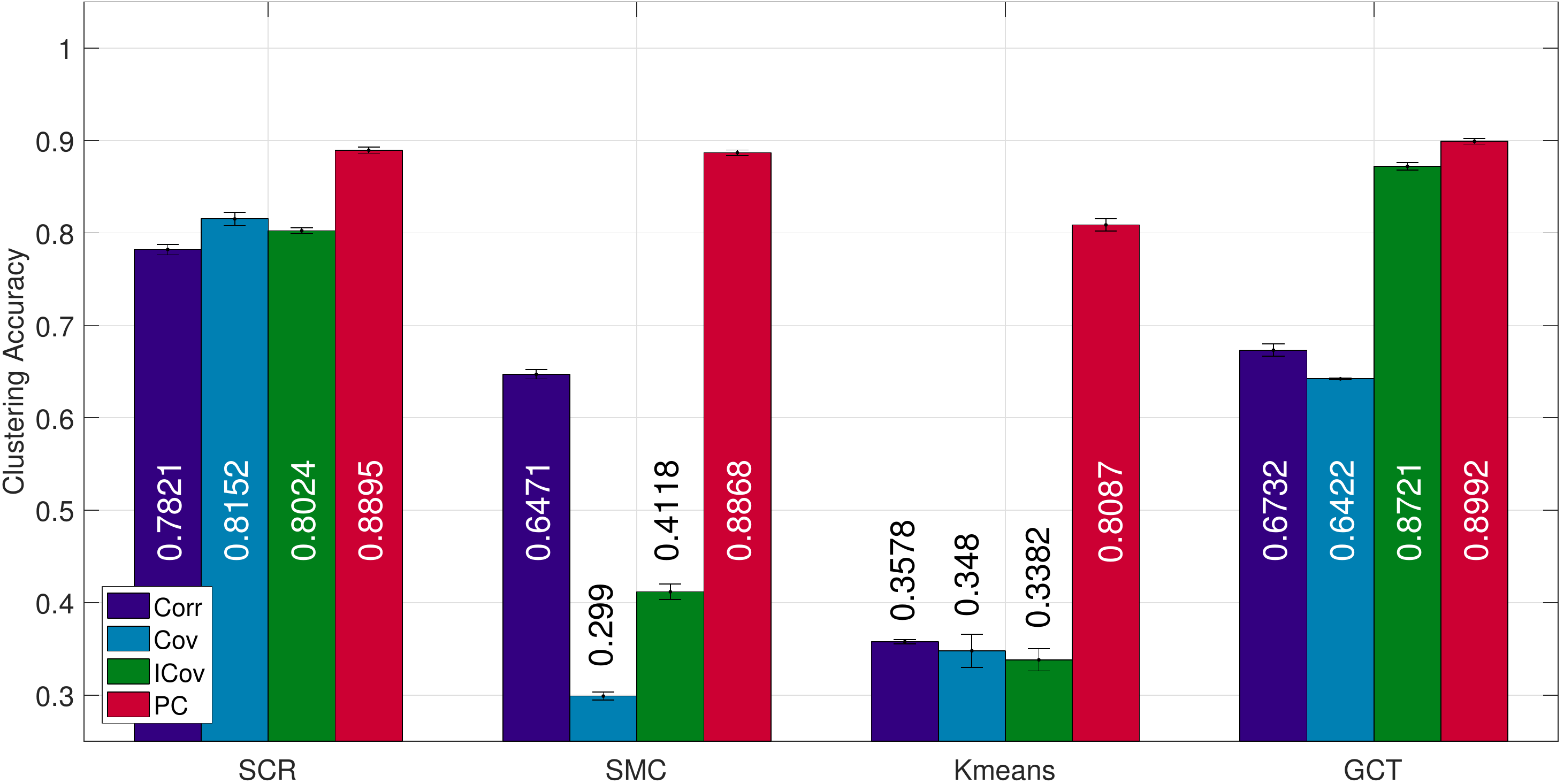}
  \caption{Single Gaussian kernel: $\sigma^2=0.5$;
    $N_{\text{NN}}^{\text{GCT}} =
    16$.}\label{fig:real.single.kernel.05}
\end{figure}

\begin{figure}[!t]
  \centering
  \includegraphics[width=.8\linewidth]{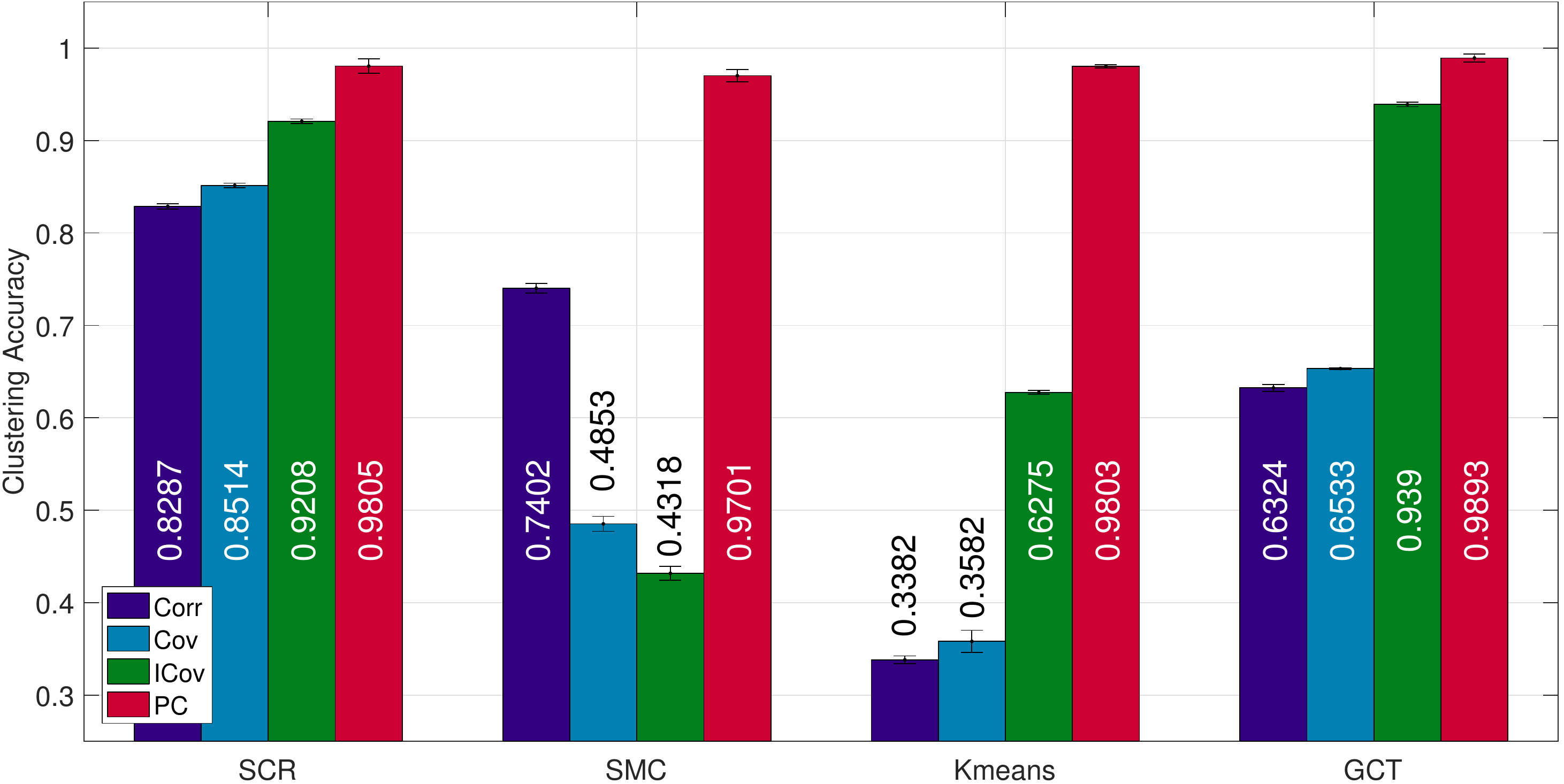}
  \caption{Single Gaussian kernel:
    $\sigma^2=1$; $N_{\text{NN}}^{\text{GCT}}
    =16$.} \label{fig:real.single.kernel.1} 
\end{figure}

\begin{figure}[!t]
  \centering
  \includegraphics[width=.8\linewidth]{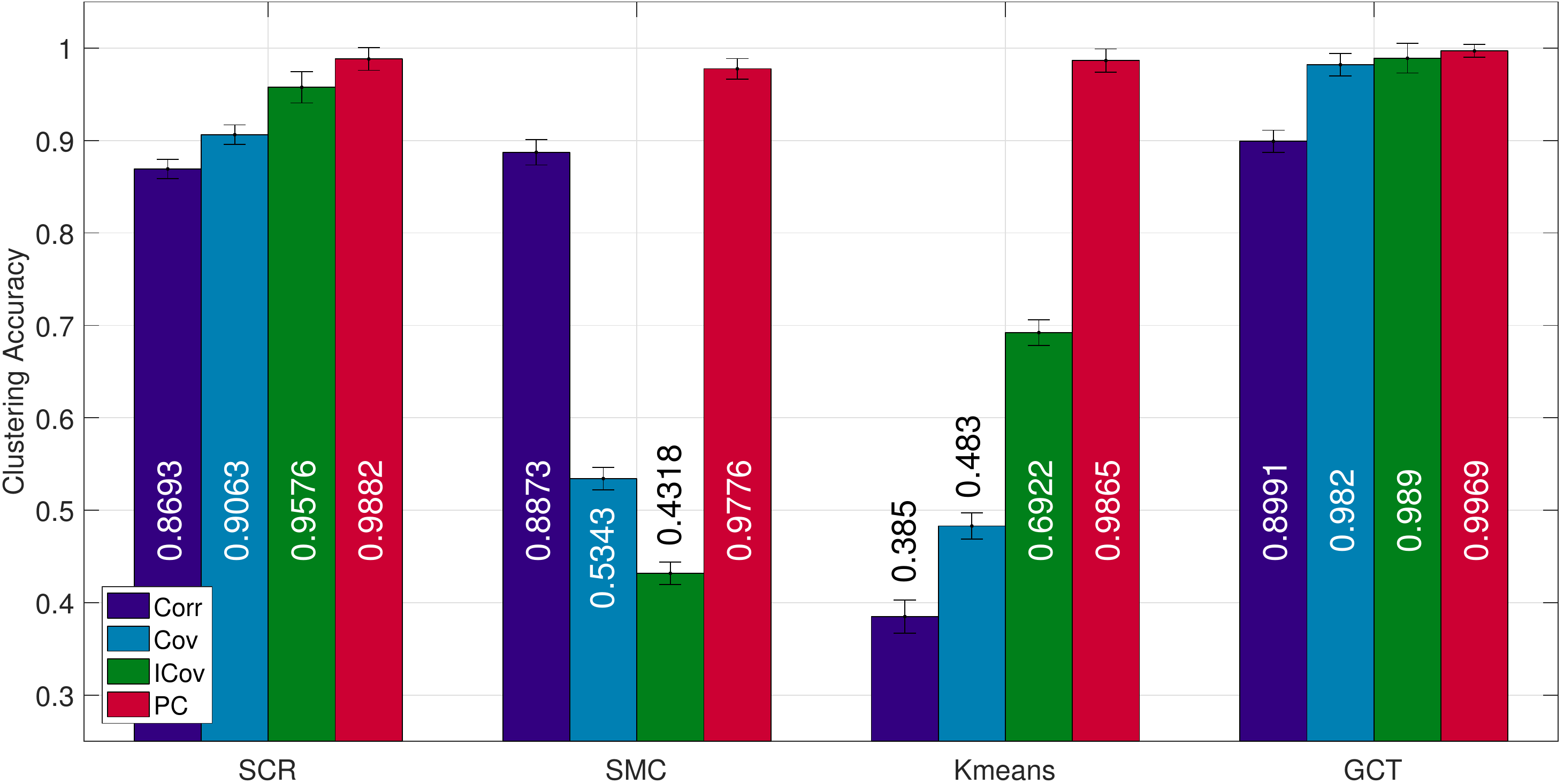}
  \caption{Multi-kernel:
    $N_{\text{NN}}^{\text{GCT}} =16$.}\label{fig:real.multi.kernel} 
\end{figure}

\section{Conclusions and the Road Ahead}\label{sec:conclusions}

This paper introduced Riemannian multi-manifold modeling (RMMM)
in the context of network-wide non-stationary time-series
analysis. Features extracted sequentially from time series were
used to define points in a Riemannian manifold, which under the
RMMM hypothesis, are located in or close to a union of multiple
Riemannian submanifolds. Two feature-generation mechanisms for
network-wide time series were introduced:
\begin{enumerate*}[label=\textbf{(\roman*)}]
\item Motivated by Granger-causality arguments, an
  auto-regressive moving average model was proposed to map
  low-rank linear vector subspaces, spanned by column vectors of
  appropriately defined observability matrices, to points into
  the Grassmann manifold; and
\item to capture dynamic (non-linear) relations among
  nodes, kernel-based partial correlations were
  introduced to generate points in the manifold of
  positive-definite matrices.
\end{enumerate*}
Furthermore, based on the very recent~\cite{MMM.arxiv.14,
  MMM.15}, a clustering algorithm was introduced to segment the
multiple Riemannian submanifolds which fit the data
patterns. Extensive numerical tests demonstrated that the
advocated framework outperforms classical and state-of-the-art
techniques. On-going research focuses on
\begin{enumerate*}[label=\textbf{(\roman*)}]
\item building an online spectral clustering scheme to alleviate
  the computational burden of Step~\ref{alg:affinity.matrix} in
  Alg.~\ref{algo:GCT}; and
\item applying the RMMM hypothesis to community detection
  scenarios, without any a-priori knowledge on the number of
  clusters.
\end{enumerate*}

\appendices

\section{Reproducing Kernels}\label{app:RKHS}

A real Hilbert space $\mathpzc{H}$, with elements denoted by $f$
and inner product $\innerp{\cdot}{\cdot}_{\mathpzc{H}}$, is
called a \textit{reproducing kernel Hilbert space (RKHS)}
\cite{Scholkopf.Smola.book, Aronszajn, Slavakis.eref.14}
whenever, for an arbitrarily fixed row vector
$\bm{y}\in\Real^{\tau_{\text{w}}}$, the mapping
$f\mapsto f(\bm{y})$ is continuous on $\mathpzc{H}$. This
condition is equivalent to the existence of a (unique)
\textit{reproducing kernel function}
$\kappa(\cdot,\cdot): \Real^{\tau_{\text{w}}} \times
\Real^{\tau_{\text{w}}} \rightarrow \Real$ which satisfies:
\begin{enumerate*}[label=\textbf{(\roman*)}]

\item $\varphi(\bm{y}) := \kappa(\bm{y},\cdot) \in\mathpzc{H}$,
  $\forall \bm{y}\in\Real^{\tau_{\text{w}}}$, and

\item the following \textit{reproducing property} holds:
  $f(\bm{y}) = \innerp{f}{\varphi(\bm{y})}_{\mathpzc{H}} =
  \innerp{f}{\kappa(\bm{y},\cdot)}_{\mathpzc{H}}$,
  $\forall \bm{y}\in\Real^{\tau_{\text{w}}}, \forall
  f\in\mathpzc{H}$.

\end{enumerate*}
If $f$ is chosen to be $\kappa(\bm{y}', \cdot)$, then the
previous reproducing property boils down to the so-called
\textit{kernel trick:}
$\kappa(\bm{y}',\bm{y}) = \innerp{\kappa(\bm{y}',\cdot)}
{\kappa(\bm{y},\cdot)}_{\mathpzc{H}}$,
$\forall \bm{y}, \bm{y}'\in \Real^{\tau_{\text{w}}}$. It turns
out that
$\mathpzc{H} = \overline{\linspan}\Set{\kappa(\bm{y},\cdot):
  \bm{y}\in \Real^{\tau_{\text{w}}}}$, where $\linspan$ stands
for the set of all linear combinations of the elements of a set,
and the overline symbol denotes closure, in the strong-topology
sense.

The previous definition has a more convenient algebraic
characterization. Kernel $\kappa$ is called \textit{positive
  definite}\/ if it is symmetric, \ie,
$\kappa(\bm{y}',\bm{y}) = \kappa(\bm{y},\bm{y}')$, for any
$\bm{y}, \bm{y}'\in \Real^{\tau_{\text{w}}}$, and
$\sum_{i=1}^I \sum_{j=1}^I \alpha_i \alpha_j \kappa(\bm{y}_i,
\bm{y}_j)\geq 0$, for any $\Set{\alpha_i}_{i=1}^I\subset \Real$,
any $\Set{\bm{y}_i}_{i=1}^I\subset \Real^{\tau_{\text{w}}}$, and
any $I\in\IntegerPP$. The positive definiteness of $\kappa$ can
be stated equivalently via the property that the \textit{kernel
  matrix\/} $\vect{K}$, defined by
$[\vect{K}]_{ij}:= \kappa(\bm{y}_i, \bm{y}_j)$, is positive
semidefinite, since
$\sum_i \sum_j \alpha_i \alpha_j \kappa(\bm{y}_i, \bm{y}_j) =
\bm{\alpha}^{\top} \vect{K}\bm{\alpha}$, for
$\bm{\alpha} := [\alpha_1, \ldots, \alpha_I]^{\top}$. Remarkably,
positive definiteness of a kernel characterizes its reproducing
property. Indeed, the reproducing kernel $\kappa$ of an RKHS
$\mathpzc{H}$ is positive definite~\cite{Slavakis.eref.14}, and
given a positive definite kernel $\kappa$, there exists a unique
RKHS $\mathpzc{H}$ s.t.\ $\kappa$ is the reproducing kernel of
$\mathpzc{H}$~\cite{Moore.PD.1916}.

Celebrated examples of reproducing kernels are the
\begin{enumerate*}[label=\textbf{(\roman*)}]

\item linear kernel:
  $\kappa_{\text{l}}(\bm{y}, \bm{y}') := \bm{y}\bm{y}'{}^{\top}$
  (recall that $\bm{y}, \bm{y}'$ are row vectors). In this case,
  $\mathpzc{H}= \Real^{\tau_{\text{w}}}$,
  $\varphi(\bm{y}) = \bm{y}$, and
  $\vect{K} = \vect{Y} \vect{Y}^{\top}$, where $\vect{Y}$ is the
  matrix whose rows are vectors $\Set{\bm{y}_i}_{i=1}^I$;

\item polynomial kernel:
  $\kappa_{\text{p}}(\bm{y}, \bm{y}') := (\bm{y} \bm{y}'{}^{\top}
  + 1)^{q}$, where $q\in\IntegerPP$; and the

\item Gaussian kernel:
  $\kappa_{\sigma}(\bm{y}, \bm{y}') := \exp[-\norm{\bm{y} -
    \bm{y}'}^2/ (2\sigma^2)]$, for some $\sigma\in\RealPP$. It
  turns out that $\dim\mathpzc{H}_{\sigma} = +\infty$, \eg,
  \cite{Slavakis.eref.14}.

\end{enumerate*}

\section{Logarithm Maps of
  $\text{Gr}(mN_{\mathpzc{G}}, p\rho)$ and  $\text{PD}(N_{\mathpzc{G}})$}\label{app:LogMap}

An efficient way to compute the logarithm map of the Grassmannian
$\text{Gr}(mN_{\mathpzc{G}}, p\rho)$, under a computational
complexity of $\mathcal{O}(mN_{\mathpzc{G}}p^2\rho^2)$, is
provided in \cite{Gallivan03efficientalgorithms}. Per point $x_t$
of $\text{Gr}(mN_{\mathpzc{G}}, p\rho)$,
\cite{Gallivan03efficientalgorithms} requires an
$mN_{\mathpzc{G}} \times mN_{\mathpzc{G}}$ orthogonal matrix
$\vect{O}$, having its first $p\rho$ columns, denoted by the
$mN_{\mathpzc{G}}\times p\rho$ matrix $\vect{L}$, span the
subspace $x_t$. Given $x_t$ and $x_{t'}$ of the Grassmannian, or
equivalently, pairs ($\vect{O}, \vect{L}$) and
($\vect{O}', \vect{L}'$), to compute $\log_{x_t} (x_{t'})$,
the SVDs of $\vect{L}^{\top}\vect{L}'$ and
$\vect{O}^{\top}\vect{L}'$ are needed.

Regarding manifold $\text{PD}(N_{\mathpzc{G}})$,
\cite{Tuzel.07} computes logarithm
$\log_{\vect{M}}(\vect{M}')$,
$\vect{M}, \vect{M}'\in \text{PD}(N_{\mathpzc{G}})$, by first
computing the Cholesky decomposition $\vect{M}= \vect{G}^2$, for a
symmetric $\vect{G}$, and by forming
$\log_{\vect{M}}(\vect{M}') = \vect{G} \log(\vect{G}^{-1}
\vect{M}' \vect{G}^{-1}) \vect{G}$, where $\log$ denotes the
matrix logarithm, under overall complexity
$\mathcal{O}(N_{\mathpzc{G}}^3)$. 

\section{Proof of Proposition~\ref{prop:kPC}}\label{app:prop.kPC} 

To reduce clutter, subscript $t$ will be dropped from all
subsequent symbols. Moreover,
$\tilde{\bm{y}}_{\nu} := \tilde{\bm{y}}_{\nu t}$,
$\tilde{\vect{Y}}_{-12} := \tilde{\vect{Y}}_{-12,t}$,
$\varphi_{\nu} := \varphi(\tilde{\bm{y}}_{\nu})$, and
$\bm{\varphi}_{-12} := \bm{\varphi}(\tilde{\vect{Y}}_{-12})$.

Assuming w.l.o.g.\ that $i<j$, then there exists
an $N_{\mathpzc{G}}\times N_{\mathpzc{G}}$ permutation matrix
$\vect{Q}$ s.t.\
\begin{align*}
  \bm{\Pi} := \vect{Q} \vect{K} \vect{Q}^{\top}
  & = 
    \begin{tikzpicture}[mymatrixenv,baseline=-0.5ex]
      \matrix (m) [mymatrix, ampersand replacement=\&]{
        \norm{\varphi_i}^2_{\mathpzc{H}} \&
        \innerp{\varphi_i}{\varphi_j}_{\mathpzc{H}} \& {} \&
        \bm{k}_{-ij,i} \\
        \innerp{\varphi_j}{\varphi_i}_{\mathpzc{H}} \&
        \norm{\varphi_j}^2_{\mathpzc{H}} \& {} \& \bm{k}_{-ij,j} \\
        {} \& {} \& {} \& {} \\
        \bm{k}_{-ij,i}^{\top} \& \bm{k}_{-ij,j}^{\top} \& {} \&
        \vect{K}_{-ij} \\ 
      };
      \draw (m-1-3.north) -- (m-4-3.south);
      \draw (m-3-1.west) -- (m-3-4.east);
    \end{tikzpicture}\\
  & =: 
    \begin{tikzpicture}[mymatrixenv,baseline=-0.5ex]
      \matrix[mymatrix, ampersand replacement=\&] (m) {
        \bm{\Pi}_{11} \& {} \& \bm{\Pi}_{12}\\
        {} \& {} \& {} \\
        \bm{\Pi}_{21} \& {} \& \bm{\Pi}_{22}\\
      };
      \mymatrixbraceleft{1}{1}{\footnotesize $2$}
      \mymatrixbracetop{1}{1}{\footnotesize $2$}
      \mymatrixbracetop{3}{3}{\footnotesize $N-2$}
      \draw (m-1-2.north) -- (m-3-2.south);
      \draw (m-2-1.west) -- (m-2-3.east);
    \end{tikzpicture}\hspace{-1.5em}
  \,.
\end{align*}
Indeed, $\vect{Q}$ can be defined by swapping the $1$st and
$i$th row, as well as the $2$nd and $j$th row of the identity matrix
$\vect{I}_{N_{\mathpzc{G}}}$. According to \eqref{gSC},
$\vect{K} / \vect{K}_{-ij} = \bm{\Pi} / \vect{K}_{-ij} = \bm{\Pi} /
\bm{\Pi}_{22}$.

By standard arguments of LS estimation, for $l\in\Set{i,j}$,
\begin{align}
  \hat{\bm{\beta}}_l
  & \in \Argmin\nolimits_{\bm{\beta}\in \Real^{N_{\mathpzc{G}}-2}}\,
    \norm*{\varphi_l - \bm{\beta}
    \bm{\varphi}_{-{ij}}}_{\mathpzc{H}}^2 \notag\\
  & = \Argmin\nolimits_{\bm{\beta}}\,
    \norm*{\varphi(\tilde{\bm{y}}_l) -
    \sum\nolimits_{\nu\in\mathpzc{V}_{-ij}} \beta_{\nu}
    \varphi(\tilde{\bm{y}}_{\nu})}_{\mathpzc{H}}^2 \label{define.beta}
\end{align}
yields the orthogonal projection
$\hat{\varphi}_l := \hat{\bm{\beta}}_l \bm{\varphi}_{-{12}}$ of
$\varphi_l$ onto the closed linear subspace spanned by
$\{\varphi_{\nu}\}_{\nu\in\mathpzc{V}_{-ij}}$. As such,
$\hat{\bm{\beta}}_l$ satisfies the normal equations
$\hat{\bm{\beta}}_l \vect{K}_{-{12}} = \bm{k}_{-12,l}$, since
$\vect{K}_{-{12}}$ in \eqref{K-ij} is the Gram matrix formed by
$\Set{\varphi_{\nu}}_{\nu\in\mathpzc{V}_{-ij}}$. Hence, the
minimum-norm $\hat{\bm{\beta}}_l$ of \eqref{define.beta} can be obtained
by $\bm{k}_{-12,l}\vect{K}_{-{12}}^{\dagger}$. Clearly,
$\hat{\varphi}_l :=
\bm{k}_{-12,l} \vect{K}_{-{12}}^{\dagger}\bm{\varphi}_{-{12}}$, which
justifies \eqref{LS.estimate}.

Now, it can be verified that%
\begin{subequations}\label{derive.innerp}
  \begin{alignat}{2}
    \innerp*{\prescript{}{\kappa}{\tilde{r}}_i}
    {\prescript{}{\kappa}{\tilde{r}}_j}_{\mathpzc{H}} 
    & \,\mathbin{=}\, && \innerp*{\varphi_i - \hat{\varphi}_i}{\varphi_j -
      \hat{\varphi}_j}_{\mathpzc{H}} \notag\\
    & \,\mathbin{=}\, && \left\langle\left.{\varphi_i -
          \sum\nolimits_{\nu\in\mathpzc{V}_{-ij}} [\bm{k}_{-ij,i}
          \vect{K}_{-{ij}}^{\dagger}]_{\nu}
          \varphi_{\nu}}\right|\right. \notag\\
    &&& \qquad\left.{\varphi_j - \sum\nolimits_{\nu'\in\mathpzc{V}_{-ij}}
        [\bm{k}_{-ij,j} \vect{K}_{-{ij}}^{\dagger}]_{\nu'}
        \varphi_{\nu'}}
    \right\rangle_{\mathpzc{H}} \notag\\
    & \,\mathbin{=}\, && \innerp{\varphi_i}{\varphi_j}_{\mathpzc{H}}
    - 2\bm{k}_{-ij,i} 
    \vect{K}_{-{ij}}^{\dagger} \bm{k}_{-ij,j}^{\top} \notag \\
    &&& + \bm{k}_{-ij,1} \vect{K}_{-{ij}}^{\dagger}
    \vect{K}_{-{ij}} \vect{K}_{-ij}^{\dagger}
    \bm{k}_{-ij,j}^{\top} \label{pi12.linearity}\\
    & \,\mathbin{=}\, && \innerp{\varphi_i}{\varphi_j}_{\mathpzc{H}}
    - 2\bm{k}_{-ij,i} 
    \vect{K}_{-{ij}}^{\dagger} \bm{k}_{-ij,j}^{\top} \notag\\
    &&& + \bm{k}_{-ij,i} \vect{K}_{-{ij}}^{\dagger}
    \bm{k}_{-ij,j}^{\top} \label{pi12.penrose}\\
    & \,\mathbin{=}\, && \innerp{\varphi_i}{\varphi_j}_{\mathpzc{H}}
    - \bm{k}_{-ij,i} \vect{K}_{-{ij}}^{\dagger}
    \bm{k}_{-ij,j}^{\top}  \notag\\ 
    & \,\mathbin{=}\, && [\bm{\Pi} / \bm{\Pi}_{22}]_{12}
    \,,\label{pi12}
  \end{alignat}
\end{subequations}%
where the linearity of the inner product was used in
\eqref{pi12.linearity}, and the properties of the Moore-Penrose
pseudoinverse in \eqref{pi12.penrose}. In a similar way to
\eqref{derive.innerp}, it can be verified that
\begin{align}
  \norm*{\prescript{}{\kappa}{\tilde{r}}_l}^2_{\mathpzc{H}}
  & = \norm*{\varphi_l - \hat{\varphi}_l}^2_{\mathpzc{H}} 
    = \norm{\varphi_l}^2_{\mathpzc{H}} - \bm{k}_{-ij,l}
    \vect{K}_{-{ij}}^{\dagger} \bm{k}_{-ij,l}^{\top} \notag\\ 
  & = [\bm{\Pi} / \bm{\Pi}_{22}]_{ll}\,, \quad
    l\in\Set{1,2}\,. \label{pi11}
\end{align}
Hence, \eqref{prop:kPC.eq1} follows from \eqref{pi12} and
\eqref{pi11}.

If $\vect{K}\succ \vect{0}$, then also $\bm{\Pi}\succ
\vect{0}$. This implies that
$\vect{K}_{-ij} = \bm{\Pi}_{22}\succ \vect{0}$,
$\bm{\Pi}_{11} \succ 0$,
$\bm{\Pi} / \bm{\Pi}_{22} \succ \vect{0}$, and
$\bm{\Pi} / \bm{\Pi}_{11} \succ \vect{0}$~\cite{Ben.Israel,
  Albert.Schur.complement}. Consequently,
$\vect{K}_{-{12}}^{\dagger} =
\vect{K}_{-{12}}^{-1}$~\cite{Ben.Israel}. If
$\bm{\Xi} := [\xi_{ll'}] := (\bm{\Pi}/ \bm{\Pi}_{22})^{-1}$, and
if $\minor_{ll'}(\cdot)$ stands for the $(l,l')$th minor of a
square matrix, Cramer's rule dictates that
$[ \bm{\Xi}^{-1}]_{ll'} = (1/\det\bm{\Xi})\cdot (-1)^{l+l'}
\cdot \minor_{l'l}(\bm{\Xi})$. Recall also the well-known
fact~\cite[p.~30]{Ben.Israel}:
\begin{align*}
  \bm{\Pi}^{-1} = 
  \begin{bmatrix}
    (\bm{\Pi}/\bm{\Pi}_{22})^{-1} &\hspace{-1em}
    -\bm{\Pi}_{11}^{-1} \bm{\Pi}_{12} 
    (\bm{\Pi}/\bm{\Pi}_{11})^{-1} \\
    -\bm{\Pi}_{22}^{-1} \bm{\Pi}_{21}
    (\bm{\Pi}/\bm{\Pi}_{22})^{-1} &\hspace{-1em}
    (\bm{\Pi}/\bm{\Pi}_{11})^{-1} 
  \end{bmatrix}\,,
\end{align*}
which suggests that $\bm{\Xi}$ is the $2\times 2$ upper-left
submatrix of $\bm{\Pi}^{-1}$. By \eqref{prop:kPC.eq1},
\eqref{prop:kPC.eq2} is established as follows:
\begin{align*}
  \prescript{}{\kappa}{\hat{\varrho}}_{ij}
  & = \frac{[\vect{K} / \vect{K}^{-ij}]_{12}}{\sqrt{[\vect{K} /
    \vect{K}_{-ij}]_{11} \cdot [\vect{K} / \vect{K}_{-ij}]_{22}}} \\
  & = \frac{[\bm{\Pi} / \bm{\Pi}_{22}]_{12}}{\sqrt{[\bm{\Pi} /
    \bm{\Pi}_{22}]_{11} \cdot [\bm{\Pi} / \bm{\Pi}_{22}]_{22}}}
    = \frac{[\bm{\Xi}^{-1}]_{12}}{\sqrt{[\bm{\Xi}^{-1}]_{11} \cdot
    [\bm{\Xi}^{-1}]_{22}}} \\
  & = \frac{(-1)^{1+2} \cdot \minor_{21}(\bm{\Xi})}
    {\left[(-1)^{1+1}\minor_{11}(\bm{\Xi})\cdot
    (-1)^{2+2}\minor_{22}(\bm{\Xi}) \right]^{1/2}} \\
  & = \frac{-\xi_{12}}{\sqrt{\xi_{22}\xi_{11}}} 
    = \frac{-[\bm{\Pi}^{-1}]_{12}}
    {\sqrt{[\bm{\Pi}^{-1}]_{22} \cdot
    [\bm{\Pi}^{-1}]_{11}}} \\
  & = \frac{-[\vect{Q} \vect{K}^{-1} \vect{Q}^{\top}]_{12}}
    {\sqrt{[\vect{Q} \vect{K}^{-1} \vect{Q}^{\top}]_{11} \cdot
    [\vect{Q} \vect{K}^{-1} \vect{Q}^{\top}]_{22}}} \\
  & = \frac{-[\vect{K}^{-1}]_{ij}}
    {\sqrt{[\vect{K}^{-1}]_{ii} \cdot
    [\vect{K}^{-1}]_{jj}}} \,.
\end{align*}

\section{Semidefinite Embedding}\label{app:SDE}

Along the lines of the discussion in Appendix~\ref{app:RKHS}, it
is likely that the geometry of $\Set{\tilde{\bm{y}}_{\nu t}}$ is
``destroyed'' during the transfer
$\Set{\tilde{\bm{y}}_{\nu t}}\mapsto
\Set{\varphi(\tilde{\bm{y}}_{\nu t})}$, if no constraints are
imposed on $\varphi$. To this end, the geometry of
$\Set{\tilde{\bm{y}}_{\nu t}}$ needs to be learned first. A graph
is built on $\Set{\tilde{\bm{y}}_{\nu t}}$, and a weighted
adjacency matrix $\bm{\Omega}_t$, as well as neighborhoods
$\Set{\mathcal{N}^{\text{SDE}}_{\nu
    t}}_{\nu=1}^{N_{\mathpzc{G}}}$ are constructed. A
straightforward way is:
\begin{enumerate*}[label=\textbf{(\roman*)}]

\item Per node $\nu$, gather in
  $\mathcal{N}^{\text{SDE}}_{\nu t}$ the (user-defined)
  $P\in\IntegerPP$ nearest neighbors (in a Euclidean-distance
  sense, for example) of $\tilde{\bm{y}}_{\nu t}$ among
  $\Set{\tilde{\bm{y}}_{\nu' t}}_{\nu'\neq \nu}$, including also
  $\tilde{\bm{y}}_{\nu t}$;

\item define $\bm{\Omega}_t := [\omega_{\nu\nu',t}]$ as follows:
  $w_{\nu\nu',t} := 1/P$, if
  $\tilde{\bm{y}}_{\nu' t}\in \mathcal{N}^{\text{SDE}}_{\nu t}$,
  and $w_{\nu\nu'} := 0$, otherwise. Clearly, data vectors
  $\tilde{\bm{y}}_{\nu t}$ and $\tilde{\bm{y}}_{\nu' t}$ belong
  to the same neighborhood iff there exists $\nu''$ s.t.\
  $\tilde{\bm{y}}_{\nu t}, \tilde{\bm{y}}_{\nu' t}\in
  \mathcal{N}^{\text{SDE}}_{\nu''t}$ iff $\exists\nu''$ with
  $\omega_{\nu'' \nu,t}\cdot \omega_{\nu''\nu',t}>0$.

\end{enumerate*}

SDE postulates that data geometry, at least within neighborhoods
defined via the previous step \textbf{(i)}, should be preserved
even after mapping data into $\mathpzc{H}$. For neighbors
$\tilde{\bm{y}}_{\nu t}, \tilde{\bm{y}}_{\nu' t}$, distances
should satisfy the \textit{isometric}\/ condition:
$\norm{\varphi(\tilde{\bm{y}}_{\nu t}) -
  \varphi(\tilde{\bm{y}}_{\nu' t})}_{\mathpzc{H}}^2 =
\norm{\tilde{\bm{y}}_{\nu t} - \tilde{\bm{y}}_{\nu' t}}^2_2$. By
the kernel trick, the previous constraint translates to
$[\vect{K}_t]_{\nu\nu} - 2[\vect{K}_t]_{\nu\nu'} +
[\vect{K}_t]_{\nu'\nu'} = \norm{\tilde{\bm{y}}_{\nu t} -
  \tilde{\bm{y}}_{\nu' t}}^2_2$. Moreover, data are required to
be ``centered'' around $0$, \ie,
$\sum_{\nu=1}^{N_{\mathpzc{G}}} \varphi(\tilde{\bm{y}}_{\nu t}) =
0$. Again, by the kernel trick,
$\sum_{\nu} \varphi(\tilde{\bm{y}}_{\nu t}) = 0 \Leftrightarrow
\innerp{\sum_{\nu} \varphi(\tilde{\bm{y}}_{\nu t})}{\sum_{\nu'}
  \varphi(\tilde{\bm{y}}_{\nu' t})}_{\mathpzc{H}} =0
\Leftrightarrow \sum_{\nu} \sum_{\nu'} [\vect{K}_t]_{\nu\nu'} =
0$. Finally, the data cloud
$\{\varphi(\tilde{\bm{y}}_{\nu t})\}_{t=1}^T$ should occupy ``as
much space as possible'' within $\mathpzc{H}$. This can be
achieved by the maximization of the ``sample variance,'' which,
according to the previous constraints, becomes:
$\sum_{\nu=1}^{N_{\mathpzc{G}}} \norm{\varphi(\tilde{\bm{y}}_{\nu
    t}) - (1/N_{\mathpzc{G}}) \sum_{\nu'=1}^{N_{\mathpzc{G}}}
  \varphi(\tilde{\bm{y}}_{\nu' t})}^2_{\mathpzc{H}} =
\sum_{\nu=1}^{N_{\mathpzc{G}}} \norm{\varphi(\tilde{\bm{y}}_{\nu
    t})}^2_{\mathpzc{H}} = \sum_{\nu=1}^{N_{\mathpzc{G}}}
\innerp{\varphi(\tilde{\bm{y}}_{\nu t})}
{\varphi(\tilde{\bm{y}}_{\nu t})}_{\mathpzc{H}}=
\sum_{\nu=1}^{N_{\mathpzc{G}}} \kappa(\tilde{\bm{y}}_{\nu t},
\tilde{\bm{y}}_{\nu t}) = \trace(\vect{K}_t)$.

SDE is posed as the following linear (convex) programming
task over the set of PSD matrices: given data
$\Set{\tilde{\bm{y}}_{\nu t}}_{\nu=1}^{N_{\mathpzc{G}}}$ per $t$, as
well as the weighted adjacency matrix $\bm{\Omega}_t$, find
\begin{align*}\label{SDE}
  \vect{K}_t \in
  & \arg\max\nolimits_{\vect{K}} \trace(\vect{K})\notag\\
  & \text{s.to}\ \left[
    \begin{aligned}
      & \vect{K}\succeq\vect{0}\,,\\
      & \sum\nolimits_{\nu=1}^{N_{\mathpzc{G}}}
      \sum\nolimits_{\nu'=1}^{N_{\mathpzc{G}}}
      [\vect{K}]_{\nu\nu'} = 0\,,\\
      & \left[
        \begin{aligned}
          & [\vect{K}]_{\nu\nu} - 2[\vect{K}]_{\nu\nu'} +
          [\vect{K}]_{\nu'\nu'} = \norm{\tilde{\bm{y}}_{\nu t} -
            \tilde{\bm{y}}_{\nu' t}}^2_2\,,\\
          & \forall (\nu,\nu')\ \text{s.t.}\ \exists\nu''\ \text{with}\
          \omega_{\nu'' \nu,t} \cdot \omega_{\nu''\nu',t}>0\,.
        \end{aligned}\right.
    \end{aligned}\right.
\end{align*}

\bibliographystyle{IEEEtranS}
\bibliography{kostas.bib}
\end{document}